\documentclass[manuscript,screen]{acmart}

\AtBeginDocument{%
  }

\setcopyright{acmlicensed}
\copyrightyear{2018}
\acmYear{2018}
\acmDOI{XXXXXXX.XXXXXXX}

\acmConference[Conference acronym 'XX]{Make sure to enter the correct
  conference title from your rights confirmation emai}{June 03--05,
  2018}{Woodstock, NY}
\acmISBN{978-1-4503-XXXX-X/18/06}

\usepackage{longtable}
\usepackage{makecell}
\usepackage{tikz}
\usepackage{forest}
\usetikzlibrary{decorations.pathreplacing}
\usetikzlibrary{arrows.meta,shapes,positioning,shadows,trees}

\begin{document}

\title{Text to Image Generation and Editing: A Survey}

\author{Pengfei Yang}
\email{pfyangcrc@gmail.com}
\orcid{0009-0007-2847-4151}
\affiliation{%
  \institution{Singapore University of Technology and Design}
  \country{Singapore}}

\author{Ngai-Man CHEUNG}
\affiliation{%
  \institution{Singapore University of Technology and Design}
  \country{Singapore}}
\email{n.man.cheung@gmail.com}

\author{Xinda Ma}
\affiliation{%
  \institution{Singapore University of Technology and Design}
  \country{Singapore}}
\email{xinda_ma@sutd.edu.sg}

\renewcommand{\shortauthors}{Yang et al.}

\begin{abstract}
    Text-to-image generation (T2I) refers to the text-guided generation of high-quality images. In the past few years, T2I has attracted widespread attention and numerous works have emerged. In this survey, we comprehensively review 141 works conducted from 2021 to 2024. First, we introduce four foundation model architectures of T2I (autoregression, non-autoregression, GAN and diffusion) and the commonly used key technologies (autoencoder, attention and classifier-free guidance). Secondly, we systematically compare the methods of these studies in two directions, T2I generation and T2I editing, including the encoders and the key technologies they use. In addition, we also compare the performance of these researches side by side in terms of datasets, evaluation metrics, training resources, and inference speed. In addition to the four foundation models, we survey other works on T2I, such as energy-based models and recent Mamba and multimodality. We also investigate the potential social impact of T2I and provide some solutions. Finally, we propose unique insights of improving the performance of T2I models and possible future development directions. In summary, this survey is the first systematic and comprehensive overview of T2I, aiming to provide a valuable guide for future researchers and stimulate continued progress in this field.
\end{abstract}



\received{20 February 2007}
\received[revised]{12 March 2009}
\received[accepted]{5 June 2009}

\maketitle

\section{Introduction} \label{Introduction}
With the development of generative models (GAN, AR, Diffusion, NAR), text-to-image generation (T2I) has made greatly progress, and countless works have emerged. Our survey provides researchers with a holistic perspective, including a comparison of important existing works in community and some emerging research directions, in order to help researchers understand the development of T2I.

The selection criteria for the T2I papers surveyed in this review are: 1) We survey T2I papers from 2021 to 2024; 2) We survey T2I papers that are highly concerned in the community and their subsequent works, such as: LDM \cite{rombach2022high, podell2023sdxl, sauer2023adversarial, esser2024scaling}, Imagen \cite{saharia2022photorealistic, chen2022re, wang2023imagen}, DALL-E \cite{ramesh2021zero, ramesh2022hierarchical, betker2023improving}, Cogview \cite{ding2021cogview, ding2022cogview2, zheng2024cogview3}, and Pixart \cite{chen2023pixart, chen2024pixart}, etc.; 3) We survey T2I papers from top papers of conferences; 4) We survey T2I papers cited (compared or mentioned in related work) in papers selected according to the previous criteria; 5) When drafting this review, we focus on the latest interesting T2I papers in arXiv, some of which have been included in conferences when this article was completed. See Table~\ref{tab:mod} for details.

Previous T2I surveys often involve reviewing only a single foundation model; such as GAN \cite{agnese2020survey, frolov2021adversarial} or Diffusion \cite{zhang2303text, cao2024controllable}. Although some survey \cite{shuai2024survey, zhan2023multimodal, zhou2023vision+, bie2023renaissance, singh2023survey, alhabeeb2024text} investigate multiple foundation models, they are not as comprehensive as our review: 1) the number of T2I papers they review is too small, 2) they do not review some recent works, such as Mamba, 3) their comparison is not sufficient, and the side-by-side comparison used in our survey (see Table~\ref{tab:mod} and Table~\ref{tab:perf} in appendix) makes the gaps between different models clear at a glance, 4) compared with these reviews, our review makes a detailed investigation of social impacts and solutions, 5) based on the existing literature, our review points out more future research directions. Please see Table~\ref{tab:survey} for comparison. Note that \cite{cao2024controllable} and \cite{shuai2024survey} contain a large number of T2I papers but have little overlap with ours. The reasons are as follows. There are about 75\% of the T2I papers in \cite{cao2024controllable} come from arXiv, which is different from our paper selection criteria. The T2I papers that overlap with our paper in this survey are works that are highly concerned in community and some papers accepted by conferences. Unlike ours focusing on both image generation and editing, \cite{shuai2024survey} only focuses on image editing and does not mention well-known generation models such as LDM \cite{rombach2022high}, Imagen \cite{saharia2022photorealistic}, etc. In addition, since it focuses on multimodal generation, it contains a large number of LLM-based image editing works, which is not the focus of our work and we only briefly mention it in Section~\ref{Others}. The T2I papers that overlap with our paper in this survey are mainly well-known image editing models, such as DreamBooth \cite{ruiz2023dreambooth}, Textual Inversion \cite{gal2022image}, InstructPix2Pix \cite{brooks2023instructpix2pix}, etc.

\begin{table}[!ht]
    \centering
    \caption{Comparison of T2I surveys. \textbf{T2I Paper Number (Total)} refers to the number of T2I papers whose input modals contain text. If the input modals are modals other than text, these papers are not considered. Papers, included in surveys, that are not T2I papers but are only related papers, such as foundation models or basic techniques, are not considered. \textbf{T2I Paper Number (Common)} refers to the number of selected T2I papers that overlap with the T2I papers in our survey.}
    \tiny
    \begin{tabular}{ccccccccc}
        \toprule
        Title & \makecell{First submission \\ on Arxiv}  & Published & \makecell{T2I Paper Number \\(Total/Common)} & Foundation Models & Input Modal & Datasets & Metrics & Social Impacts \\ 
        \midrule
        \cite{agnese2020survey} & 2019.10.21 & / & 22 / 0 & GAN & Text & \checkmark & \checkmark & \\ 
        \cite{frolov2021adversarial} & 2021.01.25 & 21'ACM Neural Networks & 46 / 0 & GAN & Text & \checkmark & \checkmark & \\ 
        \cite{zhang2303text} & 2023.03.14 & / & 44 / 19 & Diffusion & Text & & & \checkmark \\ 
        \cite{cao2024controllable} & 2024.03.07 & / & 165 / 19 & Diffusion & Text, Visual, Audio & & &  \\ 
        \cite{shuai2024survey} & 2024.06.20 & / & 131 / 24 & Diffusion & Text, Visual,  Others & & & \checkmark \\ 
        \cite{zhan2023multimodal} & 2021.12.27 & 23'TPAMI & 21 / 14 & GAN, Diffusion, AR, NeRF & Text, Visual, Audio, Others & & & \checkmark \\
        \cite{zhou2023vision+} & 2023.05.24 & 23'CVPR & / & GAN, Diffusion, AR & Text, Visual, Audio, Others & \checkmark & \checkmark & \checkmark \\
        \cite{bie2023renaissance} & 2023.09.02 & / & 29 / 19 & GAN, Diffusion, AR & Text & \checkmark & \checkmark &  \\
        \cite{singh2023survey} & 2023.11.10 & 23'IEEE AIRC & 3 / 3 & Diffusion, AR & Text & & &  \\
        \cite{alhabeeb2024text} & / & 24'IEEE Access & 70 / 41 & GAN, Diffusion & Text & \checkmark & \checkmark &  \\
        Ours & / & / & 141 / 141 & GAN, Diffusion, AR, NAR & Text & \checkmark & \checkmark & \checkmark \\
        \bottomrule
    \end{tabular}
    \label{tab:survey}
\end{table}

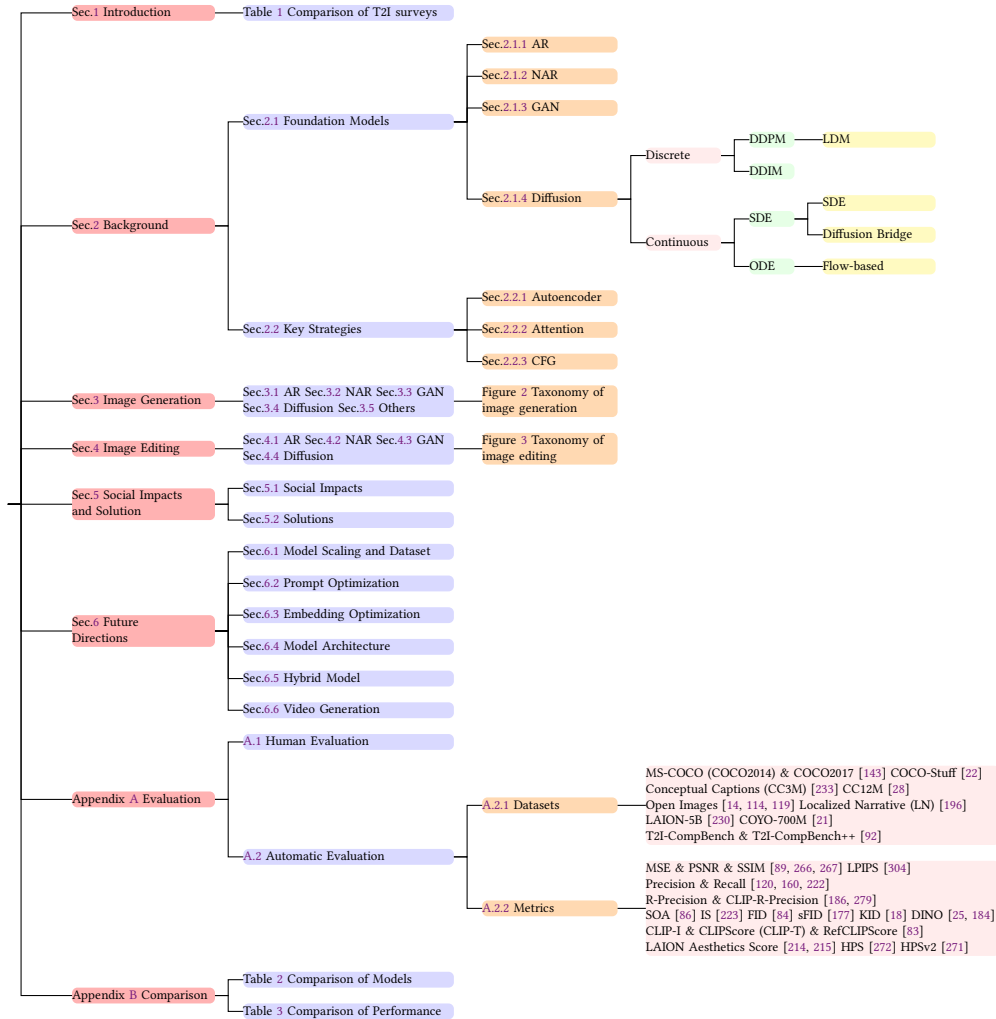
\begin{figure} [!ht]
\centering
\tikzset{
    rnode/.style = {thin, rounded corners=2pt, align=left, fill=red!30, text width=1.9cm},
    bnode/.style = {thin, rounded corners=2pt, align=left, fill=blue!15, text width=2.8cm},
    onode/.style = {thin, rounded corners=2pt, align=left, fill=orange!30, text width=1.8cm},
    pnode/.style = {thin, rounded corners=2pt, align=left, fill=pink!30, text width=1cm},
    gnode/.style = {thin, rounded corners=2pt, align=left, fill=green!10, text width=0.6cm},
    ynode/.style = {thin, rounded corners=2pt, align=left, fill=yellow!30, text width=1.5cm},
    ppnode/.style = {thin, rounded corners=2pt, align=left, fill=pink!30, text width=4.7cm},
}

\begin{forest}
for tree={
    inner sep=0pt,
    outer sep=0pt,
    fit=band,
    child anchor=west,
    parent anchor=east,
    grow'=0,
    anchor=west,
    align=left,
    edge path={\noexpand\path[\forestoption{edge}] 
         (!u.parent anchor) -- +(5pt,0pt) |- (.child anchor)
         \forestoption{edge label};},
    font=\tiny, 
}   
[
    [Sec.\ref{Introduction} Introduction, rnode
        [Table~\ref{tab:survey} Comparison of T2I surveys, bnode]
    ]
    [Sec.\ref{Background} Background, rnode
        [Sec.\ref{Foundation Models} Foundation Models, bnode
            [Sec.\ref{2Autoregression} AR, onode]
            [Sec.\ref{2NAR} NAR, onode]
            [Sec.\ref{2GAN} GAN, onode]
            [Sec.\ref{2Diffusion} Diffusion, onode
                [Discrete, pnode
                    [DDPM, gnode
                        [LDM, ynode]
                    ]
                    [DDIM, gnode]
                ]
                [Continuous, pnode
                    [SDE, gnode
                        [SDE, ynode]
                        [Diffusion Bridge, ynode]
                    ]    
                    [ODE, gnode
                        [Flow-based, ynode]
                    ]            
                ]
            ]
        ]
        [Sec.\ref{Key Strategies} Key Strategies, bnode
            [Sec.\ref{Autoencoder} Autoencoder, onode]
            [Sec.\ref{Attention} Attention, onode]
            [Sec.\ref{CFG} CFG, onode]
        ]
    ]
    [Sec.\ref{Image Generation} Image Generation, rnode
        [Sec.\ref{3Autoregression} AR Sec.\ref{3NAR} NAR Sec.\ref{3GAN} GAN \\
        Sec.\ref{3Diffusion} Diffusion Sec.\ref{Others} Others, bnode
            [Figure~\ref{fig:generation} Taxonomy of \\ image generation, onode]
        ]
    ]
    [Sec.\ref{Image Editing} Image Editing, rnode
        [Sec.\ref{4Autoregression} AR Sec.\ref{4NAR} NAR Sec.\ref{4GAN} GAN \\
        Sec.\ref{4Diffusion} Diffusion, bnode
            [Figure~\ref{fig:editing} Taxonomy of \\ image editing, onode]
        ]
    ]
    [Sec.\ref{Social Impacts and Solution} Social Impacts \\
    and Solution, rnode
        [Sec.\ref{Social Impacts} Social Impacts, bnode]
        [Sec.\ref{Solutions} Solutions, bnode]
    ]
    [Sec.\ref{Future Directions} Future \\ Directions, rnode
        [Sec.\ref{Model Scaling and Dataset} Model Scaling and Dataset, bnode]
        [Sec.\ref{Prompt Optimization} Prompt Optimization, bnode]
        [Sec.\ref{Embedding Optimization} Embedding Optimization, bnode]
        [Sec.\ref{Model Architecture} Model Architecture, bnode]
        [Sec.\ref{Hybrid Model} Hybrid Model, bnode]
        [Sec.\ref{Video Generation} Video Generation, bnode]
    ]
    [Appendix~\ref{Evaluation} Evaluation, rnode
        [\ref{Human Evaluation} Human Evaluation, bnode]
        [\ref{Automatic Evaluation} Automatic Evaluation, bnode
            [\ref{Datasets} Datasets, onode
                [MS-COCO (COCO2014) \& COCO2017 \cite{lin2014microsoft} COCO-Stuff \cite{caesar2018coco} \\
                Conceptual Captions (CC3M) \cite{sharma2018conceptual} CC12M \cite{changpinyo2021conceptual} \\
                Open Images \cite{OpenImages2, OpenImagesSegmentation, OpenImages} 
                Localized Narrative (LN) \cite{PontTuset_eccv2020} \\
                LAION-5B \cite{schuhmann2022laion} 
                COYO-700M \cite{kakaobrain2022coyo-700m} \\
                T2I-CompBench \& T2I-CompBench++ \cite{huang2023t2i}, ppnode]
            ]
            [\ref{Evaluation Metrics} Metrics, onode
                [MSE \& PSNR \& SSIM \cite{wang2009mean, hore2010image, wang2004image} LPIPS \cite{zhang2018unreasonable}  \\
                Precision \& Recall \cite{lucic2018gans, sajjadi2018assessing, kynkaanniemi2019improved} \\
                R-Precision \& CLIP-R-Precision \cite{xu2018attngan, park2021benchmark} \\ SOA \cite{hinz2020semantic} IS \cite{salimans2016improved} FID \cite{heusel2017gans} sFID \cite{nash2021generating} KID \cite{binkowski2018demystifying} DINO \cite{caron2021emerging, oquab2023dinov2} \\
                CLIP-I \& CLIPScore (CLIP-T) \& RefCLIPScore \cite{hessel2021clipscore} \\
                LAION Aesthetics Score \cite{Romain2022laion, Romain2023laion} HPS \cite{wu2023human} HPSv2 \cite{wu2023humanv2}, ppnode]
            ]
        ]    
    ]
    [Appendix~\ref{Comparison} Comparison, rnode
        [Table~\ref{tab:mod} Comparison of Models, bnode]
        [Table~\ref{tab:perf} Comparison of Performance, bnode]
    ]
]
\end{forest}

\caption{Architecture of this survey.\protect}
\label{fig:arch}
\end{figure}

\section{Background} \label{Background}
In this section, we give a short introduction of the foundation model (AR, NAR, GAN and Diffusion) and some key strategies (Autoencoder, Attention and Class-free Guidance) used in T2I. 

\subsection{Foundation Models} \label{Foundation Models}

T2I model is a class of conditional generative model. Modern machine learning methods have studied it since \cite{mansimov2015generating}. They proposed that when the generative model (Draw \cite{gregor2015draw}) is conditioned on image captions, it can also generate visual scenes. Since then, the era of T2I generation has begun - numerous text-guided image generation models have emerged. The backbone on which these models rely can be divided into four categories: Autoregression-based, GAN-based, Diffusion-based, and Non-Autoregression-based. In this section, we will systematically introduce these backbones, including their architecture and execution.

\subsubsection{Autoregression} \label{2Autoregression} 
\leavevmode \\
Autoregression (AR) models are a class of likelihood-based directed probabilistic models used to model natural language before being used for image generation. Since language has a natural sequence order, the joint probability factor of token is usually decomposed into the product of conditional probabilities through the chain rule \cite{jelinek1980interpolated, bengio2000neural}:
\begin{equation}
    p(x) = \prod\limits_{i = 1}^n {p(\mathop s\nolimits_n |\mathop s\nolimits_1 , \ldots ,\mathop s\nolimits_{n - 1} )} 
\end{equation}
where, $x$ is a given sample, $\{s_1, \ldots, s_n\}$ corresponds to a series of tokens of $x$. The prediction of joint probability is a one-way process, and only the distribution of all tokens before the token can be used to predict the distribution of this token.

Transformers \cite{vaswani2017attention}, initially proposed for sequence-to-sequence tasks, promotes the further development of AR. Motivated by advances of GPT1 \cite{radford2018improving} and GPT2 \cite{radford2019language}, which is based on Transformers, in modeling natural language, AR is successfully applied to image generation (iGPT \cite{chen2020generative}): it downscales the resolution of the original image and then reshape the image into 1D sequences, these sequences are treated as discrete tokens. The model then autoregressively predicts the next pixel or predicts the masked pixel. Similarly, Vision Transformer (ViT) \cite{dosovitskiy2020image} attempts to apply Transformers directly to images. It splits the image into patches and uses linear projection to map these patches into embedding sequences as input to the Transformer.

\subsubsection{Non-Autoregression (NAR)} \label{2NAR}
\leavevmode \\
Directed sequential generation of autoregression models is a very time-consuming generation method. Thus, can we use a parallel generation method to predict the output of all positions at once, thereby reducing the inference latency? To this end, non-autoregression generation emerges. The concept of non-autoregression generation was first introduced by \cite{gu2017non} for natural language processing (NLP), which reduces the inference latency by an order of magnitude. 

 \cite{ghazvininejad2019mask} introduces the conditional masking language model (CMLM), using a mask-predict decoding algorithm, conditioned on text and a partially masked target translation. This is the basis of non-autoregression image generation \cite{chang2022maskgit, ding2022cogview2, chang2023muse, patil2024amused, feng2023emage}, which allows efficient iterative decoding: it first non-autoregressively predicts all target embeddings, and then iteratively mask and regenerate the subset of embeddings for which the model is least confident.

\subsubsection{GAN} \label{2GAN}
\leavevmode \\
GANs \cite{creswell2018generative} are generative models composed of a generator G and a discriminator D. The generator G maps latent variable $z$ to data distribution $p_{data}$ classified as true by D, where $z$ is a noise vector drawn from, e.g. a Gaussian or uniform distribution. The discriminator D is trained to distinguish whether the input is synthesized by G or sampled from real data distribution.

A large amount of work has focused on designing adversarial objectives to improve training, and the two notable objective formulations are minimax game \cite{creswell2018generative} and hinge loss \cite{miyato2018spectral}, as shown in Equation (\ref{minimax game}) and Equation (\ref{hinge loss}) respectively:
\begin{equation}\label{minimax game}
    \min\limits_{G}{\max\limits_{D}{V\left( {D,G} \right) = E_{x\sim p_{data}{(x)}}\left\lbrack {\log{D(x)}} \right\rbrack + E_{z\sim p{(z)}}\left\lbrack {\log\left( {1 - D\left( {G(z)} \right)} \right)} \right\rbrack}}
\end{equation}

\begin{equation}\label{hinge loss}
\begin{aligned}
    L_{D} &= E_{x\sim p_{data}{(x)}}\left\lbrack {\max\left( {0,1 - D(x)} \right)} \right\rbrack + E_{z\sim p{(z)}}\left\lbrack {\max\left( {0,1 + D\left( {G(z)} \right)} \right)} \right\rbrack \\
    L_{G} &= - E_{Z\sim P{(Z)}}\left( {D\left( {G(z)} \right)} \right)
\end{aligned}
\end{equation}
During training, it is common to alternately freeze one of G and D and use adversarial objective to train the other.

GAN-INT-CLS \cite{reed2016generative} is the first to introduce text conditioning into GAN generation, which trains a deep convolutional generative adversarial network conditioned on text feature encoded by a hybrid character-level convolutional recurrent neural network. Both the generator network G and the discriminator network D perform forward inference conditioned on text feature.

\subsubsection{Diffusion} \label{2Diffusion}
\leavevmode \\
Diffusion models, such as DDPM \cite{ho2020denoising, rombach2022high}, DDIM \cite{song2020denoising}, SDE-based models \cite{sarkka2019applied, song2020score, zhou2023denoising}, ODE-based models \cite{song2020score, liu2022flow}, etc., are a class of generative models. Its training usually consists of two stages: 1) The forward diffusion process gradually adds noise to the image with time steps until the image becomes completely Gaussian noise; 2) The backward denoising process is the reverse of the forward diffusion process, which gradually subtracts noise from the noisy image to reconstruct the original image. The noise subtracted at each time step is estimated by the network (usually based on the U-Net \cite{ronneberger2015u} architecture). Note that in addition to the commonly used prediction noise ($\epsilon$-prediction), there are other prediction methods, such as $x_0$-prediction \cite{ho2020denoising} and v-prediction \cite{salimans2022progressive}.The differences and connections between different diffusion models are as follows:

The diffusion process of DDPM \cite{ho2020denoising} is usually a Markov chain, which requires a larger time step to have good performance and longer denoising time. Compared with DDPM in high-dimensional pixel space, LDM \cite{rombach2022high} adds a pre-trained encoder that encodes images into a low-dimensional latent space and a pre-trained decoder that decodes the latent back to the image in the outer layer of DDPM. The diffusion and denoising process are performed in a low-dimensional latent space, which speeds up the entire process while reducing computational resource requirements.

The diffusion process of DDIM \cite{song2020denoising} is a non-Markov chain, so the inference time can be shortened by jumping time steps. It differs from DDPM mainly in that its denoising process is deterministic, i.e., no random noise is introduced during denoising process. On the other hand, in ideal case, i.e. , the noise variance of diffusion process is large enough, the end point $x_1$ will be almost independent of the starting point. Therefore, the distribution of the ideal DDPM output (the end point of the denoising process) is completely independent of the starting point $x_1$, and it implements a random mapping. In contrast, DDIM implements a deterministic mapping: for a given $x_1$ it will always produce a fixed value, and therefore is very dependent on $x_1$ \cite{nakkiran2024step}.

DDPM is a discrete time process. In the continuous limit, when ${\Delta}t\rightarrow0$, the discrete diffusion process becomes a stochastic differential equation \cite{sarkka2019applied}. SDE \cite{song2020score} is a class of diffusion models that combines a score-based model (NCSC \cite{song2019generative}) with a stochastic differential equation. In the backward denoising process, it uses a time-dependent score function (similar to the noise estimator used in DDPM) to solve the inverse stochastic differential equation.

Denoising Diffusion Bridge Models (DDBM) \cite{zhou2023denoising} points out that the limitation of diffusion models is that they can only transform complex data distributions to standard Gaussian distributions and cannot naturally transform between two arbitrary distributions. Fortunately, the diffusion process can be placed on fixed endpoints through the famous Doob's h transformation \cite{doob1984classical}. When the starting point is fixed, the process is often called Diffusion Bridge \cite{sarkka2019applied}.

Similar to the relationship between DDPM and SDE, DDIM \cite{song2020denoising} can be extended to probability flow ordinary differential equations (PF-ODE) \cite{song2020score}. In addition, \cite{song2020score} also shows that stochastic differential equation \cite{sarkka2019applied} can be equivalently converted to PF-ODE. Rectified Flow \cite{liu2022flow} is purely based on ODE \cite{song2020score}. It uses reflow to straighten the trajectory of probability flow to accelerate the inference process.

\textbf{Analysis of diffusion model}. In the denoising process of DDPM , non-deterministic noise (variance is not 0) is added. This does not seem intuitive enough, we can understand it from the perspective of vector fields and particles: 1) We know that in a vector field, particles will move according to the direction of the vector field. 2) Particles are doing Brownian motion all the time. These two points correspond to the two terms in the SDE \cite{song2020score} formula. Now let's go back to DDPM. Recall that DDPM is a discretized special case of SDE, so it can be understood in a similar way: the vector field corresponds to the denoising process from the Gaussian distribution to the ground truth distribution, and the Brownian motion corresponds to the non-deterministic noise added during the denoising process. Similarly, DDIM and ODE can be understood in the same way.

Note that many distillation methods aim to train a one-step generator student model to match the output of the diffusion teacher model, thereby achieving high-quality sampling in one step (or a few steps), such as consistency models \cite{song2023consistency, luo2023latent} and adversarial distillation methods \cite{lin2024sdxl, xu2024ufogen, sauer2024fast}. However, distillation models are no longer diffusion models, and their samplers (even multi-step ones) are not diffusion samplers \cite{nakkiran2024step}.

T2I diffusion models are conditional image generation models. Text prompts are typically encoded by a text encoder into text embeddings, which are then used together with time-step embeddings to guide image denoising and reconstruction in backward denoising process. The most common way to use text conditions in diffusion models is classifier-free guidance \cite{ho2022classifier}, which is a form of guidance that interpolates between the predictions of conditional and unconditional diffusion models, see Section~\ref{CFG} for details.

In recent years, the number of diffusion-based T2I generation models have far exceeded GAN-based methods. This is because compared to GAN training which is prone to mode collapse \cite{salimans2016improved}, the training of the diffusion model is more stable.

\subsection{Key Strategies} \label{Key Strategies}
\subsubsection{Autoencoder} \label{Autoencoder}
\leavevmode \\
Autoencoder (AE) is an encoder-decoder architecture. The encoder encodes the input data into representations, called embeddings, of a low-dimensional latent space; the decoder receiving embeddings reconstructs input data. By training the AE to copy the input to the output, the latent representation can learn useful properties. 

To a certain extent, existing T2I generation models, such as AR, GAN, Diffusion, NAR, etc., can be regarded as using AE architecture, except for models that directly operate on pixels. They use text to guide image generation or image editing by operating in the latent space.

AE has many derivative works and variational autoencoders (VAE \cite{kingma2013auto, kingma2019introduction}) are the first that can be called "generative models". On the encoder side, VAE utilizes KL divergence between an additional distribution and the input distribution to approximate the input distribution; while the decoder uses this approximated distribution to sample the output. This approach breaks the data flow from input to output, thus preventing the input from being copied directly to the output.

Inspired by vector quantization (VQ) \cite{gray1984vector}, the vector quantized variational autoencoder (VQ-VAE) \cite{van2017neural} encodes input images into discrete latent representations based on a codebook consisting of embeddings drawn from a distribution. VQ-VAE-2 \cite{razavi2019generating} trains a hierarchical VQ-VAE to encode images into discrete latent spaces. VQ-GAN \cite{esser2021taming} introduces perceptual loss and adversarial training of patch-based discriminator to VQ-VAE \cite{van2017neural} to learn a rich codebook while pushing the compression limit.

In addition to the above that can be used as image encoders, existing image encoders also include dVAE \cite{vahdat2018dvae}, VQ-diffusion \cite{gu2022vector}, RQ-VAE \cite{lee2022autoregressive}, Transformer-VQ \cite{lingle2023transformer}, VQ-SEG and VQ-IMG \cite{gafni2024scene}, ViT-VQGAN \cite{yu2021vector}, etc., and image encoder of text-image pre-training model (such as CLIP \cite{radford2021learning}).

In the T2I generation model, in order to use text-guided generation, one need to first use a tokenizer (such as BPE \cite{sennrich2015neural}, WordPiece \cite{devlin2018bert}, SentencePiece \cite{kudo2018sentencepiece}, UnigramLM \cite{kudo2018subword}, etc.) to convert text data into text tokens, and then use text encoder to encodes text tokens into latent representations. Existing text encoders include: BERT \cite{devlin2018bert}, T5 \cite{raffel2020exploring, chung2024scaling}, and the text encoder of text-image pre-training model (such as CLIP \cite{radford2021learning}) .

\subsubsection{Attention} \label{Attention}
\leavevmode \\
Attention mechanisms can be used to model the dependency of (visual/textual) tokens regardles their distance in sequences \cite{bahdanau2014neural, kim2017structured}. Self-attention is an attention mechanism that models the correlation of different positions in a single sequence, and Transformer \cite{vaswani2017attention} is the first transduction model that entirely relies on self-attention to compute the representation of input and output. In the field of T2I generation, DALL-E \cite{ramesh2021zero} uses three attentions ("Text attends to Text", "Image attends to Text" and "Image attends to Image") to improve performance of using a single attention, and text-image alignment is improved through cross-attention.

\subsubsection{Class-free Guidance} \label{CFG}
\leavevmode \\
The implementation of Classifier-Free Guidance (CFG) \cite{ho2022classifier} refers to jointly training conditional and unconditional diffusion models at training time and combining with conditional and unconditional score estimates $\varepsilon$ at inference time. More specifically, the unconditional and conditional models are jointly trained by randomly setting the condition c to the unconditional class identifier $\varnothing$ with probability $p$ during training. At inference time, sampling is performed using a linear combination of conditional and unconditional score estimates with a guidance scale $\omega \geq 1$. The modified score estimate therefore extrapolates in the direction toward the conditional and away from the unconditional.
\begin{equation}
\begin{aligned}
    {\overset{\sim}{\varepsilon}}_{\theta}\left( {z,c_{I},c_{T}} \right) &= \varepsilon_{\theta}\left( {z,\varnothing,\varnothing} \right)~ \\
    &+ {~\omega}_{I}\left( {\varepsilon_{\theta}\left( {z,c_{I},\varnothing} \right) - \varepsilon_{\theta}\left( {z,\varnothing,\varnothing} \right)} \right) \\
    &+ ~\omega_{T}\left( {\varepsilon_{\theta}\left( {z,c_{I},c_{T}} \right) - \varepsilon_{\theta}\left( {z,c_{I},\varnothing} \right)} \right)
\end{aligned}
\end{equation}

\section{Image Generation} \label{Image Generation}

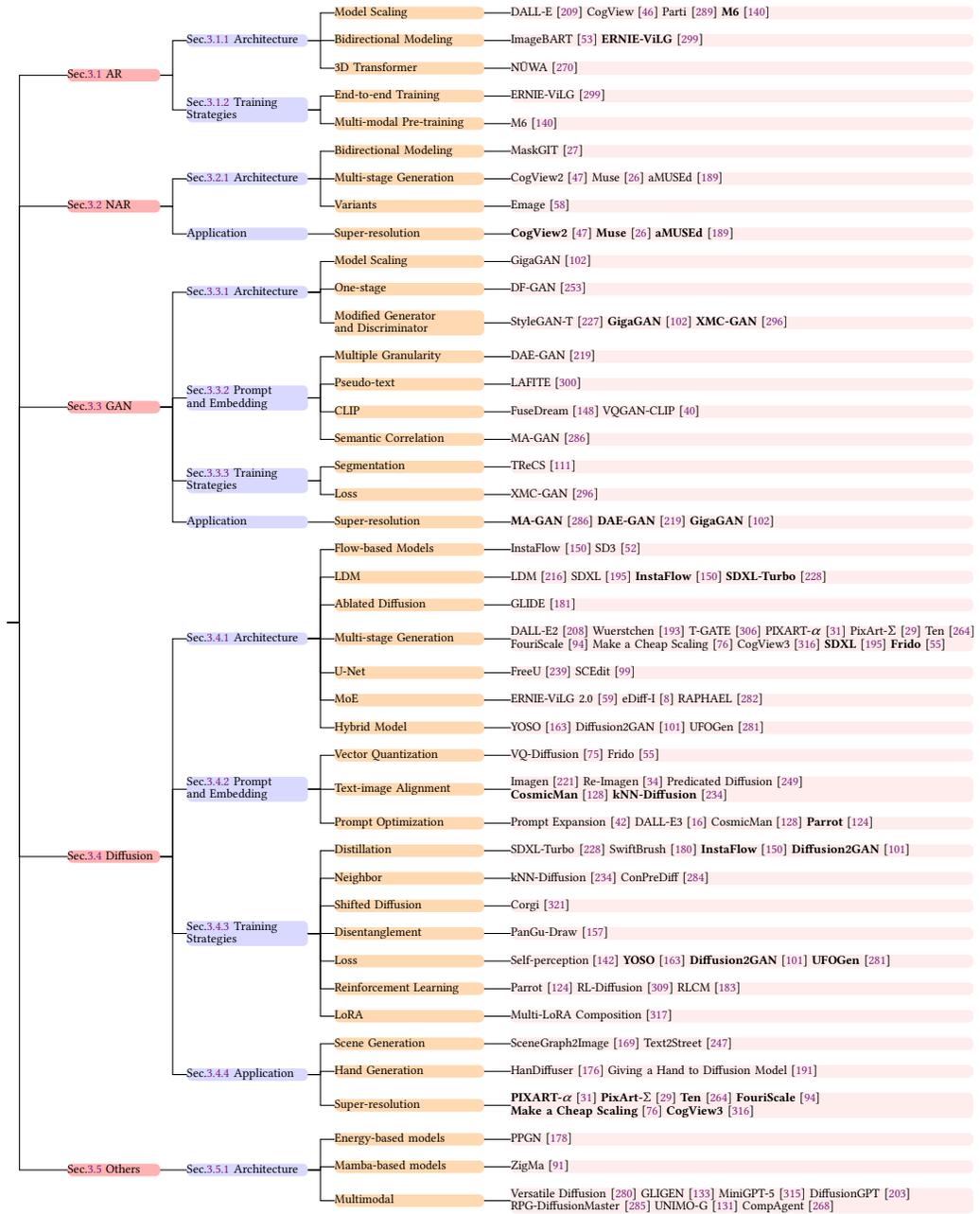
\begin{figure} [!ht]
\centering
\tikzset{
    rnode/.style = {thin, rounded corners=2pt, align=left, fill=red!30, text width=1.3cm},
    bnode/.style = {thin, rounded corners=2pt, align=left, fill=blue!15, text width=1.7cm},
    onode/.style = {thin, rounded corners=2pt, align=left, fill=orange!30, text width=2.1cm},
    pnode/.style = {thin, rounded corners=2pt, align=left, fill=pink!30, text width=6.5cm},
}

\begin{forest}
for tree={
    inner sep=0pt,
    outer sep=0pt,
    fit=band,
    child anchor=west,
    parent anchor=east,
    grow'=0,
    anchor=west,
    align=left,
    edge path={\noexpand\path[\forestoption{edge}] 
         (!u.parent anchor) -- +(5pt,0pt) |- (.child anchor)
         \forestoption{edge label};},
    font=\fontsize{5}{5}\selectfont
}   
[
    [Sec.\ref{3Autoregression} AR, rnode
        [Sec.\ref{3Autoregression Architecture} Architecture, bnode
            [Model Scaling, onode
                [DALL-E~\cite{ramesh2021zero} CogView~\cite{ding2021cogview}  Parti~\cite{yu2022scaling} \textbf{M6}~\cite{lin2021m6}, pnode]
            ]
            [Bidirectional Modeling, onode
                [ImageBART~\cite{esser2021imagebart} \textbf{ERNIE-ViLG}~\cite{zhang2021ernie}, pnode]
            ]
            [3D Transformer, onode
                [NÜWA~\cite{wu2022nuwa}, pnode]
            ]
        ]
        [Sec.\ref{3Autoregression Training Strategies} Training \\
        Strategies, bnode
            [End-to-end Training, onode
                [ERNIE-ViLG~\cite{zhang2021ernie}, pnode]
            ]
            [Multi-modal Pre-training, onode
                [M6~\cite{lin2021m6}, pnode]
            ]
        ]
    ]
    [Sec.\ref{3NAR} NAR, rnode
        [Sec.\ref{3NAR Architecture} Architecture, bnode
            [Bidirectional Modeling, onode
                [MaskGIT~\cite{chang2022maskgit}, pnode]
            ]
            [Multi-stage Generation, onode
                [CogView2~\cite{ding2022cogview2} Muse~\cite{chang2023muse} aMUSEd~\cite{patil2024amused}, pnode]
            ]
            [Variants, onode
                [Emage~\cite{feng2023emage}, pnode]
            ]
        ]
        [Application, bnode
            [Super-resolution, onode
                [\textbf{CogView2}~\cite{ding2022cogview2} \textbf{Muse}~\cite{chang2023muse} \textbf{aMUSEd}~\cite{patil2024amused}, pnode]
            ]
        ]
    ]
    [Sec.\ref{3GAN} GAN, rnode
        [Sec.\ref{3GAN Architecture} Architecture, bnode
            [Model Scaling, onode
                [GigaGAN~\cite{kang2023scaling}, pnode]
            ]
            [One-stage, onode
                [DF-GAN~\cite{tao2022df}, pnode]
            ]
            [Modified Generator \\
            and Discriminator, onode
                [StyleGAN-T~\cite{sauer2023stylegan} \textbf{GigaGAN}~\cite{kang2023scaling} \textbf{XMC-GAN}~\cite{zhang2021cross}, pnode]
            ]
        ]
        [Sec.\ref{3GAN Prompt and Embedding} Prompt \\
        and Embedding, bnode
            [Multiple Granularity, onode
                [DAE-GAN~\cite{ruan2021dae}, pnode]
            ]
            [Pseudo-text, onode
                [LAFITE~\cite{zhang2021lafite}, pnode]
            ]
            [CLIP, onode
                [FuseDream~\cite{liu2021fusedream} VQGAN-CLIP~\cite{crowson2022vqgan}, pnode]
            ]
            [Semantic Correlation, onode
                [MA-GAN~\cite{yang2021multi}, pnode]
            ]
        ]       
        [Sec.\ref{3GAN Training Strategies} Training \\
        Strategies, bnode
            [Segmentation, onode
                [TReCS~\cite{koh2021text}, pnode]
            ]
            [Loss, onode
                [XMC-GAN~\cite{zhang2021cross}, pnode]
            ]
        ]
        [Application, bnode
            [Super-resolution, onode
                [\textbf{MA-GAN}~\cite{yang2021multi} \textbf{DAE-GAN}~\cite{ruan2021dae} \textbf{GigaGAN}~\cite{kang2023scaling}, pnode]
            ]
        ]
    ]
    [Sec.\ref{3Diffusion} Diffusion, rnode
        [Sec.\ref{3Diffusion Architecture} Architecture, bnode
            [Flow-based Models, onode
                [InstaFlow~\cite{liu2023instaflow} SD3~\cite{esser2024scaling}, pnode]
            ]
            [LDM, onode
                [LDM~\cite{rombach2022high} SDXL~\cite{podell2023sdxl} \textbf{InstaFlow}~\cite{liu2023instaflow} \textbf{SDXL-Turbo}~\cite{sauer2023adversarial}, pnode]
            ]
            [Ablated Diffusion, onode
                [GLIDE~\cite{nichol2021glide}, pnode]
            ]
            [Multi-stage Generation, onode
                [DALL-E2~\cite{ramesh2022hierarchical} Wuerstchen~\cite{pernias2023wurstchen} T-GATE~\cite{zhang2024cross} PIXART-$\alpha$~\cite{chen2023pixart}
                PixArt-$\Sigma$~\cite{chen2024pixart} Ten~\cite{wang2024generative} \\ FouriScale~\cite{huang2024fouriscale} Make a Cheap Scaling~\cite{guo2024make}  
                CogView3~\cite{zheng2024cogview3} \textbf{SDXL}~\cite{podell2023sdxl} \textbf{Frido}~\cite{fan2023frido}, pnode] 
            ]
            [U-Net, onode
                [FreeU~\cite{si2024freeu} SCEdit~\cite{jiang2024scedit}, pnode]
            ]
            [MoE, onode
                [ERNIE-ViLG 2.0~\cite{feng2023ernie} eDiff-I~\cite{balaji2022ediff} RAPHAEL~\cite{xue2024raphael}, pnode]
            ]
            [Hybrid Model, onode
                [YOSO~\cite{luo2024you} Diffusion2GAN~\cite{kang2024distilling} UFOGen~\cite{xu2024ufogen}, pnode]
            ]
        ]
        [Sec.\ref{3Diffusion Prompt and Embedding} Prompt \\
        and Embedding, bnode
            [Vector Quantization, onode
                [VQ-Diffusion~\cite{gu2022vector} Frido~\cite{fan2023frido}, pnode]
            ]
            [Text-image Alignment, onode
                [Imagen~\cite{saharia2022photorealistic} Re-Imagen~\cite{chen2022re} Predicated Diffusion~\cite{sueyoshi2024predicated} \\ \textbf{CosmicMan}~\cite{li2024cosmicman} \textbf{kNN-Diffusion}~\cite{sheynin2022knn}, pnode]
            ]
            [Prompt Optimization, onode
                [Prompt Expansion~\cite{datta2023prompt} DALL-E3~\cite{betker2023improving} CosmicMan~\cite{li2024cosmicman} \textbf{Parrot}~\cite{lee2024parrot}, pnode]
            ]
        ]
        [Sec.\ref{3Diffusion Training Strategies} Training \\
        Strategies, bnode
            [Distillation, onode
                [SDXL-Turbo~\cite{sauer2023adversarial} SwiftBrush~\cite{nguyen2024swiftbrush} 
                \textbf{InstaFlow}~\cite{liu2023instaflow} \textbf{Diffusion2GAN}~\cite{kang2024distilling}, pnode]
            ]
            [Neighbor, onode
                [kNN-Diffusion~\cite{sheynin2022knn} ConPreDiff~\cite{yang2024improving}, pnode]
            ]
            [Shifted Diffusion, onode
                [Corgi~\cite{zhou2023shifted}, pnode]
            ]
            [Disentanglement, onode
                [PanGu-Draw~\cite{lu2023pangu}, pnode]
            ]
            [Loss, onode
                [Self-perception~\cite{lin2023diffusion} \textbf{YOSO}~\cite{luo2024you} \textbf{Diffusion2GAN}~\cite{kang2024distilling} \textbf{UFOGen}~\cite{xu2024ufogen}, pnode]
            ]
            [Reinforcement Learning, onode
                [Parrot~\cite{lee2024parrot} RL-Diffusion~\cite{zhang2024large} 
                RLCM~\cite{oertell2024rl}, pnode]
            ]
            [LoRA, onode
                [Multi-LoRA Composition~\cite{zhong2024multi}, pnode]
            ]
        ]
        [Sec.\ref{3Diffusion Application} Application, bnode
            [Scene Generation, onode
                [SceneGraph2Image~\cite{mishra2024scene} Text2Street~\cite{su2024text2street}, pnode]
            ]
            [Hand Generation, onode
                [HanDiffuser~\cite{narasimhaswamy2024handiffuser} 
                Giving a Hand to Diffusion Model~\cite{pelykh2024giving}, pnode]
            ]
            [Super-resolution, onode
                [\textbf{PIXART-$\alpha$}~\cite{chen2023pixart} \textbf{PixArt-$\Sigma$}~\cite{chen2024pixart} \textbf{Ten}~\cite{wang2024generative} \textbf{FouriScale}~\cite{huang2024fouriscale} \\
                \textbf{Make a Cheap Scaling}~\cite{guo2024make} \textbf{CogView3}~\cite{zheng2024cogview3}, pnode]
            ]
        ]
    ]
    [Sec.\ref{Others} Others, rnode
        [Sec.\ref{Others Architecture} Architecture, bnode
            [Energy-based models, onode
                [PPGN~\cite{nguyen2017plug}, pnode]
            ]
            [Mamba-based models, onode
                [ZigMa~\cite{hu2024zigma}, pnode]
            ]
            [Multimodal, onode
                [Versatile Diffusion~\cite{xu2023versatile} GLIGEN~\cite{li2023gligen} MiniGPT-5~\cite{zheng2023minigpt} DiffusionGPT~\cite{qin2024diffusiongpt} \\ 
                RPG-DiffusionMaster~\cite{yang2024mastering} UNIMO-G~\cite{li2024unimo} CompAgent~\cite{wang2024divide}, pnode]
            ]
        ]    
    ]
]
\end{forest}

\caption{Taxonomy of image generation. Papers are categorized according to its main contribution. Note that papers falling into other categories according to its sub-important contributions are shown in bold and the description of which are not given again for concise.\protect}
\label{fig:generation}
\end{figure}

In this section, we introduce the taxonomy of papers focusing on text-guided image generation. Based on the methods used by the papers of each foundation model, we categorize the papers from the perspective of global (architecture) to local, from front (prompts and embeddings) to back (training strategies). In addition, we also focus on the specific applications of T2I models. Note that an paper may be categorized into multiple categories due to its multiple contributions. In order to avoid redundancy, we will only introduce papers in one category (based on its main contribution), as shown in Figure~\ref{fig:generation}. The same paper in other categories is indicated by bold in Figure~\ref{fig:generation}. In summary, we investigate the papers of each base foundation from the following four perspectives (if applicable):

\begin{itemize}
\item \textbf{Architecture}: an global perspective involving new architectures (e.g. flow-based model, LDM, Hybrid Model, etc.), changes to existing architectures (e.g. Model Scaling, improved U-Net, etc.), and modeling methods by modifying the architecture (e.g. single-step/multi-step generation, bidirectional modeling, etc.).

\item \textbf{Prompt and Embedding}: a local perspective, referring to techniques from input pre-processing (e.g., Pseudo-text/Inversion, Prompt Optimization, etc.) to encoding/embedding techniques (e.g., vector quantization) and subsequent operations in embedding space (e.g., CLIP-based techniques, techniques to improve image-text alignment, etc.).

\item \textbf{Training Strategies}: a local perspective, referring to training strategies to improve performance, such as fine-tuning, MoE, reinforcement learning, etc.

\item \textbf{Application}: referring to specialized applications of T2I models, such as text-guided scene generation, hand generation, and super-resolution.

\end{itemize} 

\subsection{Autoregression} \label{3Autoregression}
\subsubsection{Architecture} \label{3Autoregression Architecture}
\leavevmode \\
\textbf{Model Scaling}. Early visual Autoregression (AR) models, such as iGPT \cite{chen2020generative}, PixelCNN \cite{van2016conditional}, PixelRNN \cite{van2016pixel} and Image Transformer \cite{parmar2018image}, usually have smaller data set sizes and model sizes. To explore whether dataset size and model size are limiting factors in model generation capabilities, DALL-E \cite{ramesh2021zero} trains a transformer \cite{vaswani2017attention} with 12B parameters on 250M text-image pairs, allowing for flexible, high-fidelity generation. A concurrent work, CogView \cite{ding2021cogview}, collects 30M high-quality text-image pairs and train a transformer with 4B parameters, and fine-tune it to adapt to various downstream tasks. In addition, CogView uses Precision Bottleneck Relaxation (PB-Relax) and Sandwich LayerNorm (Sandwich-LN) to stabilize the training of large Transformer on complex datasets. Similarly, Parti \cite{yu2022scaling} extends the parameters of transformer to 20B to achieve high-fidelity photo-realistic image generation and content-rich synthesis. In addition, to obtain high-quality results corresponding to complex prompts, Parti also proposed the concept of "growing the cherry tree"——one develops a prompt in increments while interacting with the model, in order to produce a cherry tree that offers some great outputs.

\textbf{Bidirectional Modeling}. Unidirectional and fixed ordering of sequence elements imposes a perceptually unnatural bias on attention in images by only considering contextual information to the left or above. To solve this problem, ImageBART \cite{esser2021imagebart} successively introduces bidirectional context by inverting the polynomial diffusion process in a compact image representation space, thereby not only alleviating the exposure bias often encountered by unidirectional AR models, but also enabling the use of AR models for high-fidelity images synthesis.

\textbf{3D Transformer}.NÜWA \cite{wu2022nuwa} designed a 3D transformer encoder-decoder framework that can process not only 1D text and 2D image data, but also 3D video data. It also proposes a 3D Nearby Attention (3DNA) that considers visual data continuity to improve generation quality while reducing computational complexity.

\subsubsection{Training Strategies} \label{3Autoregression Training Strategies}
\leavevmode \\
\textbf{End-to-end training}. Different from the traditional two-stage method, ERNIE-ViLG \cite{zhang2021ernie} proposes an end-to-end training method to simultaneously train the visual sequence generator and image reconstructor and adjust the sparse attention mechanism for unidirectional modeling to perform bidirectional modeling. This bidirectional generative modeling simplifies semantic alignment between vision and language.

\textbf{Multi-modal Pre-training}. M6 \cite{lin2021m6} uses a transformer architecture based on self-attention to perform multi-modal pre-training on a large-scale Chinese dataset. This is the first study to combine T2I generation with pre-training. It also incorporates Mixture-of-Experts (MoE) to extend model parameters from 10B to 100B.

\subsection{Non-Autoregression (NAR)} \label{3NAR}
\subsubsection{Architecture} \label{3NAR Architecture}
\leavevmode \\
\textbf{Bidirectional Modeling}. The generative Transformer treats the image simply as a sequence of tokens and decodes the image sequentially in raster scan order (i.e. row-by-row). This strategy is neither optimal nor efficient. MaskGIT \cite{chang2022maskgit} uses a bidirectional Transformer decoder: during training, it learns to predict randomly masked tokens by attending tokens in all directions; during inference, the model first generates all tokens of the image simultaneously, and then iterates based on the previous generation to refine the image. MaskGIT analyzed the performance of different masking scheduling functions (linear function, concave function, convex function) and found that the cosine function (concave function) performed best.

\textbf{Multi-stage Generation}. CogView2 \cite{ding2022cogview2} proposes a cross-modal universal language model (CogLM) that concatenates tokenized text and image as input, then masks various types of tokens in the text and image token sequences and learns to predict them. For low-resolution tokens generated by CogLM, they are refined using direct super-resolution and iterative super-resolution modules fine-tuned from pre-trained CogLM, respectively, and then decoded to obtain high-resolution images. In iterative super-resolution modules, most tokens are re-masked and regenerated in a local parallel autoregressive (LoPAR) manner. Similarly, Muse \cite{chang2023muse} uses a masked Transformer to generate low-resolution latent, and then trains two VQ-GAN \cite{esser2021taming} for hierarchical super-resolution. To further improve the model's ability to generate fine details, Muse increased the capacity of the VQGAN decoder, adding more residual layers and channels while keeping the capacity of the encoder frozen. Then, fine-tuned the new decoder layer while keeping the weights, codebook and transformer of the VQGAN encoder frozen. aMUSEd \cite{patil2024amused} is an open source, lightweight masking image model based on MUSE. It enables fast image generation using only 10\% of the parameters of MUSE.

\textbf{Variants}. Based on the fully non-autoregressive model, Emage \cite{feng2023emage} developed three stepwise improved variants (iterative non-autoregressive model): it uses T iterations to generate (such as 1024) image tokens. In each iteration, 1024/T image tokens are generated simultaneously, conditioning on image tokens generated in all previous iterations. Finally, the variant 3, which can correct the prediction errors of previous iterations, achieves the best performance.

\subsection{GAN} \label{3GAN}
After GAN-INT-CLS \cite{reed2016generative}, a large number of GAN-based T2I generation models have emerged \cite{xu2018attngan, zhang2017stackgan, zhang2018stackgan++, li2019object, donahue2017semantically, zhu2019dm, qiao2019mirrorgan, li2019controllable, hinz2020semantic, liu2020cp}. In recent years, there are also numerous related works. We will show them as follows.

\subsubsection{Architecture} \label{3GAN Architecture}
\leavevmode \\
\textbf{Model Scaling}. Just as proposed in Section~\ref{3Autoregression} that extending the autoregressive model to improve generation quality, GigaGAN \cite{kang2023scaling} performs similar operations. GigaGAN is larger and has better performance compared with previous models \cite{karras2020analyzing, sauer2022stylegan}. It modifies the generator and discriminator of StyleGAN2 \cite{karras2020analyzing} to generate low-resolution images and then uses a GAN-based up-sampler for super-resolution.

\textbf{One-stage Generation}. In order to solve the entanglement problem between generators of different image scales introduced by the stacked architecture, DF-GAN \cite{tao2022df} replaces the stacked backbone with a one-stage backbone. Furthermore, it uses an object-aware discriminator, consisting of match-aware gradient penalty and one way output, and a deep image text fusion block (DFBlock) to improve text-image alignment. Compared with using cross-attention, using DFBlock composed of multiple affine transformations \cite{perez2018film} has lower computational cost and does not need to consider the limitations from image scales.

\textbf{Modified Generator and Discriminator}. StyleGAN-T \cite{sauer2023stylegan} uses StyleGAN-XL \cite{sauer2022stylegan} as the baseline architecture, and improves performance by modifying the generator and discriminator. In addition, it improves text-image alignment through the guidance of CLIP \cite{radford2021learning}, and a balance can be achieved between image quality, text conditioning, and distribution diversity by adjusting the weight in overall CLIP loss.

\subsubsection{Prompt and Embedding} \label{3GAN Prompt and Embedding}
\leavevmode \\
\textbf{Multiple Granularity}. DAE-GAN \cite{ruan2021dae} uses "aspect" information in text for T2I generation to improve the fidelity and semantic consistency of images. It first extracts text semantic representation from multiple granularity, namely sentence level, word level, and aspect level, and then generates low-resolution images using sentence-level text features and random noise vectors.

\textbf{Pseudo-text}. Training of previous T2I models usually requires a large number of high-quality text-image pairs carefully annotated by humans, which is very expensive. To solve this problem, LAFITE \cite{zhang2021lafite} proposes the first method to train T2I generation models without text data. It utilizes the image-text feature alignment properties of pre-trained CLIP model in the joint semantic space to construct <image, pseudo-text> feature pairs. In addition, a modified StyleGAN2 \cite{karras2020analyzing} architecture can be trained using <image, pseudo-text> feature pairs.

\textbf{CLIP}. An interesting CLIP-based T2I generation method is to find the image with the maximum semantic relevance to the given text under the CLIP metric \cite{hessel2021clipscore} in the latent space of the off-the-shelf GAN. However, optimizing CLIP scores in GAN space is extremely challenging. To solve this problem, FuseDream \cite{liu2021fusedream} proposed a more robust AugCLIP score and introduced novel optimization methods to improve the CLIP+GAN method. Unlike FuseDream, VQGAN-CLIP \cite{crowson2022vqgan} uses CLIP \cite{radford2021learning} to guide VQ-GAN \cite{esser2021taming} for T2I generation. It uses GAN to iteratively generate images based on text prompts, using CLIP to improve image-text alignment at each step. 

\textbf{Semantic Correlation}. Most previous works only focus on learning the semantic consistency between paired images and sentences without exploring the semantic correlation between different but related sentences describing the same image, which leads to significant visual differences between synthetic images. MA-GAN \cite{yang2021multi} is proposed to solve this problem. It consists of a single-sentence generator that generates high-resolution images using target sentences and a multi-sentence discriminator that identifies different sentences and learns their semantic relatedness. In addition, it uses a progressive negative sample selection mechanism (PNSS) to improve model performance by gradually introducing negative samples into training process.

\subsubsection{Training Strategies} \label{3GAN Training Strategies}
\leavevmode \\
\textbf{Segmentation}. Given the dataset Localized Narratives [] in which image descriptions are paired with mouse trajectories, TReCS \cite{koh2021text} 1) leverages the descriptions to retrieve segmentation masks in different images and predict object labels aligned with mouse trajectories, and then 2) places the masks according to the reverse order in which the corresponding labels appear in the narrative to create scene segmentation. Finally, images are generated using mask-to-image generation models (such as SPADE \cite{park2019semantic} and CC-FPSE \cite{liu2019learning}) that are adversarially trained through the GAN framework \cite{creswell2018generative}.

\textbf{Loss}. XMC-GAN \cite{zhang2021cross} generates realistic images aligned with text by maximizing the mutual information between images and text. It uses multiple contrastive losses to capture inter- and intra-modal correspondences. It consists of a one-stage attentional self-modulation generator and a contrast discriminator that acts as an encoder to generate global and regional image features for contrast loss in addition to functioning as a discriminator to distinguish true from false.

\subsection{Diffusion} \label{3Diffusion}
\subsubsection{Architecture} \label{3Diffusion Architecture}
\leavevmode \\
\textbf{Flow-based Models}. InstaFlow \cite{liu2023instaflow} is the first T2I generation model with single-step diffusion derived from stable diffusion \cite{rombach2022high}. Inspired by Rectified Flow \cite{liu2022flow}, it adopts reflow to straighten the trajectory of probability flow, refining the coupling between noise and images, and facilitating the distillation process of student models. SD3 [] improves noise sampling techniques for training Rectified Flow models, biasing them towards perceptually relevant scales. It also proposes a T2I structure based on DiT \cite{peebles2023scalable}, using independent weights for both modalities and achieving bidirectional information flow between image and text tokens.

\textbf{LDM}. Previous diffusion models often operated in pixel space, requiring substantial computational resources. To train on limited computational resources while maintaining quality and flexibility, LDM \cite{rombach2022high} applies diffusion models in the latent space of pre-trained encoders (such as VAE), conducting diffusion and denoising processes in a low-dimensional latent space. Additionally, it avoids arbitrary high-variance latent spaces through regularization. The recent noteworthy open-source Stable Diffusion (SD) is developed based on LDM. SDXL \cite{podell2023sdxl} is an improved LDM for T2I generation. Compared to SD, SDXL utilizes a three times larger UNet backbone, designs various novel adjustments, and trains on multiple aspect ratios. Additionally, it introduces a refinement model to enhance generation fidelity.

\textbf{Ablated Diffusion}. Ablated Diffusion Model (ADM) \cite{dhariwal2021diffusion} searches for a better architecture for unconditional image generation through a series of ablations. GLIDE \cite{nichol2021glide} represents a guided language for image diffusion used for generation and editing. It employs an ADM architecture and enhances it with text as conditioning information. It compares CLIP guidance and classifier-free guidance and finds that the latter performs better in terms of realism and caption similarity. The model can also achieve text-guided image editing through fine-tuning.

\textbf{Multi-stage Generation}. DALL-E2 \cite{ramesh2022hierarchical} proposes a two-stage T2I generation and editing model: the prior model generates CLIP image embeddings based on given text prompts, and the decoder generates images conditioned on CLIP image embeddings. When human evaluations are conducted on images generated using diffusion priors and autoregressive priors, humans prefer the former. Wuerstchen \cite{pernias2023wurstchen} is a diffusion model using a three-stage training and generation method. It learns a detailed but highly compact semantic image representation to guide the diffusion process. Compared to the latent representation of language, this highly compressed image representation provides more detailed guidance, significantly reducing computational requirements. T-GATE \cite{zhang2024cross} observes that the inference process can be divided into a text-guided semantic generation stage that relies on cross-attention and a fidelity-improving stage. Ignoring text in the latter stage reduces computational complexity while maintaining model performance. Based on this, T-GATE caches the cross-attention output after it converges and keeps it unchanged during the remaining inference steps. This method can be used as a plug-in for existing generative diffusion models without the need for retraining. To support high-resolution image generation with low training cost, PIXART-$\alpha$ \cite{chen2023pixart} injects text into DiT \cite{peebles2023scalable} using cross-attention and enhances generation performance through three different training steps. Additionally, it automatically generates high-information density pseudo-captions for images using the large-scale visual language model LLaVA \cite{liu2024visual}, thereby enhancing text-image alignment. PixArt-$\Sigma$ \cite{chen2024pixart} is an improved DiT capable of directly generating 4K resolution images. It develops from low-resolution to high-resolution models by integrating high-quality data. It uses key-value (KV) compression within the DiT framework to significantly improve efficiency and facilitate the generation of ultra-high-resolution images. Ten \cite{wang2024generative} proposes joint multi-scale diffusion sampling and scaling stack representation updated during diffusion-based sampling. It generates an entire set of scaled images corresponding to a scene in a scale-consistent manner using a pre-trained T2I diffusion model. This method is guided by different text prompts at each generated scale. FouriScale \cite{huang2024fouriscale} achieves structural and scale consistency for cross-resolution generation by integrating dilated convolutions and low-pass filtering to replace the original convolution layers in the pre-trained diffusion model. It enables the use of models pre-trained on low-resolution images for arbitrary size, high-resolution, and high-quality generation. Make a Cheap Scaling \cite{guo2024make} proposes a self-cascading diffusion model. The self-cascading diffusion model can efficiently adapt to higher resolutions by leveraging a trained low-resolution model and integrating a series of multi-scale upsampling modules. It uses a pivot-guided noise rescheduling strategy to speed up inference and improve local structural details. CogView3 \cite{zheng2024cogview3} is a relayed cascading diffusion framework. It first creates low-resolution images in the base stage and then starts diffusion from the results generated in the first stage, achieving finer and faster T2I generation.

\textbf{U-Net}. Some works focus on modifying U-net \cite{ronneberger2015u} for enhancing performance. U-Net backbone helps in denoising, while skip connections introduce high-frequency features into the decoder module, causing the network to ignore backbone semantics. Therefore, FreeU \cite{si2024freeu} uses dynamic scaling factors to scale the features of the backbone and skip connections, thereby improving generation quality without additional training or fine-tuning. SCEdit \cite{jiang2024scedit} uses a lightweight fine-tuning module SC-Tuner to integrate and edit jump connections. SC-Tuner consists of a tuning operation (e.g., LoRA, Adapter, Prefix, and their combination) and a residual connection. The framework can be easily extended to controllable image synthesis by injecting different conditions into SC-Tuner.

\textbf{Mixture-of-Experts (MoE)}. Some works introducing additional modules (MoE) for enhancing performance. ERNIE-ViLG 2.0 \cite{feng2023ernie} is a large-scale Chinese T2I diffusion model. It enhances image quality by incorporating knowledge enhancement (integrating fine-grained textual and visual knowledge into images) and mixing denoising experts (using different denoising experts in different denoising stages). Using unique temporal dynamics of T2I diffusion models during generation, eDiff-I \cite{balaji2022ediff} combines multiple expert denoisers (for different noise levels, trained based on binary tree strategy) and encoders (providing input information for diffusion models) to improve generation quality. It also controls the output layout of T2I through cross-attention modulation. RAPHAEL \cite{xue2024raphael} uses a combination of temporal and spatial mixture experts to accurately depict text prompts and generate highly artistic images. It achieves billions of diffusion paths from network input to output by stacking dozens of MoEs layers. Each path intuitively acts as a "painter", depicting specific text concepts onto designated image regions during the diffusion steps.

\textbf{Hybrid Model}. There are already some models \cite{xiao2021tackling, geng2024improving} that combine the diffusion model with GAN for generation. They improve the generation quality and diversity of the model by introducing adversarial losses to diffusion model. In the field of T2I generation, YOSO \cite{luo2024you} proposes a single-step diffusion GAN model to solve the problem that the small number of step generation of diffusion GAN is not efficient enough. Inspired by cooperative learning \cite{xie2018cooperative, xie2021learning, xie2022tale, hill2022learning}, it constructs a self-cooperative learning objective. It also uses pre-trained T2I diffusion models for initialization of self-cooperative diffusion GANs, thus avoiding expensive training from scratch. Diffusion2GAN \cite{kang2024distilling} distills a pre-trained multi-step diffusion model into a one-step generator trained using conditional GAN and perceptual loss. It treats diffusion distillation as a pairwise image-to-image translation task using noise-to-image pairs of ODE trajectories. The perceptual loss used is E-LatentLPIPS, which operates directly in latent space without decoding to pixel space. UFOGen \cite{xu2024ufogen} is a one-step generative model. Unlike traditional methods that focus on improving samplers \cite{bao2022analytic, lu2022dpm, lu2022dpm++, song2020denoising, li2024snapfusion} or adopt distillation techniques for diffusion models \cite{meng2023distillation, salimans2022progressive, song2023consistency, luo2023latent}, UFOGen uses a pre-trained diffusion model for initialization and combines the diffusion model with the GAN objective to achieve fast one-step text-to-image generation.

\subsubsection{Prompt and Embedding} \label{3Diffusion Prompt and Embedding}
\leavevmode \\
\textbf{Vector Quantization}. VQ-Diffusion \cite{gu2022vector} utilizes a pre-trained VQ-VAE \cite{van2017neural} to obtain discrete image tokens, then models the discrete latent space of VQ-VAE with a conditional variant of DDPM \cite{ho2020denoising}. This is a non-autoregressive method. This approach not only eliminates the unidirectional bias of existing methods but also allows for the adoption of mask-and-replace discrete diffusion strategies to avoid error accumulation. Frido \cite{fan2023frido} models latent features from coarse to fine through a feature pyramid diffusion model. Initially, inspired by VQGAN \cite{esser2021taming}, it trains a multi-scale vector quantization model (MS-VQGAN) to encode images into latent codes at multiple spatial levels. Then, it employs the feature pyramid diffusion model to extend diffusion and denoising in a multi-scale, coarse-to-fine manner.

\textbf{Text-image Alignment}. Some works focus on the methods for enhancing text-image alignment. One key finding of Imagen \cite{saharia2022photorealistic} is that increasing the size of the language model (large-scale language models pre-trained on pure text corpora, such as T5 \cite{raffel2020exploring}) performs better than increasing the size of the diffusion model (U-Net \cite{ronneberger2015u}). Large classifier-free guidance \cite{ho2022classifier} weights can improve image-text alignment but may compromise image fidelity. Imagen addresses this issue by using dynamic thresholds to avoid pixel saturation. Re-Imagen \cite{chen2022re} is a retrieval-augmented T2I generation model. Given text prompts, Re-Imagen accesses an external multimodal knowledge base to retrieve relevant (image, text) pairs and uses the retrieved information as additional inputs to the model to generate high-fidelity and faithful images. Additionally, it adopts classifier-free guidance based on intertwined text and retrieval information to balance the alignment between text and retrieval information. Predicated Diffusion \cite{sueyoshi2024predicated} is proposed to solve the problem that existing T2I generation models cannot fully follow the prompts for correct generation. It uses predicate logic \cite{genesereth2012logical} to represent the relationship between words in the prompt as a proposition, and then uses attention maps and fuzzy logic \cite{hajek2013metamathematics, prokopowicz2017theory} to guide the model for generation with achieving the proposition (following the prompts).

\textbf{Prompt Optimization}. Users craft specific prompts to generate better images, but the images can be repetitive. Prompt Expansion \cite{datta2023prompt} takes a text query as input and outputs a set of optimized expanded text prompts that, when passed to a T2I model, can generate a wider variety of appealing images. DALL-E3 \cite{betker2023improving} fine-tunes a CLIP-based specialized image caption generator in two steps and uses it to regenerate the captions of training dataset, thereby improving the T2I model's ability to follow prompts. CosmicMan \cite{li2024cosmicman} focuses on human image generation. It uses a new data generation paradigm (Annotate Anyone) to obtain high-quality images and generate refined annotations for these images at a low cost. It also uses decomposed-attention-refocusing to model the relationship between text descriptions and image pixels: the description is decomposed into a fixed number of groups related to the human body structure, and then decomposed cross-attention is performed at the group level.

\subsubsection{Training Strategies} \label{3Diffusion Training Strategies}
\leavevmode \\
\textbf{Distillation}. Adversarial diffusion distillation (SDXL-Turbo) \cite{sauer2023adversarial} utilizes score distillation sampling (SDS) \cite{poole2022dreamfusion} to leverage large-scale image diffusion models (SD \cite{rombach2022high} / SDXL \cite{podell2023sdxl}) as teacher signals and combines adversarial loss to ensure high fidelity of images even with only one to two sampling steps. SwiftBrush \cite{nguyen2024swiftbrush} is a 3D NeRF-inspired image distillation scheme for text-to-3D synthesis. It utilizes an improved version of SDS, variational score distillation, to distill pre-trained multi-step T2I models into student networks capable of generating high-fidelity images in just a single inference step.

\textbf{Neighbor}. kNN-Diffusion \cite{sheynin2022knn} is a T2I generation and editing model based on CLIP and k-Nearest-Neighbors (kNN). It uses kNN to retrieve the nearest image embeddings to the given text embeddings in the CLIP space, thus training a T2I diffusion model without text-image pairs. Additionally, it can perform local editing based on CLIP and kNN without relying on user-provided masks. ConPreDiff \cite{yang2024improving} reinforces each point's prediction of its neighbor during the training phase, allowing each point to better reconstruct itself by retaining semantic connections with its neighbor. This new paradigm can be extended to arbitrary discrete and continuous diffusion backbones without introducing additional parameters during sampling.

\textbf{Shifted Diffusion}. Corgi \cite{zhou2023shifted} is a shifted diffusion model that introduces a shift term at each time step of the diffusion process, making the sampling trajectory closer to the distribution of real image embeddings, which lie in a small region of the embedding space. This method integrates CLIP prior knowledge into its diffusion process, bridging the gap between image and text modalities. It is versatile and applicable to various settings, such as semi-supervised and language-free T2I generation, bridging the gap in data availability.

\textbf{Disentanglement between structure and texture}. PanGu-Draw \cite{lu2023pangu} uses a resource-efficient time-disentangled training strategy to divide the T2I model into structure and texture generators and train them using different schemes. Additionally, it introduces Coop-Diffusion to cooperatively use different pre-trained diffusion models in a unified denoising process.

\textbf{Loss}. The effectiveness of classifier-free guidance partly stems from its implicit perceptual guidance. Since the diffusion model itself is a perceptual network and can be used to generate meaningful perceptual losses, Self-perception \cite{lin2023diffusion} proposes self-perception objectives, introducing perceptual losses into the diffusion training process, allowing the diffusion model to generate more realistic samples.

\textbf{Reinforcement Learning (RL)}. Parrot \cite{lee2024parrot} uses batch-wise Pareto optimal selection to balance between different rewards in reinforcement learning. It jointly optimizes the T2I model and Prompt Expansion \cite{datta2023prompt} to generate high-quality text prompts and introduces original prompt-centered guidance during inference to ensure faithful image generation. RL-Diffusion \cite{zhang2024large} proposes an effective scalable algorithm using reinforcement learning to improve diffusion model. Its goal is to fine-tune the parameters of existing diffusion models to maximize the reward function of images generated from sampling process. The reward function covers human preferences, compositionality, fairness, etc. Consistency models (CM) \cite{song2023consistency} map noise directly to data, thereby generating images in at least one sampling iteration. RLCM \cite{oertell2024rl} constructs the iterative reasoning process of CM as a reinforcement learning process, which improves the diffusion model of reinforcement learning refinement on T2I generation and trades computation for sample quality at inference time.

\textbf{LoRA}. Existing methods face challenges in combining multiple LoRAs \cite{hu2021lora}. To this end, Multi-LoRA Composition \cite{zhong2024multi} proposes two training-free methods: LoRA Switch, which alternates between different LoRAs at each denoising step, and LoRA Composite, which simultaneously integrates all LoRAs to guide more consistent image synthesis.

\subsubsection{Application} \label{3Diffusion Application}
\leavevmode \\
\textbf{Scene Generation}. SceneGraph2Image \cite{mishra2024scene} proposes scene-graph-to-image generation without the need to predict layout. A scene graph is a layout map composed of objects and their relationships described in text. SceneGraph2Image first pre-trains a scene graph encoder, then uses object embeddings and relationship embeddings to create CLIP-guided conditioning signals to fine-tune pre-trained diffusion models. Text2Street [] generates controllable street scene images using a lane-aware road topology generator, a position-based object layout generator, and a multi-control image generator that integrates road topology, object layout, and weather descriptions.

\textbf{Hand Generation}. Currently, hand generation is a prominent issue in image generation. People can easily recognize which images are fake according to errors in hands generation (e.g., artifacts, unnatural fingers or incorrect finger counts). Recently, some models have been proposed to mitigate this problem. HanDiffuser \cite{su2024text2street} is an architecture focusing on hand generation based on diffusion. It consists of two components: a Text-to-Hand-Params diffusion model for generating SMPL-Body \cite{loper2023smpl} and MANO-Hand \cite{romero2022embodied} parameters from input text, and a Text-Guided Hand-Params-to-Image diffusion model for synthesizing images conditioned on prompts and previously generated hand parameters. Giving a Hand to Diffusion Model \cite{pelykh2024giving} proposes a pose-conditioned human image generation method. It first trains a diffusion-based hand generator in a multi-task setting to generate hand images and their corresponding segmentation masks based on keypoint heatmaps; then it uses a fine-tuned ControlNet \cite{zhang2023adding} to draw bodies around generated hands to generate complete images.

\subsection{Others} \label{Others}
\subsubsection{Architecture} \label{Others Architecture}
\leavevmode \\
\textbf{Energy-based models} \cite{lecun2006tutorial, zhao2016energy} use an energy function to predict the output having the minimum energy with input. PPGN \cite{nguyen2017plug} proposed an energy-based framework for conditional image generation. It introduces additional priors for latent code to enhance sample quality and diversity. This is a plug-and-play method that can use any conditional generative backbone, even simultaneously employing multiple conditional networks for image generation.

\textbf{Mamba-based models}. Transformer \cite{vaswani2017attention} faces challenges such as quadratic computation and high memory due to its attention mechanisms, resulting in slow inference speed, low throughput, and difficulties in handling long contexts. Mamba \cite{gu2023mamba, dao2024transformers} is proposed to address these issues. Mamba is based on the State Space Model (SSM) \cite{gu2021efficiently} where the current state depends only on previous state and current input. Mamba (Selective SSM) uses selection to compress useful information in the input, thereby mitigating the problems faced by Transformers. It can handle longer contexts with performance close to Transformer. However, due to its information compression characteristics, its ability for in-context learning is not as strong as Transformers. Jamba \cite{lieber2024jamba} fully leverages the advantages of Transformers and Mamba, by interleaving blocks of Transformer and Mamba layers, along with additional Mixture of Experts (MoE).

Due to Mamba's ability to model long sequences, it has been applied in the field of image generation \cite{zhu2024vision, liu2024vmambavisualstatespace, fei2024scalable}. Subsequently, ZigMa \cite{hu2024zigma} utilized it for T2I generation. ZigMa replaces multi-head self-attention blocks in DiT with Zigzag Mamba blocks, reducing quadratic complexity to linear complexity. Additionally, it uses Zigzag scanning to address the lack of consideration for spatial continuity in most Mamba-based visual models. ZigMa can also generalize to 3D video generation by decomposing spatial and temporal information.

\textbf{Multimodal}. Besides using text prompts as conditions, multimodal T2I generation also utilizes other information (or priors) to assist its generation. Versatile Diffusion \cite{xu2023versatile} extends existing single-flow diffusion into a multi-task multi-modal network that can handle multiple flows of T2I, image-to-text, and variations in one unified model. It instantiates a unified multi-flow diffusion framework composed of shareable and swappable layer modules, enabling the cross-modal generality beyond images and text. GLIGEN \cite{li2023gligen} is a multimodal image generation diffusion model. In addition to conditioning on text, it can control conditional generation with grounding input (e.g., bounding boxes and keypoints). It freezes all weights of the pretrained model to retain its extensive conceptual knowledge and injects grounding information into new trainable layers through a gating mechanism. MiniGPT-5 \cite{zheng2023minigpt} is based on the concept of "Generative Voken" and aligns LLM with a pretrained T2I generation model to generate images with coherent textual narratives. It introduces a two-stage training strategy for multimodal generation without description, where training does not require a comprehensive description of the image. DiffusionGPT \cite{qin2024diffusiongpt} constructs domain-specific trees for various generation models. Given prompts, it uses LLM to parse prompts and selects appropriate models using thought trees and human feedback to ensure excellent performance in different domains. RPG-DiffusionMaster \cite{yang2024mastering} is a framework for untrained T2I generation and editing. It uses a multimodal Large Language Model (MLLM) as a global planner, leveraging the powerful inferencing capabilities of MLLM to decompose the process of generating complex images into multiple simple generation tasks, which are executed within subregions. Region-wise composition generation is achieved through complementary regional diffusion. UNIMO-G \cite{li2024unimo} generates images based on interleaved text and visual inputs. It includes a MLLM for encoding multimodal prompts and a conditional denoising diffusion network for generating images based on multimodal inputs. It employs a two-stage training strategy: first pretraining the diffusion model on a large-scale Chinese text-image pair dataset, then fine-tuning with multimodal prompts generated by MLLM to improve image fidelity and fidelity. CompAgent \cite{wang2024divide} uses a LLM agent to decompose the given complex text prompts, including extracting objects, their associated attributes, and predicting a coherent scene layout. The agent then uses tools to combine these objects and correct errors and refine the resulting image based on verification and human feedback.

\section{Image Editing} \label{Image Editing}

\begin{figure} [!ht]
\centering
\tikzset{
    rnode/.style = {thin, rounded corners=2pt, align=left, fill=red!30, text width=1.3cm},
    bnode/.style = {thin, rounded corners=2pt, align=left, fill=blue!15, text width=2.2cm},
    onode/.style = {thin, rounded corners=2pt, align=left, fill=orange!30, text width=1.8cm},
    pnode/.style = {thin, rounded corners=2pt, align=left, fill=pink!30, text width=6.5cm},
}

\begin{forest}
for tree={
    inner sep=0pt,
    outer sep=0pt,
    fit=band,
    child anchor=west,
    parent anchor=east,
    grow'=0,
    anchor=west,
    align=left,
    edge path={\noexpand\path[\forestoption{edge}] 
         (!u.parent anchor) -- +(3pt,0pt) |- (.child anchor)
         \forestoption{edge label};},
    font=\tiny, 
}    
[
    [Sec.\ref{4Autoregression} AR, rnode
        [Sec.\ref{4AR Application} Application, bnode
            [Scene Editing, onode
                [Make-A-Scene~\cite{gafni2024scene}, pnode]
            ]            
        ]
    ]
    [Sec.\ref{4NAR} NAR, rnode
        [Sec.\ref{4NAR Training Strategies} Training Strategies, bnode
            [Fine-tuning, onode
                [StyleDrop~\cite{sohn2023styledrop}, pnode]
            ]            
        ]
    ]
    [Sec.\ref{4GAN} GAN, rnode
        [Sec.\ref{4GAN Architecture} Architecture, bnode
            [Editing Layer, onode
                [Text2LIVE~\cite{bar2022text2live}, pnode]
            ]
        ]
        [Sec.\ref{4GAN Prompt and Embedding} Prompt \\
        and Embedding, bnode
            [CLIP, onode
                [StyleGAN-NADA~\cite{gal2022stylegan}, pnode]
            ]    
            [Disentanglement, onode
                [Edit One for All~\cite{nguyen2024edit}, pnode]
            ]            
        ]
    ]
    [Sec.\ref{4Diffusion} Diffusion, rnode
        [Sec.\ref{4Diffusion Architecture} Architecture, bnode
            [SDE, onode
                [SDEdit~\cite{meng2021sdedit}, pnode]
            ]            
            [Editing Layer, onode
                [LayerDiffusion~\cite{zhang2024transparent}, pnode]
            ]
            [MoE, onode
                [SuTI~\cite{chen2024subject}, pnode]
            ]
        ]
        [Sec.\ref{4Diffusion Prompt and Embedding} Prompt and \\
        Embedding, bnode
            [Pseudo-text, onode
                [Textual Inversion~\cite{gal2022image} Null-text Inversion~\cite{mokady2023null} 
                InST~\cite{zhang2023inversion} WaveOpt-Estimator~\cite{koo2024wavelet}, pnode]
            ]
            [Learnable Encoder, onode
                [ELITE~\cite{wei2023elite} InstantBooth~\cite{shi2024instantbooth}, pnode]
            ]
            [Text-image Alignment, onode
                [BLIP-Diffusion~\cite{li2024blip} PALP~\cite{arar2024palp} \textbf{UniTune}~\cite{valevski2023unitune}, pnode]
            ]
            [Embedding, onode
                [ProSpect~\cite{zhang2023prospect} Imagic~\cite{kawar2023imagic} Get What You Want~\cite{li2024get} PhotoMaker~\cite{li2024photomaker} SeFi-IDE~\cite{li2024sefi} \\ 
                \textbf{Attribute-Control}~\cite{baumann2024continuous} 
                \textbf{Object-Driven One-Shot Fine-tuning}~\cite{lu2024object}, pnode]
            ]
            [Soft Prompt, onode
                [DreamDistribution~\cite{zhao2023dreamdistribution}, pnode]
            ]
            [Instruction, onode
                [InstructPix2Pix~\cite{brooks2023instructpix2pix} ZONE~\cite{li2024zone}, pnode]
            ]
        ]
        [Sec.\ref{4Diffusion Training Strategies} Training Strategies, bnode
            [Fine-tune, onode
                [DreamBooth~\cite{ruiz2023dreambooth} Custom Diffusion~\cite{kumari2023multi} Cones~\cite{liu2023cones} BootPIG~\cite{purushwalkam2024bootpig} DiffusionCLIP~\cite{kim2022diffusionclip} \\
                Object-Driven One-Shot Fine-tuning~\cite{lu2024object}                 UniTune~\cite{valevski2023unitune} IP-Adapter~\cite{ye2023ip}, pnode]
            ]
            [Segmentation, onode
                [Plug-and-Play Diffusion~\cite{tumanyan2023plug} Text2Scene~\cite{hwang2023text2scene} 
                Make-A-Storyboard~\cite{su2023make} \\
                Pick-and-Draw~\cite{lv2024pick} 
                Prompt-to-Prompt~\cite{hertz2022prompt} Factorized Diffusion~\cite{geng2024factorized} \\
                MIGC~\cite{zhou2024migc} \textbf{ZONE}~\cite{li2024zone} \textbf{Composer}~\cite{huang2023composer}, pnode]
            ]
            [Mask and Predict, onode
                [Blended Diffusion~\cite{avrahami2022blended} DiffEdit~\cite{couairon2023diffedit} 
                SmartBrush~\cite{xie2023smartbrush} Imagen Editor~\cite{wang2023imagen} \\
                MasaCtrl~\cite{cao2023masactrl} InstDiffEdit~\cite{zou2024towards}, pnode]
            ]
            [Disentanglement, onode
                [DiffStyler~\cite{huang2024diffstyler} ControlStyle~\cite{chen2023controlstyle} FreeStyle~\cite{he2024freestyle}              CreativeSynth~\cite{huang2024creativesynth} MagicMix~\cite{liew2022magicmix} \\ Semantic-Visual Alignment~\cite{abreu2023addressing}              DEADiff~\cite{qi2024deadiff} SepME~\cite{zhao2024separable} Ranni~\cite{feng2024ranni} \\
                One-dimensional Adapter~\cite{lyu2024one} DisenDiff~\cite{zhang2024attention} Self-Guidance~\cite{epstein2023diffusion} Attribute-Control~\cite{baumann2024continuous} \\               \textbf{ProSpect}~\cite{zhang2023prospect} \textbf{IP-Adapter}~\cite{ye2023ip} \textbf{UniControl}~\cite{qin2023unicontrol}, pnode]
            ]
            [Attention Sharing, onode
                [StyleAligned~\cite{hertz2024style}, pnode]
            ]
            [Multimodal, onode
                [ControlNet~\cite{zhang2023adding} ControlNet-XS~\cite{zavadski2023controlnet} T2I-Adapter~\cite{mou2024t2i} Composer~\cite{huang2023composer} \\ UniControl~\cite{qin2023unicontrol} CtrlColor~\cite{liang2024control}, pnode]
            ]
        ]
        [Application, bnode
            [Personalized \\
            Generation, onode
                [\textbf{Textual Inversion}~\cite{gal2022image} \textbf{DreamBooth}~\cite{ruiz2023dreambooth} 
                \textbf{Plug-and-Play Diffusion}~\cite{tumanyan2023plug} \\
                \textbf{Custom Diffusion}~\cite{kumari2023multi}                \textbf{ELITE}~\cite{wei2023elite} \textbf{Cones}~\cite{liu2023cones} \textbf{SuTI}~\cite{chen2024subject} \textbf{InstantBooth}~\cite{shi2024instantbooth} \\
                \textbf{BLIP-Diffusion}~\cite{li2024blip} \textbf{ProSpect}~\cite{zhang2023prospect} \textbf{Text2Scene}~\cite{hwang2023text2scene} 
                \textbf{Make-A-Storyboard}~\cite{su2023make} \\ \textbf{DreamDistribution}~\cite{zhao2023dreamdistribution} 
                \textbf{PALP}~\cite{arar2024palp} \textbf{Object-Driven One-Shot Fine-tuning}~\cite{lu2024object} \\
                \textbf{PhotoMaker}~\cite{li2024photomaker} \textbf{SeFi-IDE}~\cite{li2024sefi} \textbf{BootPIG}~\cite{purushwalkam2024bootpig} \textbf{CreativeSynth}~\cite{huang2024creativesynth} \\
                \textbf{Pick-and-Draw}~\cite{lv2024pick} 
                \textbf{SepME}~\cite{zhao2024separable} 
                \textbf{Get What You Want}~\cite{li2024get} 
                \textbf{Ranni}~\cite{feng2024ranni} \\
                \textbf{One-dimensional Adapter}~\cite{lyu2024one} \textbf{MIGC}~\cite{zhou2024migc} \textbf{DisenDiff}~\cite{zhang2024attention}, pnode]
            ]
        ]
    ]
]
\end{forest}

\caption{Taxonomy of image editing. Papers are categorized according to its main contribution. Note that papers falling into other categories according to its sub-important contributions are shown in bold and the description of which are not given again for concise.\protect}

\label{fig:editing}
\end{figure}
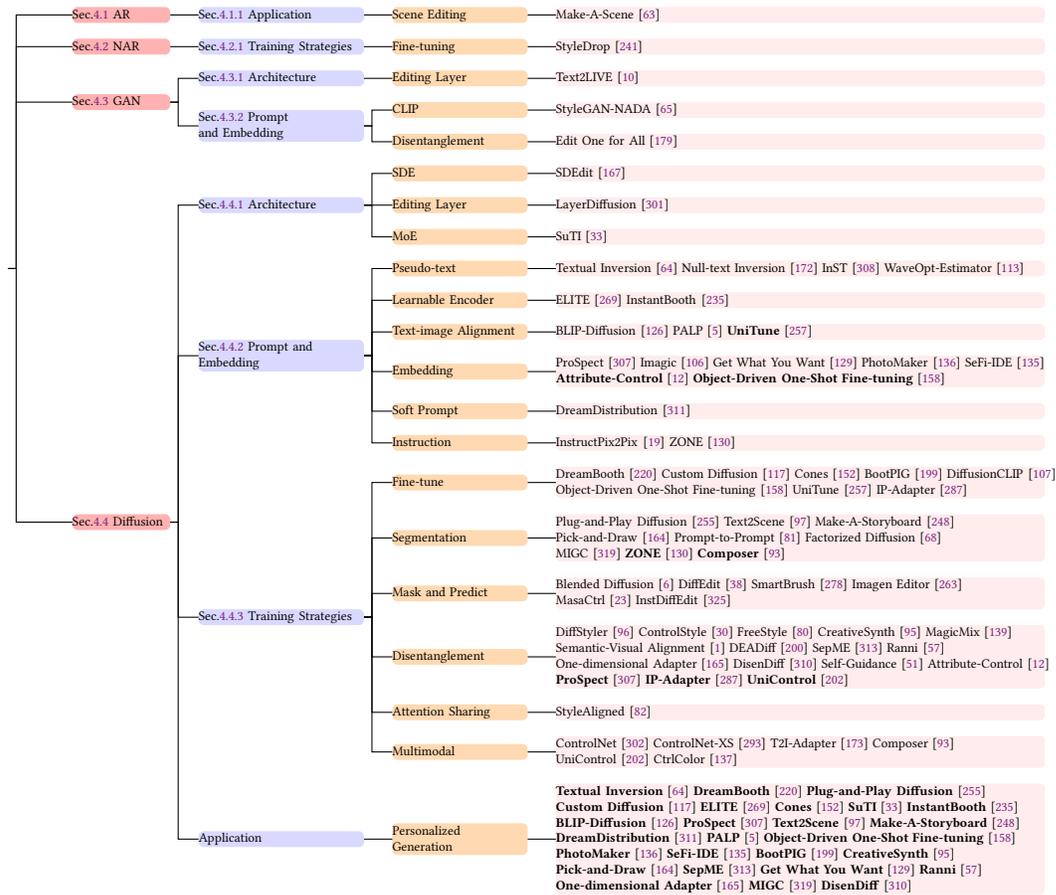

In this section, we present a taxonomy of papers focusing on text-guided image editing. For consistency, this section uses a similar taxonomy as Section~\ref{Image Generation}, see Figure~\ref{fig:editing} for details.

\subsection{Autoregression} \label{4Autoregression}
\subsubsection{Application} \label{4AR Application}
\leavevmode \\
\textbf{Scene Editing}. Make-A-Scene \cite{gafni2024scene} uses scene tokens encoded from the scene layout (segmentation map) in addition to traditional text and image tokens, allowing for layout-constrained image generation and editing. Scene tokens are created by VQ-SEG, a modified VQ-VAE for semantic segmentation.

\subsection{Non-Autoregression (NAR)} \label{4NAR}
\subsubsection{Training Strategies} \label{4NAR Training Strategies}
\leavevmode \\
\textbf{Fine-tuning}. StyleDrop \cite{sohn2023styledrop} is a Parameter Efficient Fine Tuning (PEFT) framework that focuses on Muse \cite{chang2023muse} and uses adapter tuning. It performs iterative training based on feedback (CLIP scores or human feedback) to improve generation quality. Impressive results can be achieved even if the user only provides a single image in desired style. Note that the framework is not limited to specific models and applications and can be easily applied to fine-tuning T2I and text-to-video Transformers using various PEFT methods.

\subsection{GAN} \label{4GAN}
\subsubsection{Architecture} \label{4GAN Architecture}
\leavevmode \\
\textbf{Editing Layer}. Text2LIVE \cite{bar2022text2live} uses an image layering method for text-guided image editing. Given an input image, the generator produces an editing layer (color + opacity). Image editing is then done by linearly interpolating between the color layer and the input image via opacity weighting. Apart from being used to guide the generation of edit layers, text is used to initialize the position of editing instead of using a user-supplied mask. Note that, in addition to using GAN backbone for image editing, Text2LIVE can utilize Diffusion backbone.

\subsubsection{Prompt and Embedding} \label{4GAN Prompt and Embedding}
\leavevmode \\
\textbf{CLIP}. StyleGAN-NADA \cite{gal2022stylegan} introduces CLIP into StyleGAN \cite{karras2019style} for domain adaptation guided only by text. Compared with the guidance of adversarial loss used by traditional GAN-based methods, StyleGAN-NADA exploits the properties of image-text alignment in CLIP space and uses CLIP loss to guide image editing.

\textbf{Disentanglement between different attributes}. Edit One for All \cite{nguyen2024edit} is a batch interactive image editing method. In addition to performing batch image editing based on interaction (for example, dragging a point in the image, inspired by DragGAN \cite{pan2023drag} and DragDiffusion \cite{shi2024dragdiffusion}), batch image editing based on text can also be performed. The core idea is to find latent vectors corresponding to target attributes in the disentangled latent space of StyleGAN \cite{karras2019style}, allowing the edit to be transferred to any new image.

\subsection{Diffusion} \label{4Diffusion}
\subsubsection{Architecture} \label{4Diffusion Architecture}
\leavevmode \\
\textbf{SDE}. SDEdit \cite{meng2021sdedit} is a method for image editing based on SDE diffusion prior. Given an input image with user guidance (in the form of manipulating RGB pixels), SDEdit first adds noise to input and then iteratively denoises it through SDE priors to enhance the realism of the generated image. This method employs a pre-trained SDE model on unlabeled images for guided image synthesis and editing.

\textbf{Editing layer}. LayerDiffusion \cite{zhang2024transparent} learns a latent transparency by encoding alpha channel transparency into the latent manifold of Stable Diffusion. By regulating the added transparency as a latent offset, LayerDiffusion minimally changes the original latent distribution of Stable Diffusion, preserving generation quality.

\textbf{MoE}. Some works introducing additional modules (MoE) for image editing. SuTI \cite{chen2024subject} collects millions of image clusters around different specific visual subjects from the Internet and then uses these clusters to train a large number of expert models focused on different subjects. The apprentice SuTI model learns to mimic these fine-tuned experts for image synthesis conditioning on unseen captions.

\subsubsection{Prompt and Embedding} \label{4Diffusion Prompt and Embedding}
\leavevmode \\
\textbf{Pseudo-text}. Textual Inversion \cite{gal2022image} inverses the concepts (images) provided by user into new pseudo-words in the embedding space of a pretrained T2I model, then injects these pseudo-words into new scenarios through simple natural language descriptions, thereby achieving personalized creation. Null-text Inversion \cite{mokady2023null} proposes an accurate inversion method. It uses pivotal noise rather than random noise at each time step and optimizes around it. Additionally, it only modifies unconditional text embeddings in classifier-free guidance, thus avoiding tedious tuning to the model. InST \cite{zhang2023inversion} introduces an attention-based text inversion that can quickly learn key features from images, as well as a random inversion that maintains semantic content of source images. It uses inversion to obtain text embeddings corresponding to input CLIP image embeddings. Then, the diffusion model generates new images with reference style conditioning on text embeddings. To address the time-consuming nature of null-text inversion, WaveOpt-Estimator \cite{koo2024wavelet} determines the endpoints of text optimization based on frequency features and accelerates image editing using wavelet transform and Negative-Prompt Inversion \cite{miyake2023negative}.

\textbf{Learnable Encoder}. ELITE \cite{wei2023elite} proposes a learning-based encoder composed of global and local mapping networks. The former projects hierarchical features of a given image into multiple "new" words in the word embedding space, where one main word is used for easily editable concepts, and other auxiliary words are used to exclude irrelevant interference. The latter injects encoded patch features into cross-attention layers to provide omitted details without sacrificing the editability of the main concept. InstantBooth \cite{shi2024instantbooth} achieves instant personalized text-guided image generation without any test-time fine-tuning. It learns the general concepts of input images by converting them into text tokens using a learnable image encoder and enriches visual feature representations by introducing several adapter layers, preserving fine details of pretrained models.

\textbf{Text-image Alignment}. Some works focus on the methods for enhancing text-image alignment. BLIP-Diffusion \cite{li2024blip} follows BLIP-2 \cite{li2023blip} to pretrain a multimodal encoder to generate representations visually aligned with text, and designs a subjects representation learning task to enable the diffusion model to utilize this visual representation and generate new versions of subjects. PALP \cite{arar2024palp} proposes prompt-aligned personalization to achieve single-prompt personalized generation. It uses score distillation sampling \cite{poole2022dreamfusion} to keep personalized models aligned with target prompts. Furthermore, it addresses the less diversity and oversaturation issues induced by score distillation sampling by using Delta denoising scores. 

\textbf{Embedding}, referring to embedding methods and operations in embedding space. ProSpect \cite{zhang2023prospect} introduces Prompt Spectrum Space, an extended conditioning embedding space. Various visual attributes (i.e., material/style (high frequency), content (medium frequency), and layout (low frequency) ) can be separated from embeddings, achieving attribute transfer by replacing them with embeddings of other images. Imagic \cite{kawar2023imagic} performs linear interpolation between target text embeddings and optimized embeddings (used to generate images similar to the input image) to get a representation that combines the input image and target text. It passes this representation to a fine-tuned model to generate the final edited images. Get What You Want \cite{li2024get} removes unnecessary content from text embeddings using soft weighting regularization and text embedding optimization during inference. The former regularizes text embedding matrices to effectively suppress unwanted content. The latter further suppresses unwanted content generation from prompts, encouraging generation of desired content.

PhotoMaker \cite{li2024photomaker} encodes any number of input ID images into stacked ID embeddings to retain ID information. The cross-attention layers of the generation model itself can adaptively integrate the ID information contained in stacked ID embeddings for personalized T2I generation without adding extra modules. SeFi-IDE \cite{li2024sefi} focuses on stabilizing the diffusion model for personalized generation with accurate and semantically faithful ID embeddings. It improves ID accuracy with facial attention loss and enhances semantic fidelity control by optimizing the extended text conditioning space with semantic fidelity tokens. 

\textbf{Soft Prompt}. DreamDistribution \cite{zhao2023dreamdistribution} allows pretrained T2I diffusion models to learn a set of soft prompts (capturing commonalities and variations in visual feature space), creating instances with sufficient variation while preserving commonalities in reference images by sampling from the learned prompt distribution.  

\textbf{Instruction}. InstructPix2Pix \cite{brooks2023instructpix2pix} proposes editing images based on instructions: given an input image and instructions on how to manipulate it, the model follows the instructions to edit the image. To obtain training data, it integrates two large pretrained models: a language model and a T2I model, generating a large dataset of image editing examples. ZONE \cite{li2024zone} converts instructions into specific image editing regions using InstructPix2Pix. It proposes a Region-IoU method that can precisely extract segmentation mask and develops an edge smoother based on FFT for seamless blending between layers and images.

\subsubsection{Training Strategies} \label{4Diffusion Training Strategies}
\leavevmode \\
\textbf{Fine-tune}. DreamBooth \cite{ruiz2023dreambooth} fine-tunes a pretrained T2I model with some images of the input subject, enabling it to associate unique identifiers with specific subjects. The identifiers can be used to synthesize new realistic images of the subject in different scenarios. Additionally, it uses prior-preserving losses to encourage diversity and offset language drift \cite{lee2019countering, lu2020countering}. Custom Diffusion \cite{kumari2023multi} proposes that optimizing only a few parameters in the T2I conditioning mechanism can represent new concepts and achieve rapid tuning. Furthermore, multiple concepts can be simultaneously trained through closed-form constraints optimization or merging multiple fine-tuned models into one, achieving multi-concept compositional generation. Do generative diffusion models mimic the imaginative ability of human brains using their neurons to encode different subjects? Cones \cite{liu2023cones} answers this question from the perspective of subject-driven generation. It finds a small cluster of neurons, which are parameters of the attention layer in pre-trained T2I diffusion models. By changing the values of these neurons, corresponding subjects can be generated in different content based on the semantics in the input text prompts. Object-Driven One-Shot Fine-tuning \cite{lu2024object} initializes a prototype embedding based on the appearance and class of an object, then fine-tunes the diffusion model. Class-characterizing regularization is used during fine-tuning to retain prior knowledge of object class, and object-specific losses are introduced to further improve fidelity. This method can be used to implant multiple objects. BootPIG \cite{purushwalkam2024bootpig} uses data generated by pretrained T2I models, LLM chat agents, and image segmentation models to guide personalized generation. It minimally modifies the pretrained T2I diffusion model and uses an independent UNet to guide image generation towards the desired appearance. The limited inversion capability of GAN makes it difficult to handle diverse real images. To address this issue, DiffusionCLIP \cite{kim2022diffusionclip} utilizes the full inversion capability and high-quality image generation ability of diffusion models to achieve zero-shot image manipulation across unseen domains. Additionally, it proposes a noise blending method to directly perform multi-attribute operations. UniTune \cite{valevski2023unitune} converts image generation models into image editing models by fine-tuning on a single image. It balances fidelity and prompt alignment by controlling the number of fine-tuning iterations and the weights of classifier-free guidance or replacing some initial sampling steps with noised base image. IP-Adapter \cite{ye2023ip} is an efficient and lightweight adapter with a key design of disentangled cross-attention mechanism, separating cross-attention layers of text and image features. It adds an additional cross-attention layer for image features to each cross-attention layer in UNet. During training, only new cross-attention layers are trained while the original UNet keeping frozen. 

\textbf{Segmentation}. Plug-and-Play Diffusion \cite{tumanyan2023plug} introduces T2I generation into the field of image-to-image translation. It manipulates the spatial features and their self-attention inside the model to achieve fine-grained control over the generated structure, generating images that conform to the target text while preserving the semantic layout of the guided image. Text2Scene \cite{hwang2023text2scene} adds detailed textures to labeled 3D geometry in the scene. Rather than applying stylization to the entire scene in a single step, it obtains weak semantic prompts through geometric segmentation and assigning initial colors to the segments. It then adds textural detail to each object so that their projection on image space exhibits consistency with the input embedding. Make-A-Storyboard \cite{su2023make} is a universal framework that applies disentangled control in the consistency of context-relevant roles and scenes and merges them to form harmonious images. Additionally, it enhances generative performance using balance-aware merging and alternate control. Pick-and-Draw \cite{lv2024pick} is a training-free semantic-guided personalization method that uses appearance selection guidance to enhance identity consistency and combines diffusion priors with layout drawing to improve generation diversity. This method can be applied to any personalized diffusion model and requires only one reference image. Prompt-to-Prompt \cite{hertz2022prompt} observes that cross-attention layers are crucial in controlling the spatial layout of images with respect to each word in prompts. Based on this, it edits images by injecting cross-attention maps related to new prompts during the diffusion process, controlling which pixels attend to which tokens in the prompt text during which diffusion steps. Representing an image as a sum of linear components in terms of frequency, color space, motion blur, masking, scaling, etc., Factorized Diffusion \cite{geng2024factorized} proposes a zero-shot method to control each component using diffusion model sampling by assigning different prompts to each component. This method can also be extended to the inverse problem \cite{chung2022diffusion, kawar2022denoising, lugmayr2022repaint, song2021solving, wang2022zero} (generating hybrid images from real images). Multi-instance Generation Controller (MIGC) \cite{zhou2024migc} proposes a multi-instance generation task that requires the generated instances to be accurately located at the specified location and the attributes of the instances to conform to their corresponding descriptions. MIGC decomposes the task into several subtasks (each involving the colorization of a single instance) and introduces an instance enhancement attention to ensure the accurate colorization of each instance. Finally, all the colorized instances are aggregated to achieve accurate multi-instance generation. 

\textbf{Mask and Predict}. Similar to typical modeling methods in AR (sequential prediction) and NAR (parallel prediction), some works perform image editing by masking part of images and performing text-guided prediction. Blended Diffusion \cite{avrahami2022blended} introduces the first local editing method based on natural language descriptions and region-of-interest masks. It seamlessly blends the edited regions into the unchanged parts of the image by spatially mixing noised input images with local text-guided diffusion latent at a range of noise levels. DiffEdit \cite{couairon2023diffedit} automatically generates masks highlighting regions of input images that need editing by contrasting predictions from diffusion models conditioning on different textual prompts. It relies on latent inference to preserve the content of these regions of interest, synergizing well with mask-based diffusion. SmartBrush \cite{xie2023smartbrush} uses text and shape guides to complete a missing region with an object. It better preserves the background by augmenting the diffusion U-net with object-mask prediction. Furthermore, it introduces a multi-task training strategy to utilize more training data by jointly training image inpainting and T2I generation. Imagen Editor \cite{wang2023imagen} is a cascade diffusion model fine-tuned from Imagen \cite{saharia2022photorealistic}. It focuses on text-guided image inpainting: users provide an image, masked regions, and text prompts, and the model fills masked regions to match prompts and image backgrounds. Additionally, it introduces a new benchmark, EditBench, to evaluate text-guided image inpainting across multiple dimensions (attributes, objects, scenes). MasaCtrl \cite{cao2023masactrl} converts existing self-attention in the diffusion model into mutual self-attention, enabling querying relevant local content and texture from source images for consistency. To further alleviate query confusion between foreground and background, it proposes a mask-guided mutual self-attention strategy where masks can be easily extracted from cross-attention maps. InstDiffEdit \cite{zou2024towards} uses cross-modal attention for instant mask guidance during diffusion steps. It comes with a training-free refinement scheme for automatic and accurate mask generation and achieves global semantic consistency through image inpainting. 

\textbf{Disentanglement}. Some works focus on the disentanglement between content and style. DiffStyler \cite{huang2024diffstyler} proposes a dual diffusion processing framework to balance content and style in diffusion generation. It introduces learnable noise based on content images, which is used in denoising process to better preserve structural information of stylized results. ControlStyle \cite{chen2023controlstyle} upgrades a pretrained T2I model with a trainable modulation network to handle more text prompts and style image conditions. Additionally, it promotes learning modulation networks through diffusion style and content regularization. FreeStyle \cite{he2024freestyle} presents an architecture with dual-flow encoders and a single-flow decoder. Encoders take input from content images and style text prompts, and decoder modulates features based on content images and corresponding style text prompts to achieve style transfer. Combining with multimodal features and customized attention mechanisms, CreativeSynth \cite{huang2024creativesynth} introduces real-world semantic content into the field of arts through inversion and real-time style transfer, achieving accurate manipulation of style and content of art images. MagicMix \cite{liew2022magicmix} performs semantic mixing using pretrained text-conditioned diffusion models. Diffusion models have progressive generation properties: layout/shape appears in the early denoising stage, while semantic details appear in the later denoising stage. Based on this, semantic mixing is decomposed into layout semantic generation and content semantic generation. Semantic-Visual Alignment \cite{abreu2023addressing} uses data augmentation (diffusion-based semantic mixing) to increase the semantic alignment of class with visual relationships, enhancing model robustness against adversarial perturbations (augmented data from semantic mixing). DEADiff \cite{qi2024deadiff} is a style transfer model conditioned on text and reference style images. It uses Q-Formers \cite{zhang2024vision} guided by different text descriptions to obtain style and semantic representations from reference images. Then, different cross-attention layers are responsible for injecting semantic and style representations respectively.

Some works focus on the disentanglement between different concepts. SepME \cite{zhao2024separable} includes generation of concept-independent representations and weight disentangling. The former avoids forgetting concept-independently important information. The latter separates optimizable model weights, allowing each weight increment to connect to erasure of specific concepts without affecting generation performance on other concepts. Ranni \cite{feng2024ranni} decomposes text-to-image generation into two subtasks: 1) Text-to-panel: LLM is used to translate text into so-called semantic panel (containing all visual concepts); 2) Panel-to-image: panel is encoded as control signals to guide the diffusion model to capture the details of each concept. This method has the potential for chat-based continuous image editing. One-dimensional Adapter \cite{lyu2024one} builds the concept erasing framework on a one-dimensional adapter to erase multiple concepts from most diffusion models at once in multiple erasing applications. It injects the concept-semipermeable module into any diffusion models as an adapter to control what to erase and how strong the erasure is. In addition, it dynamically adjusts the permeability of each adapter in response to different input prompts to further reduce the impact on other concepts. DisenDiff \cite{zhang2024attention} proposes an attention calibration mechanism to solve the problems of existing personalized generation methods: they cannot maintain consistency with the reference image and eliminate the mutual influence between concepts. DisenDiff uses new learnable modifiers bound to categories to learn multiple decoupled concepts from a single image and generate novel customized images with the learned concepts. By suppressing the attention activation of different categories, the mutual influence between concepts can be mitigated.  

Some works focus on the disentanglement between different attributes. Self-Guidance \cite{epstein2023diffusion} allows direct control of the shape, position, and appearance of objects in generated images. It utilizes rich representations learned by pretrained T2I diffusion models to guide attributes of entities and their interactions. These constraints can be user-specified or transferred from other images and rely solely on internal knowledge of the diffusion model. Attribute-Control \cite{baumann2024continuous} observes that diffusion models can interpret local deviations in token-wise CLIP text embedding space in a semantically meaningful way. Thus, by identifying semantic directions corresponding to specific attributes, rough prompts can be enhanced into fine-grained, continuous control to attribute expression for specific objects. This idea is similar to Edit One for All \cite{nguyen2024edit}, but it utilizes CLIP's latent space, while Edit One for All uses StyleGAN's latent space.

\textbf{Attention Sharing}. By sharing attention layers between generated images, StyleAligned \cite{hertz2024style} utilizes self-attention mechanisms (allowing communication between individual generated images) and DDIM inversion [DDIM] to achieve style-consistent generation. However, full attention sharing suffers from content leakage and low diversity. Therefore, StyleAligned adopts minimal attention sharing (sharing attention to only one image).

\textbf{Multimodal}. Some work focuses on using multimodal information to enhance image editing. Note that the multimodal here is not the commonly said multimodal (e.g., Audio, Video, etc.), but other forms of images (e.g., Canny edges, Hough lines, etc.). ControlNet \cite{zhang2023adding} obtains a conditional injection module through fine-tuning, which adds various conditional controls (including Canny edges, Hough lines, user doodles, human keypoints, segmentation maps, shape normals, depth, etc.) to a large pretrained T2I diffusion model using zero convolution layers. It enhances generation quality through classifier-free guided resolution weighting. The current popular image synthesis faces a problem of delayed flow of information between generation and control processes. ControlNet-XS \cite{zavadski2023controlnet} proposes a new architecture to address this issue. Compared to ControlNet, ControlNet-XS has fewer parameters in control network and is approximately twice as fast during inference and training, resulting in higher-quality images and better fidelity in control. T2I-Adapter \cite{mou2024t2i} consists of a pre-trained stable diffusion (SD) model and multiple T2I (T2I) adapters. Adapters are used to extract guided features from different types of conditions, such as sketches, edge detection, keypoints, color, depth, and semantic segmentation. The pre-trained SD has fixed parameters to generate images based on input text features and external guided features. Composer \cite{huang2023composer}, based on the idea of compositionality, decomposes images into representative factors and then uses all these factors as conditions to recompose inputs. This method supports various levels of conditions, such as text descriptions of global information, depth maps and sketches as local guidance, and color histograms for low-level detail. UniControl \cite{qin2023unicontrol} introduces a task-aware HyperNet to the pre-trained T2I diffusion model to modulate the diffusion model so as to integrate a wide range of controllable condition-to-image (C2I) tasks into a single framework, enabling it to adapt to different C2I tasks simultaneously. It enables pixel-level accurate image generation, where visual conditions primarily influence generated structure, while text prompts guide style and content. CtrlColor \cite{liang2024control} performs unconditional and conditional (text prompts, strokes, examples) image colorization and handles color overflow and incorrect colors within a unified framework. It effectively encodes user strokes for precise local color manipulation and constrains color distribution aligning to examples.

\section{Social Impacts and Solution} \label{Social Impacts and Solution}
Social impact remains a serious topic in the field of artificial intelligence generated content (AIGC). In this section, we introduce the possible social impacts of AIGC and the solutions.

\subsection{Social Impacts} \label{Social Impacts}
\textbf{Harmful Content}. Generative AI can generate photorealistic content. As a result, they may be used to generate harmful content, such as content with violence, blood, adult content, and other offensive themes. In addition, this ability can be used to generate fake photos or videos of real people (DeepFakes), for example, to fake scenes or create images or videos that tarnish the reputation of celebrities by swapping faces.

\textbf{Bias and fairness}. Similar to most machine learning methods, generative models may inadvertently learn and persist biases in their training data. For example, CLIP \cite{radford2021learning} points out that it inevitably introduces social biases into the model since it uses unfiltered image-text pairs from the Internet for training. Images generated by PixArt-$\Sigma$ \cite{chen2024pixart} may present stereotypes or discriminate against certain groups. For example, the generated images may show an unbalanced gender ratio and inaccurate content of some uncommon concepts. Imagen \cite{saharia2022photorealistic} encodes a variety of social biases and stereotypes, including an overall bias to generate images of people with lighter skin, and images depicting different professions tend to align with Western gender stereotypes. When GLIDE \cite{nichol2021glide} is asked to generate “girl toys, ” it generates more pink toys and stuffed animals than when prompt with “boy toys.” Additionally, when prompted with generic cultural imagery such as “religious sites, ” the model tended to reinforce Western stereotypes.

\textbf{Data-related issues such as copyright, privacy, credibility, and liability}. The large-scale data requirements of T2I models have led researchers to rely heavily on large and mostly uncurated web-scraped datasets, which may contain content without the creator’s consent, may be protected by law, or may contain sensitive personal information. On the other hand, generative AI can synthesize new data that may not always be real, or the model may be unconsciously trained on data that contains factual errors. These issues raise questions about credibility. Some results may even have legal consequences, raising challenge of liability. In addition, training models with uncurated datasets may cause the model’s generative ability to degrade. \cite{alemohammad2023self} points out that repeated use of synthetic data can form an autophagous (“self-consuming”) cycle. Without enough fresh real data in each generation of the cycle, future generative models are destined to gradually decline in quality (precision) or diversity (recall).

\textbf{Economic impact}. Generative AI has accelerated concerns about worker displacement. For example, generative models that generate high-quality images will have an impact on the field of painting. And video generation models like Sora are expected to have a significant impact on many industries such as video production and visual effects.

\subsection{Solutions} \label{Solutions}
DALL-E3 (System Card) filters explicit content (e.g., sex and violence) from the training data by deploying specialized filters for important subcategories (graphic sexualization and hateful imagery). In addition, it also mitigates possible social impacts in terms of input prompts and output images. For example, for prompts related to explicit content, celebrities, or third-party copyrighted or trademarked concepts, characters, or designs, the model mitigates these results by transforming or rejecting the input prompts. On the output side, it uses a classifier to guide the identification of explicit content and guides the model to sample images away from explicit content.

\textbf{Data}. GLIDE \cite{nichol2021glide} filters the training data and uses the data to train a filtered GLIDE to mitigate potential harmful generation.

\textbf{Prompts}. \cite{wu2024universal} transforms the toxic prompts input by user into modified clean prompts through prompt optimization. The appropriate part of the user prompt is retained (i.e., keeping text alignment), while avoiding the generation of harmful content.

\textbf{Deepfake}. In addition to image steganography and watermarking \cite{fridrich2009steganography} that directly embed information into images, some indirect methods \cite{yu2019attributing, yu2021artificial, yu2020responsible} have been proposed to detect and attribute deepfake. \cite{yu2019attributing} points out that different generative models leave unique artifacts (fingerprints) in their generated images, which can be used for image attribute. \cite{yu2021artificial} uses image steganography to embed fingerprints into the training dataset, which is used to train the generative model to embed fingerprints into it. The images generated by the model carry the fingerprints. \cite{yu2020responsible} embeds fingerprints into generative model by using the fingerprint embedding to modulate the convolutional filters of the generator backbone.

\section{Future Directions} \label{Future Directions}
In this section, we explore the direction of improving the generation capability of T2I models based on the existing literature, and look forward to future development directions.

\subsection{Model Scaling and Dataset} \label{Model Scaling and Dataset}
Existing language models often have tens or even hundreds of billions of parameters, while visual models are much smaller in comparison. Intuitively, the more parameters a model has, the larger its capacity and the better its performance. For example, EVA-CLIP-18B \cite{sun2024eva} expands EVA-CLIP \cite{sun2023eva} to 18B, and observes consistent performance improvements as the model size increased. Therefore, expanding the size of visual models is a direction worth exploring in the future.

However, there are many problems when scaling the model size. In addition to the computational cost required for training, large-scale training data is also required. Since, as well know, overfitting will occur if the dataset is too small for complex large-scale models. Some literature \cite{yang2021data, zhao2020differentiable} use data augmentation (training with real data and synthetic data.) to solve the scarcity of training data. However, there are some challenges in training models with synthetic data. \cite{alemohammad2023self} proposes that not having enough fresh real data in each generation of the autophagy cycle will lead to degradation of model performance. This finding is also confirmed by experiments in SynthCLIP \cite{hammoud2024synthclip}. It trains CLIP models using fully synthetic image-text pairs. Compared with models trained on real datasets, its performance is on par (using larger synthetic datasets) or worse (using synthetic datasets of similar size). Therefore, before training, datasets should be carefully collected and curated to ensure that the model does not "go to MAD" \cite{alemohammad2023self} while mitigating overfitting.

\subsection{Prompt Optimization} \label{Prompt Optimization}
Text prompts are very important for high-quality T2I generation. Using longer and finer prompts can usually generate visually better images. \cite{datta2023prompt, lee2024parrot} use prompt expansion \cite{datta2023prompt} to convert coarse prompts into optimized expanded prompts. DALL-E3 \cite{betker2023improving} converts a language model into an image captioner through fine-tuning to generate long captions that are highly descriptive of the image content.

\subsection{Embedding Optimization} \label{Embedding Optimization}
Recent T2I generation models rely on the embedding space of pre-trained models (e.g., CLIP \cite{radford2021learning}). In addition, their objectives are usually related to image embedding alignment and image-text embedding alignment. Therefore, research in this direction is indispensable for improving the performance of image generation models.

\subsection{Model Architecture} \label{Model Architecture}
Some new methods or model architectures are worth to be explored for T2I generation.

In addition to Transformer \cite{vaswani2017attention}, Mamba \cite{gu2023mamba, dao2024transformers}, and Jamba \cite{lieber2024jamba}, which have been mentioned in Section~\ref{Others}, some new methods have been proposed recently to reduce the cost of training models or improve existing models. Transformers can learn to dynamically allocate computations to specific positions in a sequence. MoD Transformer \cite{raposo2024mixture} calculates weights for different tokens through routers to decide which tokens to participate in the calculation of the block (self-attention and MLP) and which tokens to bypass the block, thereby reducing the total computational budget. Similarly, to solve the problem of slow inference speed of autoregressive generative models, Switchable Decision \cite{zhang2024switchable} uses reinforcement learning to dynamically allocate computing resources (automatically decide where to skip and how to balance quality and computational cost) to accelerate the inference process. Infini-Attention \cite{munkhdalai2024leave} introduces a compressed memory into attention mechanism and builds masked local attention and long-term linear attention mechanisms in a single Transformer block. It implements compressed memory by reusing the query, key and value states in the dot product attention calculation. xLSTM \cite{beck2024xlstm} introduces exponential gating and modifies the memory structure of LSTM to extend LSTM, where sLSTM comes with scalar memory, scalar update and new memory hybrid, and fully parallelizable mLSTM with matrix memory and covariance update rule (similar to QKV of Transformer).

\textbf{Idempotent Generative Networks} \cite{shocher2023idempotent} is a model that can generate output in one step and can be seen as a special case of Energy-Based Networks (EBM \cite{zhao2016energy}). Its optimization objective is based on three principles: every sample of the target distribution can be mapped back to itself; the model should be idempotent, that is, applying the mapping twice should produce the same result as applying it once; the subset of instances mapped to itself should be as small as possible.

\textbf{Visual Autoregressive Modeling} \cite{tian2024visual} is a new paradigm that redefines image autoregressive learning. It reimagines autoregressive modeling on images by shifting the generation strategy from the "next label prediction" of traditional image generation to the coarse-to-fine "next scale prediction".

\textbf{DiffEnc} \cite{nielsen2023diffenc} is based on the variational diffusion model and introduces a time-dependent learnable encoder to diffusion process by introducing a data- and depth-dependent mean function, thereby improving the diffusion loss.

\textbf{Quantum Denoising Diffusion Model} \cite{kolle2024quantum} introduces Q-Dense and QU-Net. In addition, it introduces the unitary single-sample quantum consistency model, which achieves one-step image generation by integrating the diffusion process into a unitary matrix.

Perception, memory, and other high-level cognitive functions in the brain are achieved through synchronized oscillatory networks in the brain. Inspired by this, the \textbf{Deep Oscillating Neural Network} \cite{rohan2024deep} uses alternating static dense or convolutional layers (with nonlinear activations) and dynamic oscillator layers. It can be generalized to CNN and is called oscillating convolutional neural network. Both networks can be applied to various benchmark problems in signal and image/video processing.

MLP has a fixed activation function and learnable weights, while \textbf{KAN} \cite{liu2024kan} has no linear weights, but uses a two-layer, learnable activation function composed of splines. Compared with MLP, KAN has better interpretability and requires fewer parameters to achieve the same or better accuracy as MLP. However, KAN is slow to train, and there is still a lot of room for exploration in applications that require a lot of training resources, such as T2I generation.

\subsection{Hybrid Model} \label{Hybrid Model}
In addition to the models \cite{xiao2021tackling, geng2024improving, luo2024you, kang2024distilling} that combine diffusion with GAN in Section~\ref{3Diffusion} and Jamba \cite{lieber2024jamba} that combines Transformer with Mamba \cite{gu2023mamba} and DiS \cite{fei2024scalable} that combines Diffusion with Mamba in Section~\ref{Others}, there are also some literatures that integrates different methods and models to explore their advantages. For example, DiT \cite{peebles2023scalable} that combines Transformer with Diffusion, ImageBART \cite{esser2021imagebart} that combines diffusion and autoregression, and VQ-Diffusion \cite{gu2022vector} that combines diffusion and non-autoregression.

\subsection{Video Generation} \label{Video Generation}
Text-guided generation is a hot topic recently, especially after the emergence of Sora \cite{videoworldsimulators2024}. Therefore, one future research direction is to extend the existing T2I model to text-to-video (T2V) generation. For example, Pix2Gif \cite{kandala2024pix2gif} introduces a motion-guided deformation module, which contains an optical flow network, to InstructPix2Pix \cite{brooks2023instructpix2pix}. It learns to warp the source image feature space to the target image feature space to obtain an image-to-GIF (video) generation model guided by text and motion amplitude prompts. AnimateDiff \cite{guo2023animatediff} adds a plug-and-play motion module to a personalized T2I model to form a personalized animation generator. The motion module learns transferable motion prior knowledge from real-world videos.

\begin{acks}
To Robert, for the bagels and explaining CMYK and color spaces.
\end{acks}

\bibliographystyle{ACM-Reference-Format}
\bibliography{main.bbl}

\appendix

\section{Evaluation} \label{Evaluation}
Evaluation of images is the final but still important step in T2I. In this section, we investigate different evaluation methods. First, we survey human evaluation, which is the evaluation method that best matches human preferences. However, it has some limits: 1) The preferences of people are various, hence, there are biases in the evaluation of the same images; 2) The stopping of model training usually uses the quality of generated images as an indicator, human evaluation is not suitable, and even if human evaluation is used, it is time-consuming and laborious. Therefore, automatic evaluation is needed. We survey the commonly used training and evaluation datasets and automatic evaluation metrics.

\subsection{Human Evaluation} \label{Human Evaluation}
A common setup in human evaluation of T2I generation is: given multiple prompts, two models to be compared are used to generate images pairs conditioning on each prompt, and then the subjects are asked two questions, 1) which image is higher quality? 2) which image aligns better with the prompt? Subjects could answer ‘the first is better’, ‘the second is better’, or ‘not sure’; alternatively, subjects give each image a rating (out of 5). By comparing people's responses, a final human evaluation is obtained.

\subsection{Automatic Evaluation} \label{Automatic Evaluation}
\subsubsection{Datasets} \label{Datasets}
\leavevmode \\
\textbf{MS-COCO (Microsoft Common Objects in Context)} \cite{lin2014microsoft} is a large-scale object detection, segmentation, key point detection and captioning dataset. It contains 91 common object categories, 82 of which have more than 5K labeled instances. It contains 2.5M annotated instances distributed across 328K images. The first version of MS-COCO was released in 2014. It contains 164K images, divided into training set (83K), validation set (41K) and test set (41K). In 2015, the additional test set contains 81K images, including all previous test images and 40K new images. \textbf{COCO2017}. In 2017, the training/validation split was changed from 83K/41K images to 118K/5K images. The new split uses the same images and annotations. The 2017 test set is a subset of the 41K images from the 2015 test set. Additionally, the 2017 release contains a new unannotated dataset containing 123K images. \textbf{COCO-Stuff} \cite{caesar2018coco} extends COCO by providing pixel-wise annotations for 91 stuff categories. The original COCO provides outline-level annotations for 80 thing categories, but it is not enough to understand the content of scenes. COCO-Stuff provides more comprehensive annotations. It includes all 164K images from COCO 2017 (training 118K, validation 5K, test-development 20K, test-challenge 20K), covering 172 categories: 80 thing categories, 91 stuff categories and 1 category "unlabeled".

\textbf{Conceptual Captions (CC3M)} \cite{sharma2018conceptual} is a caption annotations dataset that contains various types of images: natural images, product images, professional photos, comics, paintings, etc. It contains about 3.3M image-description pairs (an order of magnitude more than COCO), of which 28K is used for validation set, and 22.5K is used for test set. Compared to the curated style of COCO images, CC3M's images and original descriptions are collected from the web (using an automated pipeline to extract, filter and transform candidate image/caption pairs) and therefore represent a wider range of style. By relaxing the data collection pipeline used by CC3M, \textbf{CC12M} \cite{changpinyo2021conceptual} is obtained, which contains 12.4M image-text pairs and is specifically for visual-language pre-training. Compared with CC3M, it has a lower token (word count) to type (vocab size) ratio, indicating that it has a longer tail distribution and higher concept diversity. Additionally, the average captions length of CC12M is much longer.

\textbf{Open Images} \cite{OpenImages2, OpenImagesSegmentation, OpenImages} is a dataset containing about 9M images with image-level labels, object bounding boxes, object segmentation masks, visual relationships and Localized Narrative. It can be used for image classification, target detection, visual relationship detection, instance segmentation and multimodal image description.

\textbf{Localized Narrative (LN)} \cite{PontTuset_eccv2020} is a new form of multi-modal image annotations that connects vision and language. It requires the annotator to hover their mouse over the described regions while describing the image using voice. Since voice and mouse pointer are synchronized, voice (the words in descriptions/annotations) corresponds to mouse trajectory (the image regions). It annotates 849K images, including: COCO (123K), Flickr30k \cite{plummer2015flickr30k} and ADE20K \cite{zhou2017scene} datasets, and 671K images from Open Images \cite{OpenImages2}.

\textbf{LAION-5B} \cite{schuhmann2022laion} is a dataset with 5.85M CLIP-filtered image-text pairs for large-scale multi-modal model training. It contains 2.32B English image-text examples (LAION-2B-en/LAION-2B), 2.26B multi-language examples (e.g.: Russian, French, German, Spanish and Chinese, etc.), and 1.27B examples (e.g. place, product, etc.) that are not specific to particular language. Laion High Resolution is a subset of Laion5B with resolution >= 1024$\times$1024, containing 170M samples. According to the aesthetics scores calculated by LAION Aesthetics Predictor V1 \cite{Romain2022laion} and LAION Aesthetics Predictor V2 \cite{Romain2023laion} models, Laion aesthetics dataset and Laion aesthetics dataset V2 are subsets extracted from LAION-5B.

\textbf{COYO-700M} \cite{kakaobrain2022coyo-700m} is an English dataset used to train large-scale foundation models, containing 747M image-text pairs, as well as many other meta-attributes (text and image information), to increased usability of various trained models.

\textbf{T2I-CompBench} \cite{huang2023t2i} is a comprehensive open-world compositional T2I generation benchmark, including 6000 compositional text prompts from 3 categories (attribute binding, object relations, and complex composition) and 6 subcategories (color binding, shape binding, texture binding, non-spatial relationships and complex compositions). \textbf{T2I-CompBench++} is an enhanced T2I-CompBench, including 8000 compositional text prompts from 4 categories (attribute binding, object relations, generative numeracy and complex combination) and 8 subcategories (color binding, shape binding, texture binding, 2D/3D spatial relationships, non-spatial relationships, numeracy and complex compositions).

\subsubsection{Evaluation Metrics} \label{Evaluation Metrics}
\leavevmode \\
\textbf{MSE, PSNR and SSIM} \cite{wang2009mean, hore2010image, wang2004image}. Mean Square Error (MSE) describes the similarity/fidelity of two signals or, conversely, the error/distortion between them. Given two discrete signals of finite length, the MSE formula has the following two forms:
\begin{equation}
    MSE\left( {\mathbf{x},\mathbf{~}\mathbf{y}} \right) = \frac{1}{N}{\sum\limits_{i = 1}^{N}\left( {x_{i} - y_{i}} \right)^{2}}
\end{equation}
\begin{equation}    
    d_{p}\left( {\mathbf{x},\mathbf{~}\mathbf{y}} \right) = {~\left( {\sum\limits_{i = 1}^{N}\left| {x_{i} - y_{i}} \right|^{p}} \right)~}^{1/p}
\end{equation}
where $x_i,y_i$ are sample values  (e.g., pixels) from the two signals and N is the number of samples.

In image processing, MSE is often converted to Peak Signal-to-Noise Ratio (PSNR):
\begin{equation}
    PSNR = 10{\log_{10}\frac{L^{2}}{MSE}}
\end{equation}
where L is the dynamic range of pixel intensity. For example, for a grayscale image with 8 bits/pixel, $L=2^8-1=255$. PSNR is useful if comparing images with different dynamic ranges, otherwise does not contain new information comparing to MSE.

When using MSE to evaluate signal fidelity, there are the following implicit assumptions \cite{wang2009mean}: 1) Fidelity is independent of the temporal or spatial relationship between samples. 2) For a given error, the MSE remains unchanged regardless of which original signal the error is added to. 3) Fidelity is independent of the symbol of the error. 4) All samples are equally important to fidelity. However, when measuring the visual perception of image fidelity, none of these assumptions hold. SSIM \cite{wang2004image} alleviates this problem. SSIM is implemented by modeling any image distortion as a combination of brightness distortion (l), contrast distortion (c), and structural comparison loss (s). SSIM is defined as:
\begin{equation}
    SSIM\left( {\mathbf{x},\mathbf{~}\mathbf{y}} \right) = \left\lbrack l\left( {\mathbf{x},\mathbf{~}\mathbf{y}} \right) \right\rbrack^{\alpha} \cdot \left\lbrack c\left( {\mathbf{x},\mathbf{~}\mathbf{y}} \right) \right\rbrack^{\beta} \cdot \left\lbrack s\left( {\mathbf{x},\mathbf{~}\mathbf{y}} \right) \right\rbrack^{\gamma}
\end{equation}
where
\begin{equation}
    l\left( {\mathbf{x},\mathbf{~}\mathbf{y}} \right) = \frac{2\mu_{x}\mu_{y} + C_{1}}{\mu_{x}^{2} + \mu_{y}^{2} + C_{1}},~c\left( {\mathbf{x},\mathbf{~}\mathbf{y}} \right) = \frac{2\sigma_{x}\sigma_{y} + C_{2}}{\sigma_{x}^{2} + \sigma_{y}^{2} + C_{2}},~s\left( {\mathbf{x},\mathbf{~}\mathbf{y}} \right) = \frac{\sigma_{xy} + C_{3}}{\sigma_{x}\sigma_{y} + C_{3}} \nonumber
\end{equation}
\begin{equation}
    C_{1} = {~\left( K_{1}L \right)~}^{2},~C_{2} = {~\left( K_{2}L \right)~}^{2}  \nonumber
\end{equation}
where $\alpha ,\beta ,\gamma > 0$ are constants used to adjust the importance of the three components. The constant C is used to avoid instability when other values tend to 0. Constants $K_1,K_2 \ll 1$, L is the dynamic range of pixel intensity. In \cite{wang2004image}, setting $\alpha  = \beta  = \gamma  = 1$, $C_3=C_2/2$. $\mu_x,\mu_y$ are the means of signals $\textbf{x,y}$, $\sigma_x,\sigma_y$ are the standard deviations of signals $\textbf{x,y}$, and $\sigma_{xy}$ is the covariance of signals $\textbf{x,y}$.

\textbf{LPIPS}. Classical pixel-by-pixel metrics such as MSE or PSNR are not sufficient for evaluating structured outputs like images because they assume that pixels are independent. To this end, Learned Perceptual Image Patch Similarity (LPIPS) \cite{zhang2018unreasonable} introduces perceptual distance, which is consistent with human judgment. For two patches $x,x_0$, LPIPS is formulated as
\begin{equation}
    d\left( {x,~x_{0}} \right) = {\sum\limits_{l}\frac{1}{H_{l}W_{l}}}{\sum\limits_{h,~w}\left\| {w_{l}\bigodot~\left( {\hat{y}}_{hw}^{l} - {\hat{y}}_{0hw}^{l} \right)~} \right\|_{2}^{2}}
\end{equation}
where $w_l$ is a trainable channel-wise weight, ${\hat{y}}_{hw}^{l},~{\hat{y}}_{0hw}^{l}$ are the features of patch $x,x_0$ extracted from the layer l of a pre-trained network (e.g., VGG).

\textbf{Precision and Recall}. Precision measures how much of the distribution Q can be generated by "parts" of the reference distribution P, while recall measures how much of the reference distribution P can be generated by "parts" of the distribution Q \cite{sajjadi2018assessing}. F1 score is the harmonic mean of precision and recall \cite{lucic2018gans}.

\cite{lucic2018gans} constructs a data manifold and then calculates the distance from the sample to the manifold. However, this method is not applicable to more complex datasets. Instead, \cite{sajjadi2018assessing} defines precision and recall by the relative probability density of two distributions, which is formulated as
\begin{equation}
    P = R = 1 - \delta\left( {P,~Q} \right)
\end{equation}
where $\delta\left( {P,~Q} \right)$ represents the total variation distance between distributions P and Q. However, since PR relies on relative density, it cannot reliably estimate extreme values: for example, it cannot correctly account for cases where a large number of samples are clustered together (as a result of mode collapse or truncation). Therefore, \cite{kynkaanniemi2019improved} maps the manifolds of real and generated data into explicit non-parametric representations and uses the k-nearest neighbor algorithm to improve \cite{sajjadi2018assessing}.

\textbf{R-Precision and CLIP R-Precision}. If there are R relevant documents for a query, and r documents are relevant to the query among the top R retrieval results, then R-Precision is defined as r/R \cite{xu2018attngan}. R-Precision is calculated using the cosine similarity of the embeddings of query image and candidate text extracted by a pre-trained CNN-RNN retrieval model \cite{xu2018attngan}. CLIP R-Precision \cite{park2021benchmark} is calculated by replacing the embedding model with CLIP \cite{radford2021learning}.

\textbf{Semantic Object Accuracy} (SOA) \cite{hinz2020semantic} is used to measure whether the generated image contains the object mentioned in the prompt, including SOA-C (the recall as a class average) and SOA-I (the recall as an image average), formulated as:
\begin{equation}
    SOA\text{-}C = \frac{1}{|C|}{\sum\limits_{c \in C}\frac{1}{\left| I_{c} \right|}}{\sum\limits_{i_{c} \in I_{c}}{Object\text{-}Detector~\left( i_{c} \right)~}}
\end{equation}
\begin{equation}
    SOA\text{-}I = \frac{1}{\sum_{c \in C}\left| I_{c} \right|}{\sum\limits_{c \in C}{\sum\limits_{i_{c} \in I_{c}}{Object\text{-}Detector~\left( i_{c} \right)~}}}      
\end{equation}
\begin{equation}
    Object\text{-}Detector~\left( i_{c} \right)~ = \left\{ \begin{matrix}
    1 \\
    0
    \end{matrix} \right.\begin{matrix}
    {if~an~object~of~class~c~is~detected} \\
    {otherwise}
    \end{matrix} \nonumber    
\end{equation}
where $C$ is the set of object categories; $I_C$ is the set of images belonging to class $c$.

\textbf{Inception Score (IS)} \cite{salimans2016improved} is a metric for evaluating generation quality and diversity, and has a good relevance with human evaluation. It applies Inception model \cite{szegedy2016rethinking} to each generated image to obtain conditional label distribution $p (y|x)$. And the conditional label distribution $p (y|x)$ of images containing meaningful objects should have low entropy. In addition, the marginal distribution $p(y) = {\int{p~\left( y \middle| x = G~(z)~dz \right)~}}$ should have high entropy for generating diverse images. Therefore, the Inception Score can be defined as
\begin{equation}
    {Exp}\left( {\mathbb{E}_{x}KL\left( p\left( y \middle| x \right) \middle| \middle| p~(y)~ \right)} \right)
\end{equation}
The results are indexed so that the values are easier to compare (the larger the IS, the better).

The disadvantage of IS is that it only considers generated images, not real images, so it cannot reflect the gap between generated images and real images. In addition, \cite{barratt2018note} points out two disadvantages of IS: 1) IS is sensitive to small changes in network weights, for example, different implementations of the Inception model \cite{szegedy2016rethinking} (e.g., Keras or Torch) will obtain different IS; 2) Inception model is trained on ImageNet dataset \cite{deng2009imagenet}, and misleading results may be obtained when calculating IS on other datasets. Other metrics (e.g., FID \cite{heusel2017gans}, KID \cite{binkowski2018demystifying}) that use Inception model to obtain features may also face this problem.

\textbf{Fréchet Inception Distance (FID)} \cite{heusel2017gans} is an evaluation metric that improves IS and is used to evaluate the difference between two distributions. As the distribution difference increases, IS fluctuates, remains stable, or even decreases in the worst case, while FID captures the difference level well by monotonically increasing. FID obtains vision-related features from the encoding layer of inception model \cite{szegedy2016rethinking} and assumes that these features follow a multi-dimensional Gaussian distribution. The difference between two Gaussian distributions (generated image and real image) is measured by Fréchet Distance \cite{frechet1957distance}, also known as Wasserstein-2 Distance \cite{vaserstein1969markov}, and is formulated as:
\begin{equation}
    d^{2}\left( {\left( {m,~C} \right),~\left( {m_{w},~C_{w}} \right)} \right) = \left\| {m - m_{w}} \right\|_{2}^{2} + Tr\left( {C + C_{w} - 2\left( {CC_{W}} \right)^{\frac{1}{2}}} \right)
\end{equation}
where the mean $m$ and covariance $C$ of the distribution are obtained from inception model.

The smaller the FID, the more two distributions overlap. When FID=0, two distributions completely overlap. Note that, since there is an operation of taking the real part of a complex value when practically calculating the covariance matrix \cite{Seitzer2020FID}, one possible case is that the FID value for two completely overlapping distributions (such as real images and real images) may not be equal to 0, but a negative value. 

The disadvantage of FID is that FID is sensitive to the increase of spurious modes and mode dropping \cite{lucic2018gans}. In addition, it assumes that data follows multi-dimensional Gaussian distribution, which may not hold.

\textbf{sFID} \cite{nash2021generating} is equivalent to FID \cite{heusel2017gans}, but uses the intermediate spatial features in Inception model instead of the spatial pooling features used in the standard FID.

\textbf{Kernel Inception Distance (KID)} \cite{binkowski2018demystifying} is a metric similar to FID, which is the squared maximum mean difference (MMD) between Inception representations \cite{szegedy2016rethinking}. It uses a kernel function
\begin{equation}
    K\left( {x,~y} \right) = k\left( {\varphi(x),~\varphi(y)} \right)
\end{equation}
\begin{equation}
    k\left( {x,~y} \right) = \left( {\frac{1}{d}x^{T}y + 1} \right)^{3} \nonumber
\end{equation}
where $k$ is a polynomial kernel, $\varphi$ is a function that maps an image to an Inception representation, and d is the dimension of the representation. KID is unbiased compared to FID and is more consistent with human perception. However, it has a very high Spearman rank-order \cite{kurach2018gan} correlation with FID.

\textbf{CLIP-I, CLIPScore (CLIP-T) and RefCLIPScore}. CLIP-I is the average cosine similarity between the CLIP embeddings of generated images and real images, which is used to measure their alignment. CLIP-T is the average cosine similarity between the CLIP text embedding of prompts and the CLIP embedding of the corresponding generated image, referred as CLIPScore \cite{hessel2021clipscore}, which is used to measure the alignment of the generated images with the corresponding prompts.
\begin{equation}
    CLIPScore\left( {\mathbf{c},~\mathbf{v}} \right) = w*{\max\left( {{\cos\left( {\mathbf{c},~\mathbf{v}} \right)},~0} \right)}
\end{equation}
where $w = 2.5$, $c$ and $v$ are CLIP embeddings of prompts and images, respectively. CLIPScore is a reference-free metric, while RefCLIPScore \cite{hessel2021clipscore} is used under the condition of reference, which is calculated as the harmonic mean of CLIPScore and the maximal reference cosine similarity, that is, 
\begin{equation}
    RefCLIPScore = H\text{-}Mean\left( {CLIPScore\left( {\mathbf{c},~\mathbf{v}} \right),~{\max\limits_{r \in R}{~\left( {\cos\left( {\mathbf{c},~\mathbf{r}} \right)},~0 \right)~}}} \right)
\end{equation}
where R is the set of CLIP text embeddings of the reference prompts.

\textbf{DINO metric} \cite{ruiz2023dreambooth} is the average pairwise cosine similarity between the ViT-S/16 DINO embeddings of the generated and real images.

The Self-DIstillation with NO labels (DINO) network \cite{caron2021emerging} is a self-supervised network consisting of a backbone network (ViT or ResNet) and a projection head, and the features used in downstream tasks are the outputs of the backbone. The self-supervised ViT features explicitly include the scene layout, especially object boundaries, which are directly accessible in the self-attention module of the last block. Compared with supervised networks or convolutional networks, DINO network does not ignore the differences between objects of the same class during training. Instead, its self-supervised training objective encourages unique features that distinguish objects or images. The DINOv2 model \cite{oquab2023dinov2} uses a discriminative self-supervised staged training method to extract powerful image features by optimizing image-level and patch-level objectives respectively, without fine-tuning on downstream tasks.

\textbf{LAION Aesthetics Score} is calculated by LAION Aesthetics Predictor V1 \cite{Romain2022laion} or LAION Aesthetics Predictor V2 \cite{Romain2023laion}, where the input image is first converted to an embedding by Open CLIP \cite{cherti2023reproducible} ViT-L/14. LAION Aesthetics Predictor V2 adds five MLP layers on top of (frozen) Open CLIP ViT-L/14 and fine-tunes these MLP layers with a large number of images using only regression loss terms (such as MSE and MAE).

\textbf{Human Preference Score (HPS)} \cite{wu2023human} is a metric aligned with human preferences. It first collects a dataset reflecting human preferences (HPD) through the Stable Foundation Discord. Then CLIP ViT-L/14 is fine-tuned on this dataset to obtain a Human Preference Classifier (HPC) that can calculate HPS. HPS is calculated as
\begin{equation}
    HPS\left( {img,~txt} \right) = 100{\cos{~\left( {enc}_{v}\left( {img} \right),~{enc}_{t}~(txt)~ \right)~}}
\end{equation}
where $enc_v,enc_t$ are the visual encoder and text encoder of HPC respectively. Compared with CLIPScore, HPS is more aligned with human preferences. In addition, compared with metrics such as IS, FID and KID (based on Inception networks pre-trained on the ImageNet dataset), HPS has better generalization ability.

The HPC of HPSv2 \cite{wu2023humanv2} is trained on HPDv2. Compared with HPD, HPDv2 contains more data (98, 807 images generated from 25,205 prompts vs 798k image-prompt pairs containing 434k images). In addition, HPDv2 is collected from contract workers, while HPD is collected from Discord channels, showing a strong bias for certain styles.

\section{Comparison} \label{Comparison}
{
\scriptsize
\begin{longtable}[!ht]{ccccccccc}
    \caption{Comparison of Models} \\    
        \hline
        \makecell{First submission \\ on Arxiv} & Publish & Model & Text Encoder & Image Encoder & Attention & Masking & \makecell{Class-free \\ Guidance} & Source \\
        \hline
        \endhead
        \hline
        \endfoot
        
        \multicolumn{9}{c}{\textbf{Autoregression}} \\
        2021.02.24 & 21'ICML & DALL-E \cite{ramesh2021zero} & BPE & dVAE & \checkmark & \checkmark & ~ & \href{https://github.com/openai/DALL-E/tree/master}{Github} \\ 
        2021.05.01 & 21'ACM SIGKDD & M6 \cite{lin2021m6} & WordPiece & VQ-GAN & \checkmark & \checkmark & ~ & ~ \\ 
        2021.05.26 & 21'NIPS & CogView \cite{ding2021cogview} & SentencePiece & VQ-VAE & \checkmark & \checkmark & ~ & \href{https://github.com/THUDM/CogView}{Github} \\ 
        2021.08.19 & 21'NIPS & ImageBART \cite{esser2021imagebart} & CLIP & CLIP & \checkmark & \checkmark & ~ & \href{https://github.com/CompVis/imagebart}{Github} \\ 
        2021.11.24 & 22'ECCV & NÜWA \cite{wu2022nuwa} & BPE & VQ-VAE & \checkmark & \checkmark & ~ & \href{https://github.com/microsoft/NUWA}{Github} \\ 
        2021.12.31 & ~ & Ernie-ViLG \cite{zhang2021ernie} & ~ & VQ-VAE & \checkmark & \checkmark & ~ & \href{https://github.com/PaddlePaddle/ERNIE/tree/repro/ernie-vil}{Github} \\ 
        2022.07.22 & 22'TMLR & Parti \cite{yu2022scaling} & ~ & ViT-VQGAN & \checkmark & \checkmark & \checkmark & \href{https://github.com/google-research/parti}{Github} \\ 
        
        \multicolumn{9}{c}{\textbf{Autoregression Editing}} \\
        2022.03.23 & 22'ECCV & Make-A-Scene \cite{gafni2024scene} & BPE & VQ-SEG, VQ-IMG & \checkmark & ~ & \checkmark & \href{https://github.com/CasualGANPapers/Make-A-Scene}{Github} \\ 
        
        \multicolumn{9}{c}{\textbf{Non-Autoregression}} \\ 
        2022.02.08 & 22'CVPR & MaskGIT \cite{chang2022maskgit} & ~ & VQ-VAE & \checkmark & \checkmark & ~ & \href{https://github.com/google-research/maskgit}{Github} \\ 
        2022.04.28 & 22'NIPS & CogView2 \cite{ding2022cogview2} & icetk & VQ-VAE & \checkmark & \checkmark & ~ & \href{https://github.com/THUDM/CogView2}{Github} \\ 
        2023.01.02 & 23'ICML & Muse \cite{chang2023muse} & T5-XXL & VQ-GAN & \checkmark & \checkmark & \checkmark & \href{https://github.com/lucidrains/muse-maskgit-pytorch}{Github} \\ 
        2023.12.22 & ~ & Emage \cite{feng2023emage} & BERT, CLIP & VQ-GAN & \checkmark & \checkmark & ~ & ~ \\ 
        2024.01.03 & ~ & aMUSEd \cite{patil2024amused} & CLIP-L/14 & VQ-GAN & \checkmark & \checkmark & \checkmark & \href{https://github.com/huggingface/amused}{Github} \\ 
        
        \multicolumn{9}{c}{\textbf{Non-Autoregression Editing}} \\ 
        2023.06.01 & 23'NIPS & StyleDrop \cite{sohn2023styledrop} & T5-XXL & VQ-GAN & \checkmark & \checkmark & \checkmark & \href{https://styledrop.github.io/}{Page} \\ 
        
        \multicolumn{9}{c}{\textbf{GAN}} \\
          & 21'IEEE TIP & MA-GAN \cite{yang2021multi} & ~ & ~ & \checkmark & ~ & ~ & ~ \\ 
        2020.08.13 & 21'CVPR & DF-GAN \cite{tao2022df} & ~ & ~ & ~ & ~ & ~ & \href{https://github.com/tobran/DF-GAN}{Github} \\ 
        2020.11.07 & 21'WACV & TReCS \cite{koh2021text} & ~ & ~ & ~ & \checkmark & ~ & \href{https://github.com/google-research/trecs_image_generation}{Github} \\ 
        2021.08.27 & 21'ICCV & DAE-GAN \cite{ruan2021dae} & LSTM & Inception-v3 & \checkmark & ~ & ~ & ~ \\ 
        2021.01.12 & 21'CVPR & XMC-GAN \cite{zhang2021cross} & ~ & VGG & \checkmark & ~ & ~ & \href{https://github.com/google-research/xmcgan_image_generation}{Github} \\ 
        2021.11.27 & 22'CVPR & LAFITE \cite{zhang2021lafite} & CLIP & CLIP & ~ & ~ & ~ & \href{https://github.com/drboog/Lafite}{Github} \\ 
        2021.12.02 & ~ & FuseDream \cite{liu2021fusedream} & CLIP & CLIP & ~ & ~ & ~ & \href{https://github.com/gnobitab/FuseDream}{Github} \\ 
        2022.04.18 & 22'ECCV & VQGAN-CLIP \cite{crowson2022vqgan} & CLIP & CLIP & ~ & ~ & ~ & \href{https://github.com/EleutherAI/vqgan-clip/tree/main/notebooks}{Github} \\ 
        2023.01.23 & 23'ICML & StyleGAN-T \cite{sauer2023stylegan} & CLIP & CLIP & \checkmark & ~ & ~ & \href{https://github.com/autonomousvision/stylegan-t}{Github} \\ 
        2023.03.09 & 23'CVPR & GigaGAN \cite{kang2023scaling} & CLIP ViT-L/14 & CLIP & \checkmark & ~ & ~ & \href{https://github.com/lucidrains/gigagan-pytorch}{Github} \\ 
        
        \multicolumn{9}{c}{\textbf{GAN Editing}} \\
        2021.08.02 & 22'ACM TOG & StyleGAN-NADA \cite{gal2022stylegan} & CLIP & CLIP & ~ & ~ & ~ & \href{https://stylegan-nada.github.io/}{Github} \\ 
        2022.04.05 & 22'ECCV & Text2LIVE \cite{bar2022text2live} & CLIP & CLIP & \checkmark & ~ & ~ & \href{https://text2live.github.io/}{Github} \\ 
        2024.01.18 & ~ & Edit One for All \cite{nguyen2024edit} & ~ & ~ & ~ & ~ & ~ & \href{https://thaoshibe.github.io/edit-one-for-all/}{Github} \\ 
        
        \multicolumn{9}{c}{\textbf{Diffusion}} \\
        2021.12.20 & 22'CVPR & LDM (SD) \cite{rombach2022high} & BERT & VAE & \checkmark & \checkmark & \checkmark & \href{https://github.com/CompVis/latent-diffusion}{Github} \\ 
        2023.06.04 & 24'ICLR & SDXL \cite{podell2023sdxl} & OpenCLIP & VAE & \checkmark & ~ & \checkmark & \href{https://github.com/Stability-AI/generative-models}{Github} \\ 
        2023.11.28 & ~ & SDXL-Turbo \cite{sauer2023adversarial} & CLIP-ViT-g-14 & DINOv2 ViT-L & \checkmark & ~ & ~ & \href{https://github.com/Stability-AI/generative-models}{Github} \\ 
        2024.02.05 & 24‘ICML’ & SD3 \cite{esser2024scaling} & ~ & ~ & \checkmark & ~ & ~ & \href{https://stability.ai/news/stable-diffusion-3}{Page} \\ 
        2022.05.23 & 22'NIPS & Imagen \cite{saharia2022photorealistic} & T5-XXL & ~ & \checkmark & ~ & \checkmark & \href{https://imagen.research.google/}{Page} \\ 
        2022.09.29 & 23'ICLR & Re-Imagen \cite{chen2022re} & ~ & ~ & \checkmark & ~ & \checkmark & ~ \\ 
        2021.11.29 & 22'CVPR & VQ-Diffusion \cite{gu2022vector} & CLIP & VQ-VAE & \checkmark & \checkmark & ~ & \href{https://github.com/cientgu/VQ-Diffusion}{Github} \\ 
        2021.12.20 & 22'ICML & Glide \cite{nichol2021glide} & ~ & ~ & \checkmark & \checkmark & \checkmark & \href{https://github.com/openai/glide-text2im}{Github} \\ 
        2022.04.06 & 22'ICLR & kNN-Diffisuion \cite{sheynin2022knn} & CLIP & CLIP & \checkmark & ~ & \checkmark & ~ \\ 
        2022.04.13 & ~ & DALL-E2 (unCLIP) \cite{ramesh2022hierarchical} & CLIP & CLIP & \checkmark & \checkmark & \checkmark & \href{https://github.com/lucidrains/DALLE2-pytorch}{Github} \\ 
        2022.08.29 & 23'AAAI & Frido \cite{fan2023frido} & BERT & MS-VQGAN & \checkmark & ~ & \checkmark & \href{https://github.com/davidhalladay/Frido}{Github} \\ 
        2022.10.27 & 23'CVPR & Ernie-ViLG 2.0 \cite{feng2023ernie} & transformer & ~ & \checkmark & ~ & \checkmark & \href{https://wenxin.baidu.com/ernie-vilg}{Page} \\ 
        2022.11.02 & ~ & eDiff-I \cite{balaji2022ediff} & T5, CLIP & CLIP & \checkmark & \checkmark & \checkmark & \href{https://deepimagination.cc/eDiff-I/}{Page} \\ 
        2022.11.24 & 23'CVPR & Corgi \cite{zhou2023shifted} & CLIP & CLIP & \checkmark & ~ & \checkmark & \href{https://github.com/drboog/Shifted_Diffusion}{Github} \\ 
        2023.05.29 & 23'NIPS & RAPHAEL \cite{xue2024raphael} & OpenCLIP & VAE & \checkmark & \checkmark & \checkmark & ~ \\ 
        2023.06.01 & 24'ICLR & Würstchen \cite{pernias2023wurstchen} & CLIP & VQGAN & \checkmark & ~ & \checkmark & \href{https://github.com/dome272/Wuerstchen}{Github} \\ 
        2023.09.12 & 24'ICLR & InstaFlow \cite{liu2023instaflow} & CLIP ViT-L/14 & AE & ~ & ~ & ~ & \href{https://github.com/gnobitab/InstaFlow}{Github} \\ 
        2023.09.20 & ~ & FreeU \cite{si2024freeu} & ~ & ~ & ~ & ~ & ~ & \href{https://chenyangsi.top/FreeU/}{Page} \\ 
        2023.09.30 & 24'ICLR & PixArt-$\alpha$ \cite{chen2023pixart} & T5 & VAE & \checkmark & ~ & \checkmark & \href{https://pixart-alpha.github.io/}{Github} \\ 
         & ~ & DALL-E3 \cite{betker2023improving} & CLIP & VAE & \checkmark & ~ & ~ & \href{https://openai.com/index/dall-e-3/}{Page} \\ 
        2023.10.03 & 24'CVPR & Predicated Diffusion \cite{sueyoshi2024predicated} & ~ & ~ & \checkmark & ~ & \checkmark & ~ \\ 
        2023.11.14 & 24'CVPR & UFOGen \cite{xu2024ufogen} & CLIP ViT-L/14 & VAE & ~ & ~ & ~ & ~ \\ 
        2023.12.04 & 24'CVPR & Ten \cite{wang2024generative} & ~ & ~ & \checkmark & \checkmark & \checkmark & \href{https://powers-of-10.github.io/}{Github} \\ 
        2023.12.08 & 24'CVPR & SwiftBrush \cite{nguyen2024swiftbrush} & ~ & VAE & ~ & ~ & \checkmark & \\ 
        2023.12.18 & 24'CVPR & SCEdit \cite{jiang2024scedit} & ~ & ~ & ~ & \checkmark & ~ & \href{https://scedit.github.io/}{Github} \\ 
        2023.12.27 & ~ & PanGu-Draw \cite{lu2023pangu} & ~ & VAE & \checkmark & ~ & \checkmark & \href{https://pangu-draw.github.io/}{Github} \\ 
        2023.12.27 & ~ & Prompt Expansion \cite{datta2023prompt} & ~ & ~ & ~ & ~ & ~ & ~ \\ 
        2023.12.30 & ~ & Self-Perception \cite{lin2023diffusion} & ~ & ~ & ~ & ~ & \checkmark & ~ \\ 
        2024.01.04 & 23'NIPS & ConPreDiff \cite{yang2024improving} & T5, CLIP & ~ & \checkmark & \checkmark & ~ & ~ \\ 
        2024.01.11 & ~ & Parrot \cite{lee2024parrot} & CLIP & CLIP ViT-B & \checkmark & ~ & \checkmark & ~ \\ 
        2024.01.20 & ~ & RL-Diffusion \cite{zhang2024large} & ~ & ~ & ~ & ~ & \checkmark & \href{https://pinterest.github.io/atg-research/rl-diffusion/}{Github} \\ 
        2024.01.25 & ~ & SceneGraph2Image \cite{mishra2024scene} & CLIP & VQGAN & ~ & ~ & ~ & ~ \\ 
        2024.02.07 & ~ & Text2Street \cite{su2024text2street} & CLIP & ~ & \checkmark & \checkmark & ~ & ~ \\ 
        2024.02.16 & ~ & Make a Cheap Scaling \cite{guo2024make} & CLIP & ~ & \checkmark & ~ & ~ & \href{https://github.com/GuoLanqing/Self-Cascade/}{Github} \\ 
        2024.02.26 & ~ & Multi-LoRA Composition \cite{zhong2024multi} & ~ & ~ & \checkmark & ~ & ~ & \href{https://maszhongming.github.io/Multi-LoRA-Composition/}{Github} \\ 
        2024.03.04 & 24'CVPR & HanDiffuser \cite{narasimhaswamy2024handiffuser} & CLIP-ViT-B/32 & ~ & \checkmark & ~ & \checkmark & ~ \\ 
        2024.03.08 & ~ & CogView3 \cite{zheng2024cogview3} & T5-XXL & VAE & ~ & ~ & \checkmark & ~ \\ 
        2024.03.07 & ~ & PixArt-$\Sigma$ \cite{chen2024pixart} & Flan-T5-XXL & VAE & \checkmark & ~ & ~ & \href{https://pixart-alpha.github.io/PixArt-sigma-project/}{Github} \\ 
        2024.03.15 & ~ & Giving a Hand to DM \cite{pelykh2024giving} & CLIP & VAE & \checkmark & \checkmark & ~ & \href{https://github.com/apelykh/hand-to-diffusion}{Github} \\ 
        2024.03.19 & ~ & FouriScale \cite{huang2024fouriscale} & ~ & ~ & \checkmark & \checkmark & \checkmark & \href{https://github.com/LeonHLJ/FouriScale}{Github} \\ 
        2024.03.19 & ~ & YOSO \cite{luo2024you} & ~ & VAE & ~ & ~ & ~ & \href{https://github.com/Luo-Yihong/YOSO}{Github} \\ 
        2024.03.25 & ~ & RLCM \cite{oertell2024rl} & ~ & CLIP & ~ & ~ & ~ & \href{https://rlcm.owenoertell.com/}{Page} \\ 
        2024.04.01 & 24'CVPR & CosmicMan \cite{li2024cosmicman} & CLIP & AE & \checkmark & \checkmark & ~ & \href{https://cosmicman-cvpr2024.github.io/}{Github} \\ 
        2024.04.03 & ~ & T-GATE \cite{zhang2024cross} & ~ & ~ & \checkmark & ~ & \checkmark & \href{https://github.com/HaozheLiu-ST/T-GATE}{Github} \\ 
        2024.05.09 & ~ & Diffusion2GAN \cite{kang2024distilling} & ~ & VAE & \checkmark & ~ & \checkmark & \href{https://mingukkang.github.io/Diffusion2GAN/}{Github} \\ 
        
        \multicolumn{9}{c}{\textbf{Diffusion Editing}} \\ 
        2021.08.02 & 21'ICLR & SDEdit \cite{meng2021sdedit} & ~ & ~ & ~ & \checkmark & ~ & \href{https://github.com/ermongroup/SDEdit}{Github} \\ 
        2021.10.06 & 22'CVPR & DiffusionCLIP \cite{kim2022diffusionclip} & CLIP & CLIP & \checkmark & ~ & ~ & \href{https://github.com/gwang-kim/DiffusionCLIP}{Github} \\ 
        2021.11.29 & 22'CVPR & Blended Diffusion \cite{avrahami2022blended} & CLIP & CLIP & ~ & \checkmark & ~ & \href{https://omriavrahami.com/blended-diffusion-page/}{Page} \\ 
        2022.08.02 & 23'ICLR & Prompt-to-Prompt \cite{hertz2022prompt} & ~ & ~ & \checkmark & ~ & \checkmark & \href{https://github.com/google/prompt-to-prompt}{Github} \\ 
        2022.08.02 & 23'ICLR & Textual Inversion \cite{gal2022image} & BERT & ~ & ~ & ~ & ~ & \href{https://textual-inversion.github.io/}{Github} \\ 
        2022.08.25 & 23'CVPR & DreamBooth \cite{ruiz2023dreambooth} & ~ & ~ & ~ & ~ & ~ & \href{https://dreambooth.github.io/}{Github} \\ 
        2022.10.17 & 23'ACM TOG & UniTune \cite{valevski2023unitune} & T5-XXL & ~ & \checkmark & ~ & \checkmark & ~ \\ 
        2022.10.17 & 23'CVPR & Imagic \cite{kawar2023imagic} & ~ & AE & \checkmark & ~ & \checkmark & \href{https://imagic-editing.github.io/}{Github} \\ 
        2022.10.20 & 23'ICLR & DiffEdit \cite{couairon2023diffedit} & ~ & DDIM & ~ & \checkmark & \checkmark & \href{https://github.com/huggingface/diffusers}{Github} \\ 
        2022.10.28 & ~ & MagicMix \cite{liew2022magicmix} & ~ & AE & \checkmark & ~ & ~ & \href{https://magicmix.github.io/}{Github} \\ 
        2022.11.17 & 23'CVPR & InstructPix2Pix \cite{brooks2023instructpix2pix} & ~ & VAE & \checkmark & ~ & \checkmark & \href{https://www.timothybrooks.com/instruct-pix2pix}{Page} \\ 
        2022.11.17 & 23'CVPR & Null-text Inversion \cite{mokady2023null} & ~ & VQ-AE & \checkmark & ~ & \checkmark & \href{https://null-text-inversion.github.io/}{Github} \\ 
        2022.11.19 & 24'IEEE TNNLS & DiffStyler \cite{huang2024diffstyler} & CLIP & CLIP & \checkmark & ~ & ~ & \href{https://github.com/haha-lisa/Diffstyler}{Github} \\ 
        2022.11.22 & 23'CVPR & Plug-and-Play Diffusion \cite{tumanyan2023plug} & ~ & ~ & \checkmark & \checkmark & \checkmark & \href{https://pnp-diffusion.github.io/}{Github} \\ 
        2022.11.23 & 23'CVPR & InST \cite{zhang2023inversion} & ~ & CLIP & \checkmark & ~ & ~ & \href{https://github.com/zyxElsa/InST}{Github} \\ 
        2022.12.08 & 23'CVPR & Custom Diffusion \cite{kumari2023multi} & CLIP & ~ & \checkmark & ~ & \checkmark & \href{https://www.cs.cmu.edu/~custom-diffusion/}{Page} \\ 
        2022.12.09 & 23'CVPR & SmartBrush \cite{xie2023smartbrush} & ~ & ~ & ~ & \checkmark & \checkmark & ~ \\ 
        2022.12.13 & 23'CVPR & Imagen Editor \cite{wang2023imagen} & T5-XXL & ~ & \checkmark & \checkmark & \checkmark & \href{https://imagen.research.google/editor/}{Page} \\ 
        2023.02.10 & 23'ICCV & ControlNet \cite{zhang2023adding} & CLIP & ~ & \checkmark & ~ & \checkmark & \href{https://github.com/lllyasviel/ControlNet}{Github} \\ 
        2023.02.16 & AAAI & T2I-Adapter \cite{mou2024t2i} & CLIP & AE & ~ & ~ & ~ & \href{https://github.com/TencentARC/T2I-Adapter}{Github} \\ 
        2023.02.20 & 23'ICML & Composer \cite{huang2023composer} & CLIP ViT-L/14 & CLIP ViT-L/14 & \checkmark & \checkmark & \checkmark & \href{https://github.com/ali-vilab/composer}{Github} \\ 
        2023.02.27 & 23'ICCV & ELITE \cite{wei2023elite} & CLIP & CLIP & \checkmark & \checkmark & \checkmark & \href{https://github.com/csyxwei/ELITE}{Github} \\ 
        2023.03.09 & 23'ICML & Cones \cite{liu2023cones} & CLIP & CLIP & \checkmark & \checkmark & ~ & ~ \\ 
        2023.04.01 & 23'NIPS & SuTI \cite{chen2024subject} & ~ & ~ & \checkmark & ~ & \checkmark & \href{https://open-vision-language.github.io/suti/}{Page} \\ 
        2023.04.06 & CVPR & InstantBooth \cite{shi2024instantbooth} & CLIP & CLIP & \checkmark & \checkmark & ~ & \\ 
        2023.04.17 & 23'ICCV & MasaCtrl \cite{cao2023masactrl} & ~ & AE & \checkmark & \checkmark & \checkmark & \href{https://github.com/TencentARC/MasaCtrl}{Github} \\ 
        2023.05.04 & 23'NIPS & BLIP-Diffusion \cite{li2024blip} & BLIP-2 & CLIP ViT-L/14 & \checkmark & \checkmark & ~ & \href{https://github.com/salesforce/LAVIS/tree/main/projects/blip-diffusion}{Github} \\ 
        2023.05.18 & 23'NIPS & UniControl \cite{qin2023unicontrol} & CLIP & ~ & \checkmark & \checkmark & \checkmark & \href{https://github.com/salesforce/UniControl}{Github} \\ 
        2023.05.25 & 23'ACM TOG & ProSpect \cite{zhang2023prospect} & CLIP & ~ & \checkmark & ~ & ~ & \href{https://github.com/zyxElsa/ProSpect}{Github} \\ 
        2023.06.01 & ~ & Semantic-Visual Alignment \cite{abreu2023addressing} & ~ & ~ & ~ & ~ & ~ & ~ \\ 
        2023.06.01 & 23'NIPS & Self-Guidance \cite{epstein2023diffusion} & ~ & ~ & \checkmark & \checkmark & \checkmark & \href{https://dave.ml/selfguidance/}{Page} \\ 
        2023.08.13 & ~ & IP-Adapter \cite{ye2023ip} & CLIP & CLIP & \checkmark & ~ & \checkmark & \href{https://ip-adapter.github.io/}{Github} \\ 
        2023.08.31 & 23'CVPR & Text2Scene \cite{hwang2023text2scene} & CLIP & CLIP & ~ & ~ & ~ & ~ \\ 
        2023.11.09 & 23'ACM MM & ControlStyle \cite{chen2023controlstyle} & ~ & VAE & \checkmark & ~ & \checkmark & ~ \\ 
        2023.11.28 & 24'CVPR & Ranni \cite{feng2024ranni} & CLIP & ~ & \checkmark & ~ & ~ & \href{https://ranni-t2i.github.io/Ranni/}{Github} \\ 
        2023.12.04 & 24'CVPR & StyleAligned \cite{hertz2024style} & ~ & ~ & \checkmark & ~ & ~ & \href{https://github.com/google/style-aligned}{Github} \\ 
        2023.12.06 & ~ & Make-A-Storyboard \cite{su2023make} & ~ & VAE & \checkmark & \checkmark & ~ & ~ \\ 
        2023.12.07 & 24'CVPR & PhotoMaker \cite{li2024photomaker} & CLIP & CLIP & \checkmark & \checkmark & \checkmark & \href{https://photo-maker.github.io/}{Github} \\ 
        2023.12.11 & ~ & ControlNet-XS \cite{zavadski2023controlnet} & ~ & ~ & \checkmark & ~ & \checkmark & \href{https://github.com/vislearn/ControlNet-XS}{Github} \\ 
        2023.12.21 & ~ & DreamDistribution \cite{zhao2023dreamdistribution} & CLIP & AE & ~ & ~ & ~ & \href{https://briannlongzhao.github.io/DreamDistribution/}{Github} \\ 
        2023.12.26 & 24'CVPR & One-Dimensional Adapter \cite{lyu2024one} & CLIP & ~ & ~ & ~ & \checkmark & \href{https://lyumengyao.github.io/projects/spm}{Github} \\ 
        2023.12.28 & 24'CVPR & ZONE \cite{li2024zone} & CLIP & AE & \checkmark & \checkmark & \checkmark & ~ \\ 
        2024.01.11 & ~ & PALP \cite{arar2024palp} & ~ & ~ & \checkmark & ~ & \checkmark & \href{https://prompt-aligned.github.io/}{Github} \\ 
        2024.01.15 & 24'AAAI & InstDiffEdit \cite{zou2024towards} & CLIP & ~ & \checkmark & \checkmark & \checkmark & \href{https://github.com/xiaotianqing/InstDiffEdit}{Github} \\ 
        2024.01.18 & ~ & WaveOpt-Estimator \cite{koo2024wavelet} & ~ & ~ & \checkmark & ~ & ~ & ~ \\ 
        2024.01.25 & ~ & BootPIG \cite{purushwalkam2024bootpig} & ~ & VAE & \checkmark & \checkmark & \checkmark & \href{https://github.com/SalesforceAIResearch/bootpig}{Github} \\ 
        2024.01.25 & ~ & CreativeSynth \cite{huang2024creativesynth} & ~ & VAE & \checkmark & ~ & ~ & \href{https://github.com/haha-lisa/CreativeSynth}{Github} \\ 
        2024.01.28 & ~ & Object-Driven One-Shot FT \cite{lu2024object} & CLIP & CLIP & \checkmark & \checkmark & ~ & ~ \\ 
        2024.01.28 & ~ & FreeStyle \cite{he2024freestyle} & ~ & ~ & \checkmark & ~ & ~ & \href{https://github.com/FreeStyleFreeLunch/FreeStyle}{Github} \\ 
        2024.01.30 & ~ & Pick-and-Draw \cite{lv2024pick} & ~ & ~ & \checkmark & \checkmark & ~ & ~ \\ 
        2024.01.31 & ~ & SeFi-IDE \cite{li2024sefi} & CLIP & AutoVAE & \checkmark & ~ & ~ & \href{https://com-vis.github.io/SeFi-IDE/}{Github} \\ 
        2024.02.03 & ~ & SepME \cite{zhao2024separable} & ~ & ~ & \checkmark & ~ & ~ & \href{https://github.com/Dlut-lab-zmn/SRS-ME}{Github} \\ 
        2024.02.08 & 24'CVPR & MIGC \cite{zhou2024migc} & CLIP & ~ & \checkmark & \checkmark & \checkmark & \href{https://migcproject.github.io/}{Github} \\ 
        2024.02.27 & ~ & LayerDiffusion \cite{zhang2024transparent} & ~ & VAE & \checkmark & ~ & ~ & ~ \\ 
        2024.02.08 & 24'ICLR & Get What You Want \cite{li2024get} & CLIP & ~ & \checkmark & ~ & ~ & \href{https://github.com/sen-mao/SuppressEOT}{Github} \\ 
        2024.02.16 & ~ & CtrlColor \cite{liang2024control} & ~ & CLIP & \checkmark & \checkmark & ~ & \href{https://zhexinliang.github.io/Control_Color/}{Github} \\ 
        2024.03.11 & 24'CVPR & DEADiff \cite{qi2024deadiff} & ~ & CLIP ViT-L/14 & \checkmark & ~ & \checkmark & \href{https://tianhao-qi.github.io/DEADiff/}{Github} \\ 
        2024.03.25 & ~ & Attribute-Control \cite{baumann2024continuous} & CLIP & CLIP & ~ & ~ & \checkmark & \href{https://compvis.github.io/attribute-control/}{Github} \\ 
        2024.03.27 & 24'CVPR & DisenDiff \cite{zhang2024attention} & CLIP & VAE & \checkmark & \checkmark & ~ & \href{https://github.com/Monalissaa/DisenDiff}{Github} \\ 
        2024.04.17 & ~ & Factorized Diffusion \cite{geng2024factorized} & CLIP & CLIP & ~ & \checkmark & ~ & \href{https://dangeng.github.io/factorized_diffusion/}{Github} \\ 
        
        \multicolumn{9}{c}{\textbf{Energy-Based Mode}} \\
        2016.11.30 & 17'CVPR & PPGN \cite{nguyen2017plug} & ~ & DAE & ~ & \checkmark & ~ & \\ 
        
        \multicolumn{9}{c}{\textbf{Mamba}} \\
        2024.03.20 & ~ & ZigMa \cite{hu2024zigma} & CLIP & VAE & ~ & ~ & ~ & \href{https://taohu.me/zigma/}{Page} \\ 
        \multicolumn{9}{c}{\textbf{Multi-modal}} \\
        2022.11.15 & 23'ICCV & Versatile Diffusion \cite{xu2023versatile} & CLIP & VAE & \checkmark & \checkmark & \checkmark & \href{https://github.com/SHI-Labs/Versatile-Diffusion}{Github} \\ 
        2023.01.17 & 23'CVPR & GLIGEN \cite{li2023gligen} & CLIP & CLIP & \checkmark & \checkmark & \checkmark & \href{https://gligen.github.io/}{Github} \\ 
        2023.10.03 & ~ & MiniGPT-5 \cite{zheng2023minigpt} & ~ & ViT & ~ & ~ & \checkmark & \href{https://github.com/eric-ai-lab/MiniGPT-5}{Github} \\ 
        2024.01.18 & ~ & DiffusionGPT \cite{qin2024diffusiongpt} & ~ & ~ & ~ & ~ & ~ & \href{https://diffusiongpt.github.io/}{Github} \\ 
        2024.01.22 & ~ & RPG-DiffusionMaster \cite{yang2024mastering} & ~ & ~ & \checkmark & \checkmark & ~ & \href{https://github.com/YangLing0818/RPG-DiffusionMaster}{Github} \\ 
        2024.01.24 & ~ & UNIMO-G \cite{li2024unimo} & MLLM & MLLM & \checkmark & \checkmark & \checkmark & \href{https://unimo-ptm.github.io/}{Github} \\ 
        2024.01.28 & ~ & CompAgent \cite{wang2024divide} & BLIP-2 & multi-modal & \checkmark & \checkmark & ~ & \href{https://zhenyuw16.github.io/CompAgent/}{Github} \\  
    \label{tab:mod}
\end{longtable}
}

{
\tiny
\begin{longtable}[!ht]{ccccccccccccccc}
    \caption{Comparison of Performance. \textbf{Para}: parameters. \textbf{Res}: resolution. \textbf{CLIP-S}: CLIP-Score.} \\
        \hline
        Model & Para & Datasets & Res & IS & FID & CLIP-I & CLIP-S & DINO & Laion & \makecell{Inference \\ Speed [s]} & \makecell{Training \\GPU days} \\
        \hline
        \endhead
        \hline
        \endfoot
        
        \multicolumn{12}{c}{\textbf{Autoregression}} \\ 
        DALL-E \cite{ramesh2021zero} & 12B & MS-COCO & ~ & 17.80  & 28.20  & ~ & ~ & ~ & ~ & ~ & V100 \\ 
        M6 \cite{lin2021m6} & 10B / 100B & ~ & ~ & ~ & ~ & ~ & ~ & ~ & ~ & ~ & ~ \\ 
        CogView \cite{ding2021cogview} & 4B & MS-COCO & ~ & 18.20  & 27.10  & ~ & ~ & ~ & ~ & ~ & V100 \\ 
        ImageBART \cite{esser2021imagebart} & 800M & CC & ~ & 15.27  & 22.61  & ~ & 23.00  & ~ & ~ & ~ & ~ \\ 
        NÜWA \cite{wu2022nuwa} & 870M & MS-COCO & 256  & 27.20  & 12.90  & ~ & ~ & ~ & ~ & 50.00  & 768 A100 \\ 
        Ernie-ViLG \cite{zhang2021ernie} & 10B & MS-COCO & 256  & ~ & 14.70  & ~ & ~ & ~ & ~ & ~ & ~ \\ 
        Parti \cite{yu2022scaling} & 20B & MS-COCO & 256  & ~ & 7.23  & ~ & ~ & ~ & ~ & ~ & ~ \\ 

        \multicolumn{12}{c}{\textbf{Autoregression Editing}} \\
        Make-A-Scene \cite{gafni2024scene} & 4B & MS-COCO & 256  & ~ & 7.55  & ~ & ~ & ~ & ~ & 25.00  & ~ \\ 
        
        \multicolumn{12}{c}{\textbf{Non-Autoregression}} \\ & ~ & ~ & ~ & ~ & ~ & ~ & ~ & ~ & ~ & ~ & ~ \\ 
        MaskGIT \cite{chang2022maskgit} & 227M & ImageNet & 256  & 182.10  & 6.18  & ~ & ~ & ~ & ~ & ~ & ~ \\ 
        MaskGIT \cite{chang2022maskgit} & 227M & ImageNet & 512  & 156.00  & 7.32  & ~ & ~ & ~ & ~ & ~ & ~ \\ 
        CogView2 \cite{ding2022cogview2} & 6B & MS-COCO & 256  & 22.40  & 24.00  & ~ & ~ & ~ & ~ & ~ & ~ \\ 
        Muse \cite{chang2023muse} & 3B & MS-COCO & 256  & ~ & 7.88  & ~ & 32.00  & ~ & ~ & 0.50  & TPUv4 \\ 
        Muse \cite{chang2023muse} & 3B & MS-COCO & 512  & ~ & ~ & ~ & ~ & ~ & ~ & 1.30  & ~ \\ 
        Emage \cite{feng2023emage} & 346M & MS-COCO & ~ & ~ & 24.97  & ~ & ~ & ~ & ~ & 0.95  & V100 \\ 
        aMUSEd \cite{patil2024amused} & 603M & MS-COCO & 256  & 23.82  & 38.70  & ~ & 25.97  & ~ & ~ & 0.20  & A100 \\ 
        aMUSEd \cite{patil2024amused} & 608M & ~ & 512  & 26.16  & 34.87  & ~ & 24.78  & ~ & 6.00  & 0.24  & A100 \\ 

        \multicolumn{12}{c}{\textbf{NAR Editing}} \\
        StyleDrop \cite{sohn2023styledrop} & ~ & ~ & ~ & ~ & ~ & ~ & 33.00  & ~ & ~ & ~ & A100 \\ 

        \multicolumn{12}{c}{\textbf{GAN}} \\
        GAN-INT-CLS \cite{reed2016generative} & ~ & MS-COCO & ~ & ~ & ~ & ~ & ~ & ~ & ~ & ~ & ~ \\ 
        AttnGAN \cite{xu2018attngan} & 230M & MS-COCO & 256  & 25.89  & 35.49  & ~ & ~ & ~ & ~ & ~ & ~ \\ 
        StackGAN \cite{zhang2017stackgan} & ~ & MS-COCO & 256  & 8.45  & ~ & ~ & ~ & ~ & ~ & ~ & ~ \\ 
        StackGAN++ \cite{zhang2018stackgan++} & ~ & MS-COCO & 256  & 8.45  & 74.05  & ~ & ~ & ~ & ~ & ~ & ~ \\ 
        Obj-GAN \cite{li2019object} & ~ & MS-COCO & 256  & 27.37  & 25.85  & ~ & ~ & ~ & ~ & ~ & ~ \\ 
        SD-GAN \cite{donahue2017semantically} & ~ & MS-COCO & 256  & 35.69  & ~ & ~ & ~ & ~ & ~ & ~ & RTX-3090 \\ 
        DM-GAN \cite{zhu2019dm} & 223M & MS-COCO & 256  & 30.49  & 32.64  & ~ & ~ & ~ & ~ & ~ & ~ \\ 
        MirrorGAN \cite{qiao2019mirrorgan} & ~ & MS-COCO & 256  & 26.47  & ~ & ~ & ~ & ~ & ~ & ~ & ~ \\ 
        ControlGAN \cite{li2019controllable} & ~ & MS-COCO & 256  & 24.06  & ~ & ~ & ~ & ~ & ~ & ~ & ~ \\ 
        OP-GAN \cite{hinz2020semantic} & 1019M & MS-COCO & 256  & 27.88  & 24.70  & ~ & ~ & ~ & ~ & ~ & ~ \\ 
        CP-GAN \cite{liu2020cp} & 318M & MS-COCO & 256  & 52.73  & ~ & ~ & ~ & ~ & ~ & ~ & ~ \\ 
        MA-GAN \cite{yang2021multi} & ~ & CUB & 256  & 4.76  & 21.66  & ~ & ~ & ~ & ~ & ~ & GTX 1080Ti \\ 
        DF-GAN \cite{tao2022df} & 19M & MS-COCO & 256  & ~ & 19.32  & ~ & ~ & ~ & ~ & ~ & ~ \\ 
        TReCS \cite{koh2021text} & ~ & LN-COCO & ~ & 21.30  & 48.70  & ~ & ~ & ~ & ~ & ~ & ~ \\ 
        DAE-GAN \cite{ruan2021dae} & 98M & MS-COCO & 256  & 35.08  & 28.12  & ~ & ~ & ~ & ~ & ~ & V100 \\ 
        XMC-GAN \cite{zhang2021cross} & 166M & MS-COCO & 256  & 30.45  & 9.33  & ~ & ~ & ~ & ~ & ~ & ~ \\ 
        LAFITE \cite{zhang2021lafite} & 75M & MS-COCO & 256  & 32.34  & 8.12  & ~ & ~ & ~ & ~ & 0.02  & V100 \\ 
        FuseDream \cite{liu2021fusedream} & 81.5M & MS-COCO & 256  & 34.26  & 21.16  & ~ & ~ & ~ & ~ & ~ & GTX 3090 \\ 
        VQGAN-CLIP \cite{crowson2022vqgan} & 227M & ~ & ~ & ~ & ~ & ~ & ~ & ~ & ~ & ~ & V100 \\ 
        StyleGAN-T \cite{sauer2023stylegan} & 1B & MS-COCO & 256  & ~ & 13.90  & ~ & 29.30  & ~ & ~ & 0.10  & 70 A100 \\ 
        GigaGAN \cite{kang2023scaling} & 1B & MS-COCO & 512  & ~ & 9.09  & ~ & 32.00  & ~ & ~ & 0.13  & 4,783 A100 \\ 

        \multicolumn{12}{c}{\textbf{GAN Editing}} \\
        StyleGAN-NADA \cite{gal2022stylegan} & ~ & ~ & 256  & ~ & ~ & ~ & ~ & ~ & ~ & ~ & V100 \\ 
        Text2LIVE \cite{bar2022text2live} & ~ & ~ & 512  & ~ & ~ & ~ & ~ & ~ & ~ & ~ & RTX 6000 \\ 
        Edit One for All \cite{nguyen2024edit} & ~ & ~ & ~ & ~ & ~ & ~ & ~ & ~ & ~ & ~ & ~ \\

        \multicolumn{12}{c}{\textbf{Diffusion}} \\ 
        LDM \cite{rombach2022high} & 1.45B & MS-COCO & 256  & 30.29  & 12.63  & ~ & - & ~ & ~ & 3.70  & A100 \\ 
        \href{https://huggingface.co/CompVis/stable-diffusion-v1-1}{SD1.1} & 860M & COCO2017 & 512  & ~ & 15.80  & ~ & 24.95  & ~ & ~ & ~ & 6,250 A100 \\ 
        \href{https://huggingface.co/CompVis/stable-diffusion-v1-2}{SD1.1 $\rightarrow$ SD1.2} & 860M & COCO2017 & 512  & ~ & 16.30  & ~ & 25.55  & ~ & ~ & ~ & A100 \\ 
        \href{https://huggingface.co/CompVis/stable-diffusion-v1-3}{SD1.2 $\rightarrow$ SD1.3} & 860M & COCO2017 & 512  & ~ & 15.40  & ~ & 25.75  & ~ & ~ & ~ & A100 \\ 
        \href{https://huggingface.co/CompVis/stable-diffusion-v1-4}{SD1.2 $\rightarrow$ SD1.4} & 860M & COCO2017 & 512  & ~ & 15.45  & ~ & 25.75  & ~ & ~ & ~ & A100 \\ 
        \href{https://huggingface.co/runwayml/stable-diffusion-v1-5}{SD1.2 $\rightarrow$ SD1.5} & 860M & COCO2017 & 512  & ~ & 15.38  & ~ & 25.85  & ~ & ~ & ~ & A100 \\ 
        \href{https://huggingface.co/stabilityai/stable-diffusion-2-base}{SD2-Base} & 865M & COCO2017 & 512  & ~ & ~ & ~ & ~ & ~ & ~ & ~ & 8,333 A100 \\ 
        \href{https://huggingface.co/stabilityai/stable-diffusion-2}{SD2-Base $\rightarrow$ SD2} & 865M & COCO2017 & 512  & ~ & 15.00  & ~ & 31.90  & ~ & ~ & ~ & A100 \\ 
        \href{https://huggingface.co/stabilityai/stable-diffusion-2-1}{SD2 $\rightarrow$ SD2.1} & 865M & COCO2017 & 512  & ~ & ~ & ~ & ~ & ~ & ~ & ~ & A100 \\ 
        SDXL \cite{podell2023sdxl} & 2.6B & COCO2017 & 256  & ~ & 4.40  & ~ & ~ & ~ & ~ & ~ & ~ \\ 
        SDXL-Turbo \cite{sauer2023adversarial} & 860M & ~ & 512  & ~ & 19.70  & ~ & 32.60  & ~ & ~ & 0.09  & ~ \\ 
        SD3 \cite{esser2024scaling} & 8B & COCO-2014 & 512  & ~ & 44.39  & ~ & 25.30  & ~ & ~ & ~ & ~ \\ 
        Imagen \cite{saharia2022photorealistic} & 6.6B & MS-COCO & 256  & ~ & 7.27  & ~ & ~ & ~ & ~ & 9.10  & 7,132 A100 \\ 
        Re-Imagen \cite{chen2022re} & 3.6B & MS-COCO & 256  & ~ & 5.25  & ~ & ~ & ~ & ~ & ~ & ~ \\ 
        \href{https://huggingface.co/DeepFloyd/IF-I-M-v1.0}{DeepFloyd-IF-I-M} & 400M & MS-COCO & ~ & ~ & 8.86  & ~ & 30.65  & ~ & ~ & ~ & A100 \\ 
        \href{https://huggingface.co/DeepFloyd/IF-I-L-v1.0}{DeepFloyd-IF-I-L} & 900M & MS-COCO & ~ & ~ & 8.06  & ~ & 30.35  & ~ & ~ & ~ & A100 \\ 
        \href{https://huggingface.co/DeepFloyd/IF-I-XL-v1.0}{DeepFloyd-IF-I-XL} & 4.3B & MS-COCO & ~ & ~ & 6.66  & ~ & 30.67  & ~ & ~ & ~ & A100 \\ 
        \href{https://huggingface.co/DeepFloyd/IF-II-M-v1.0}{DeepFloyd-IF-II-M} & 450M & MS-COCO & ~ & ~ & ~ & ~ & ~ & ~ & ~ & ~ & A100 \\ 
        \href{https://huggingface.co/DeepFloyd/IF-II-L-v1.0}{DeepFloyd-IF-II-L} & 1.2B & MS-COCO & ~ & ~ & ~ & ~ & ~ & ~ & ~ & ~ & A100 \\ 
        VQ-Diffusion \cite{gu2022vector} & 370M & MS-COCO & 256  & ~ & 19.75  & ~ & ~ & ~ & ~ & ~ & V100 \\ 
        Glide \cite{nichol2021glide} & 5B & MS-COCO & 256  & ~ & 12.89  & ~ & ~ & ~ & ~ & 15.00  & A100 \\ 
        kNN-Diffisuion \cite{sheynin2022knn} & 400M & MS-COCO & ~ & ~ & 12.50  & ~ & ~ & ~ & ~ & 7.00  & ~ \\ 
        DALL-E2 (unCLIP) \cite{ramesh2022hierarchical} & 4.5B & MS-COCO & 256  & ~ & 10.39  & ~ & ~ & ~ & ~ & ~ & 41,667 A100 \\ 
        Frido \cite{fan2023frido} & 698M & MS-COCO & 256  & 26.82  & 11.24  & 0.7046  & ~ & ~ & ~ & 0.92  & V100 \\ 
        Ernie-ViLG 2.0 \cite{feng2023ernie} & 3.5B & MS-COCO & 256  & ~ & 8.07  & ~ & ~ & ~ & ~ & ~ & 18 A100 \\ 
        Ernie-ViLG 2.0 \cite{feng2023ernie} & 24B & MS-COCO & 256  & ~ & 7.23  & ~ & ~ & ~ & ~ & ~ & ~ \\ 
        eDiff-I \cite{balaji2022ediff} & 6.8B & MS-COCO & 256  & ~ & 7.35  & ~ & ~ & ~ & ~ & ~ & A100 \\ 
        eDiff-I \cite{balaji2022ediff} & 7.1B & MS-COCO & 256  & ~ & 7.26  & ~ & ~ & ~ & ~ & ~ & A100 \\ 
        eDiff-I \cite{balaji2022ediff} & 8.1B & MS-COCO & 256  & ~ & 7.11  & ~ & ~ & ~ & ~ & ~ & A100 \\ 
        eDiff-I \cite{balaji2022ediff} & 9.1B & MS-COCO & 256  & ~ & 6.95  & ~ & ~ & ~ & ~ & ~ & A100 \\ 
        Corgi \cite{zhou2023shifted} & ~ & MS-COCO & ~ & ~ & 10.88  & ~ & ~ & ~ & ~ & ~ & A100 \\ 
        RAPHAEL \cite{xue2024raphael} & 3B & MS-COCO & 256  & ~ & 6.61  & ~ & ~ & ~ & ~ & ~ & 60,000 A100 \\ 
        Wuerstchen \cite{pernias2023wurstchen} & 0.99B & MS-COCO & 256  & 40.90  & 23.60  & ~ & ~ & ~ & ~ & ~ & 1,025 A100 \\ 
        InstaFlow \cite{liu2023instaflow} & 1.7B & MS-COCO & 512  & ~ & 11.83  & ~ & ~ & ~ & ~ & 0.12  & 199 A100 \\ 
        FreeU \cite{si2024freeu} & ~ & ~ & ~ & ~ & ~ & ~ & ~ & ~ & ~ & ~ & ~ \\ 
        PixArt-$\alpha$ \cite{chen2023pixart} & 0.6B & MS-COCO & ~ & ~ & 10.65  & ~ & ~ & ~ & ~ & ~ & 675 A100 \\ 
        DALL-E3 \cite{betker2023improving} & ~ & MS-COCO & ~ & ~ & ~ & ~ & 32.00  & ~ & ~ & ~ & ~ \\ 
        Predicated Diffusion \cite{sueyoshi2024predicated} & ~ & ~ & ~ & ~ & ~ & ~ & ~ & ~ & ~ & ~ & ~ \\ 
        UFOGen \cite{xu2024ufogen} & 0.9B & COCO2017 & ~ & ~ & 22.50  & ~ & 31.10  & ~ & ~ & 0.09  & ~ \\ 
        Ten \cite{wang2024generative} & ~ & ~ & ~ & ~ & ~ & ~ & ~ & ~ & ~ & ~ & ~ \\ 
        SwiftBrush \cite{nguyen2024swiftbrush} & ~ & MS-COCO & ~ & ~ & 16.67  & ~ & 29.00  & ~ & ~ & 0.04  & 4.11 A100 \\ 
        SCEdit \cite{jiang2024scedit} & ~ & COCO2017 & ~ & ~ & 13.82  & ~ & ~ & ~ & ~ & ~ & ~ \\ 
        PanGu-Draw \cite{lu2023pangu} & 5B & MS-COCO & ~ & ~ & 7.99  & ~ & ~ & ~ & ~ & ~ & ~ \\ 
        Prompt Expansion \cite{datta2023prompt} & ~ & ~ & ~ & ~ & ~ & ~ & ~ & ~ & 6.19  & ~ & ~ \\ 
        Self-Perception \cite{lin2023diffusion} & ~ & MS-COCO & ~ & 28.07  & 24.42  & ~ & ~ & ~ & 6.00  & ~ & ~ \\ 
        ConPreDiff \cite{yang2024improving} & ~ & MS-COCO & 256  & ~ & 6.21  & ~ & ~ & ~ & ~ & ~ & A100 \\ 
        Parrot \cite{lee2024parrot} & ~ & ~ & 512  & 31.32  & ~ & ~ & 16.67  & ~ & ~ & ~ & A100 \\ 
        RL-Diffusion \cite{zhang2024large} & ~ & PartiPrompts & ~ & ~ & ~ & ~ & ~ & ~ & 5.92  & ~ & A100 \\ 
        SceneGraph2Image \cite{mishra2024scene} & ~ & COCO-stuff & 256  & 30.18  & 38.12  & ~ & ~ & ~ & ~ & ~ & ~ \\ 
        Text2Street \cite{su2024text2street} & ~ & nuScenes & 512  & ~ & 53.92  & ~ & 17.81  & ~ & ~ & ~ & ~ \\ 
        Make a Cheap Scaling \cite{guo2024make} & ~ & Laion-5B & 1024  & ~ & 12.40  & ~ & ~ & ~ & ~ & ~ & ~ \\ 
        Multi-LoRA Composition \cite{zhong2024multi} & ~ & ComposLoRA & ~ & ~ & ~ & ~ & ~ & ~ & ~ & ~ & ~ \\ 
        HanDiffuser \cite{narasimhaswamy2024handiffuser} & ~ & ~ & ~ & ~ & 13.92  & ~ & ~ & ~ & ~ & ~ & A100 \\ 
        CogView3 \cite{zheng2024cogview3} & 3B & MS-COCO & 1024  & ~ & 31.63  & ~ & ~ & ~ & 6.01  & 10.33  & ~ \\ 
        PixArt-$\Sigma$ \cite{chen2024pixart} & 0.6B & ~ & 1024  & ~ & 5.51  & ~ & 27.30  & ~ & ~ & ~ & 20 V100 \\ 
        ~ & ~ & ~ & 4K & ~ & ~ & ~ & ~ & ~ & ~ & ~ & 12 A800 \\ 
        Giving a Hand to DM \cite{pelykh2024giving} & ~ & HaGRID & 512  & ~ & 1.81 (finger) & ~ & 34.01  & ~ & ~ & ~ & A1000 \\ 
        FouriScale \cite{huang2024fouriscale} & ~ & Laion-5B & 1024  & ~ & 23.62  & ~ & ~ & ~ & ~ & 540.00  & A100 \\ 
        YOSO \cite{luo2024you} & 0.9B & Laion & 512  & ~ & ~ & ~ & ~ & ~ & 5.84  & ~ & 10 A800 \\ 
        RLCM \cite{oertell2024rl} & ~ & ~ & ~ & ~ & ~ & ~ & ~ & ~ & ~ & ~ & ~ \\ 
        CosmicMan \cite{li2024cosmicman} & ~ & ~ & ~ & ~ & 36.78  & ~ & 28.47  & ~ & ~ & ~ & 224 A100 \\ 
        T-GATE \cite{zhang2024cross} & 815M & MS-COCO & 256  & ~ & 19.94  & ~ & ~ & ~ & ~ & 9.87  & 1080 Ti \\ 
        Diffusion2GAN \cite{kang2024distilling} & 0.9B & COCO2014 & ~ & ~ & 9.29  & ~ & ~ & ~ & ~ & 0.09  & 43.6 A100 \\

        \multicolumn{12}{c}{\textbf{Diffusion Editing}} \\
        SDEdit \cite{meng2021sdedit} & ~ & LSUN & 256  & ~ & ~ & ~ & ~ & ~ & ~ & 29.10  & 2080Ti \\ 
        DiffusionCLIP \cite{kim2022diffusionclip} & ~ & ~ & 256  & ~ & ~ & ~ & ~ & ~ & ~ & ~ & RTX 6000 \\ 
        Blended Diffusion \cite{avrahami2022blended} & ~ & ~ & ~ & ~ & ~ & ~ & ~ & ~ & ~ & 27.00  & A10 \\ 
        Prompt-to-Prompt \cite{hertz2022prompt} & ~ & ~ & ~ & ~ & ~ & ~ & ~ & ~ & ~ & ~ & ~ \\ 
        Textual Inversion \cite{gal2022image} & 1.4B & ~ & ~ & ~ & ~ & 0.7800  & 25.50  & ~ & ~ & ~ & V100 \\ 
        DreamBooth \cite{ruiz2023dreambooth} & ~ & ~ & ~ & ~ & ~ & 0.8030  & 30.50  & 0.668  & ~ & ~ & A100 \\ 
        UniTune \cite{valevski2023unitune} & ~ & ~ & ~ & ~ & ~ & ~ & ~ & ~ & ~ & ~ & ~ \\ 
        Imagic \cite{kawar2023imagic} & ~ & TEdBench & 512  & ~ & ~ & ~ & ~ & ~ & ~ & ~ & A100 \\ 
        DiffEdit \cite{couairon2023diffedit} & ~ & MS-COCO & 512  & ~ & ~ & ~ & ~ & ~ & ~ & 10.00  & Quadro GP100 \\ 
        MagicMix \cite{liew2022magicmix} & ~ & ~ & ~ & ~ & ~ & ~ & ~ & ~ & ~ & ~ & ~ \\ 
        InstructPix2Pix \cite{brooks2023instructpix2pix} & ~ & ~ & ~ & ~ & ~ & 0.8000  & 15.00  & ~ & ~ & ~ & 1.1 A100 \\ 
        Null-text Inversion \cite{mokady2023null} & ~ & MS-COCO & ~ & ~ & ~ & ~ & ~ & ~ & ~ & 10.00  & A100 \\ 
        DiffStyler \cite{huang2024diffstyler} & ~ & ~ & 256  & ~ & ~ & ~ & 28.69  & ~ & ~ & 18.00  & RTX 3090 \\ 
        Plug-and-Play Diffusion \cite{tumanyan2023plug} & ~ & ~ & ~ & ~ & ~ & ~ & 28.90  & ~ & ~ & ~ & ~ \\ 
        InST \cite{zhang2023inversion} & ~ & ~ & ~ & ~ & ~ & ~ & ~ & ~ & ~ & ~ & RTX3090 \\ 
        Custom Diffusion \cite{kumari2023multi} & ~ & MS-COCO & ~ & ~ & ~ & 0.7950  & ~ & ~ & ~ & ~ & A100 \\ 
        SmartBrush \cite{xie2023smartbrush} & ~ & MS-COCO & 512  & ~ & 5.76  & ~ & 24.90  & ~ & ~ & ~ & A100 \\ 
        Imagen Editor \cite{wang2023imagen} & 6.6B & EditBench & 512  & ~ & ~ & ~ & 31.50  & ~ & ~ & ~ & ~ \\ 
        ControlNet \cite{zhang2023adding} & 361M & ~ & 512  & ~ & 15.27  & ~ & 26.00  & ~ & 6.31  & 71.00  & 20.8 A100 \\ 
        T2I-Adapter \cite{mou2024t2i} & ~ & MS-COCO & ~ & ~ & ~ & ~ & ~ & ~ & ~ & ~ & 12 V100 \\ 
        Composer \cite{huang2023composer} & 4.4B & MS-COCO & 256  & ~ & 9.20  & ~ & 28.00  & ~ & ~ & ~ & ~ \\ 
        ELITE \cite{wei2023elite} & ~ & ~ & 512  & ~ & ~ & 0.7620  & 25.50  & 0.652  & ~ & ~ & V100 \\ 
        Cones \cite{liu2023cones} & ~ & ~ & ~ & ~ & ~ & 0.7250  & 36.10  & ~ & ~ & ~ & A100 \\ 
        SuTI \cite{chen2024subject} & 2.5B & DreamBench & 1024  & ~ & ~ & 0.8190  & 30.40  & 0.741  & ~ & 30.00  & Cloud TPU v4 \\ 
        InstantBooth \cite{shi2024instantbooth} & ~ & PPR10K & ~ & ~ & ~ & ~ & 31.40  & ~ & ~ & 6.00  & A100 \\ 
        MasaCtrl \cite{cao2023masactrl} & ~ & ~ & ~ & ~ & ~ & ~ & ~ & ~ & ~ & ~ & ~ \\ 
        BLIP-Diffusion \cite{li2024blip} & ~ & DreamBench & ~ & ~ & ~ & 0.8050  & 30.20  & 0.670  & ~ & ~ & 96 A100 \\ 
        UniControl \cite{qin2023unicontrol} & 1.5B & MS-COCO & ~ & ~ & ~ & ~ & ~ & ~ & ~ & ~ & 208.3 A100 \\ 
        ProSpect \cite{zhang2023prospect} & ~ & ~ & ~ & ~ & ~ & ~ & ~ & ~ & ~ & 3.00  & RTX3090 \\ 
        Semantic-Visual Alignment \cite{abreu2023addressing} & ~ & CIFAR100 & ~ & ~ & ~ & ~ & ~ & ~ & ~ & ~ & ~ \\ 
        Self-Guidance \cite{epstein2023diffusion} & ~ & ~ & ~ & ~ & ~ & ~ & ~ & ~ & ~ & ~ & ~ \\ 
        IP-Adapter \cite{ye2023ip} & ~ & COCO2017 & 512  & ~ & ~ & 0.8280  & ~ & ~ & ~ & ~ & V100 \\ 
        Text2Scene \cite{hwang2023text2scene} & ~ & ~ & ~ & ~ & ~ & ~ & ~ & ~ & ~ & ~ & V100 \\ 
        ControlStyle \cite{chen2023controlstyle} & ~ & MS-COCO & 512  & ~ & ~ & ~ & ~ & ~ & 6.09  & ~ & ~ \\ 
        Ranni \cite{feng2024ranni} & ~ & ~ & ~ & ~ & ~ & ~ & ~ & ~ & ~ & ~ & A100 \\ 
        StyleAligned \cite{hertz2024style} & ~ & ~ & ~ & ~ & ~ & ~ & 28.70  & 0.510  & ~ & 29/4 img & A100 \\ 
        Make-A-Storyboard \cite{su2023make} & ~ & ~ & 512  & ~ & ~ & 0.7380  & ~ & ~ & ~ & ~ & RTX 3090 \\ 
        PhotoMaker \cite{li2024photomaker} & ~ & Mystyle & 1024  & ~ & ~ & 0.7360  & 26.10  & 0.515  & ~ & 10.00  & 112 A100 \\ 
        ControlNet-XS \cite{zavadski2023controlnet} & 55M & MS-COCO & ~ & ~ & 16.36  & ~ & 29.21  & ~ & 6.09  & 38.00  & 8.3 A100 \\ 
        DreamDistribution \cite{zhao2023dreamdistribution} & ~ & ~ & ~ & ~ & ~ & 0.8400  & ~ & 0.500  & ~ & ~ & ~ \\ 
        One-Dimensional Adapter \cite{lyu2024one} & ~ & COCO2014 & ~ & ~ & 13.26  & ~ & ~ & ~ & ~ & ~ & 1 A100 \\ 
        ZONE \cite{li2024zone} & ~ & MS-COCO & 512  & ~ & ~ & 0.9688  & 29.69  & ~ & ~ & ~ & V100 \\ 
        PALP \cite{arar2024palp} & ~ & ~ & ~ & ~ & ~ & 0.6810  & 34.00  & ~ & ~ & ~ & ~ \\ 
        InstDiffEdit \cite{zou2024towards} & ~ & ImageNet & ~ & ~ & 55.30  & ~ & 24.90  & ~ & ~ & 10.80  & A100 \\ 
        WaveOpt-Estimator \cite{koo2024wavelet} & ~ & ~ & ~ & ~ & ~ & ~ & ~ & ~ & ~ & 28.00  & RTX 8000 \\ 
        BootPIG \cite{purushwalkam2024bootpig} & ~ & ~ & ~ & ~ & ~ & 0.7970  & 31.10  & 0.674  & ~ & ~ & A100 \\ 
        CreativeSynth \cite{huang2024creativesynth} & ~ & ~ & 1024  & ~ & ~ & 0.5207  & 59.12  & ~ & 7.56  & 5.00  & L40 \\ 
        Object-Driven One-Shot FT \cite{lu2024object} & ~ & ~ & ~ & ~ & ~ & 0.6431  & 28.00  & ~ & ~ & ~ & V100 \\ 
        FreeStyle \cite{he2024freestyle} & ~ & ~ & 1024  & ~ & ~ & ~ & 25.62  & ~ & 6.31  & 31.00  & A100 \\ 
        Pick-and-Draw \cite{lv2024pick} & ~ & DreamBench & ~ & ~ & ~ & 0.7900  & 30.30  & 0.696  & ~ & ~ & ~ \\ 
        SeFi-IDE \cite{li2024sefi} & ~ & ~ & 512  & ~ & ~ & ~ & 26.13  & ~ & ~ & ~ & ~ \\ 
        SepME \cite{zhao2024separable} & ~ & ~ & ~ & ~ & ~ & ~ & ~ & ~ & ~ & ~ & RTX 3090 \\ 
        MIGC \cite{zhou2024migc} & ~ & COCO-Position & ~ & ~ & 24.52  & ~ & 24.66  & ~ & ~ & 15.61  & 25 V100 \\ 
        LayerDiffusion \cite{zhang2024transparent} & ~ & ~ & ~ & ~ & ~ & ~ & ~ & ~ & ~ & ~ & 14.6 A100 \\ 
        Get What You Want \cite{li2024get} & ~ & MS COCO & ~ & ~ & ~ & 0.6857  & ~ & ~ & ~ & 35.00  & RTX 3090 \\ 
        CtrlColor \cite{liang2024control} & ~ & COCO-Stuff & 512  & ~ & 15.27  & 0.7871  & ~ & ~ & ~ & ~ & RTX 3090 \\ 
        DEADiff \cite{qi2024deadiff} & ~ & ~ & ~ & ~ & ~ & ~ & 28.40  & ~ & ~ & ~ & A100 \\ 
        Attribute-Control \cite{baumann2024continuous} \cite{} & ~ & ~ & ~ & ~ & ~ & ~ & ~ & ~ & ~ & ~ & ~ \\ 
        DisenDiff \cite{zhang2024attention} & ~ & ~ & ~ & ~ & ~ & ~ & ~ & ~ & ~ & ~ & ~ \\ 
        Factorized Diffusion \cite{geng2024factorized} & ~ & ~ & ~ & ~ & ~ & ~ & ~ & ~ & ~ & ~ & ~ \\ 

        \multicolumn{12}{c}{\textbf{Energy-Based Model}} \\
        PPGN \cite{nguyen2017plug} & ~ & ImageNet & 256  & 60.60  & ~ & ~ & ~ & ~ & ~ & ~ & ~ \\ 

        \multicolumn{12}{c}{\textbf{Mamba}} \\
        ZigMa \cite{hu2024zigma} & 134M & MS-COCO & 256  & ~ & 33.80  & ~ & ~ & ~ & ~ & ~ & 32 A100 \\ 

        \multicolumn{12}{c}{\textbf{Multimodal}} \\
        Versatile Diffusion \cite{xu2023versatile} & ~ & COCO-caption & 512  & ~ & 11.10  & ~ & ~ & ~ & ~ & ~ & A100 \\ 
        GLIGEN \cite{li2023gligen} & ~ & MS-COCO & ~ & ~ & 5.61  & ~ & ~ & ~ & ~ & ~ & V100 \\ 
        MiniGPT-5 \cite{zheng2023minigpt} & ~ & CC3M, & ~ & 24.38  & 59.48  & 0.7000  & ~ & ~ & ~ & ~ & A6000 \\ 
        DiffusionGPT \cite{qin2024diffusiongpt} & ~ & ~ & ~ & ~ & ~ & ~ & ~ & ~ & 5.70  & ~ & ~ \\ 
        RPG-DiffusionMaster \cite{yang2024mastering} & ~ & T2I-CompBench & ~ & ~ & ~ & ~ & ~ & ~ & ~ & ~ & ~ \\ 
        UNIMO-G \cite{li2024unimo} & 11B & MS-COCO & 256  & ~ & 8.36  & 0.8410  & 32.90  & 0.668  & ~ & ~ & ~ \\ 
        CompAgent \cite{wang2024divide} & ~ & T2I-CompBench & ~ & ~ & ~ & ~ & ~ & ~ & ~ & ~ & ~ \\ 
    \label{tab:perf}
\end{longtable}
}

\end{document}